\documentclass[sigconf]{acmart}
\usepackage{graphicx}
\usepackage{amsmath}

\usepackage{color}
\usepackage{bm}
\usepackage{algorithm}
\usepackage{algorithmic}
\usepackage{multirow}
\usepackage{subfigure}
\usepackage{amsthm}
\usepackage{booktabs}
\usepackage{makecell}
\usepackage{diagbox}

\usepackage{balance}
\usepackage{float}
\usepackage{subeqnarray}
\usepackage{cases}
\usepackage{tabularx}
\usepackage{lscape}

\usepackage{bbding}
\usepackage{wasysym}
\usepackage{arydshln} 
\usepackage{hyperref}
\hypersetup{colorlinks = true, 
	linkcolor = blue, 
	urlcolor = blue,
	citecolor = blue} 

\AtBeginDocument{%
  }

\copyrightyear{2023}
\acmYear{2023}
\setcopyright{acmlicensed}\acmConference[MM '23]{Proceedings of the 31st ACM International Conference on Multimedia}{October 29-November 3, 2023}{Ottawa, ON, Canada}
\acmBooktitle{Proceedings of the 31st ACM International Conference on Multimedia (MM '23), October 29-November 3, 2023, Ottawa, ON, Canada}
\acmPrice{15.00}
\acmDOI{10.1145/3581783.3612561}
\acmISBN{979-8-4007-0108-5/23/10}

\begin{document}

\title{Little Strokes Fell Great Oaks: Boosting the Hierarchical Features for Multi-exposure Image Fusion}

\author{Pan Mu}
\affiliation{%
  \institution{College of Computer Science and Technology, Zhejiang University of Technology}
  \streetaddress{}
  \city{}
  \state{}
  \country{}
  \postcode{}
}
\email{panmu@zjut.edu.cn}

\author{Zhiying Du}
\affiliation{%
  \institution{College of Computer Science and Technology, Zhejiang University of Technology}
  \streetaddress{}
  \city{}
  \country{}}
\email{202003150103@zjut.edu.cn}

\author{Jinyuan Liu}
\authornote{Corresponding author.}
\affiliation{%
  \institution{School of Mechanical Engineering, Dalian University of Technology}
  \city{}
  \country{}
}
\email{atlantis918@hotmail.com}

\author{Cong Bai}
\affiliation{%
 \institution{College of Computer Science and Technology, Zhejiang University of Technology}
 \streetaddress{}
 \city{}
 \state{}
 \country{}
}
\email{congbai@zjut.edu.cn}

\renewcommand{\shortauthors}{Pan Mu, Zhiying Du, Jinyuan Liu, \& Cong Bai}
\begin{abstract}
 In recent years, deep learning networks have made remarkable strides in the domain of multi-exposure image fusion. Nonetheless, prevailing approaches often involve directly feeding over-exposed and under-exposed images into the network, which leads to the under-utilization of inherent information present in the source images. Additionally, unsupervised techniques predominantly employ rudimentary weighted summation for color channel processing, culminating in an overall desaturated final image tone. To partially mitigate these issues, this study proposes a gamma correction module specifically designed to fully leverage latent information embedded within source images. Furthermore, a modified transformer block, embracing with self-attention mechanisms, is introduced to optimize the fusion process. Ultimately, a novel color enhancement algorithm is presented to augment image saturation while preserving intricate details. The source code is available at   \href{https://github.com/ZhiyingDu/BHFMEF}{https://github.com/ZhiyingDu/BHFMEF}.
\end{abstract}

\ccsdesc[500]{Networks~Network architectures}
\ccsdesc[300]{Theory of computation~Models of computation}
\ccsdesc[300]{Computer systems organization~Architectures}

\keywords{Deep Learning, Multi-Exposure Image Fusion, Gamma Correction}


\maketitle

\begin{figure}[t]
	\centering
	\includegraphics[width=\linewidth]{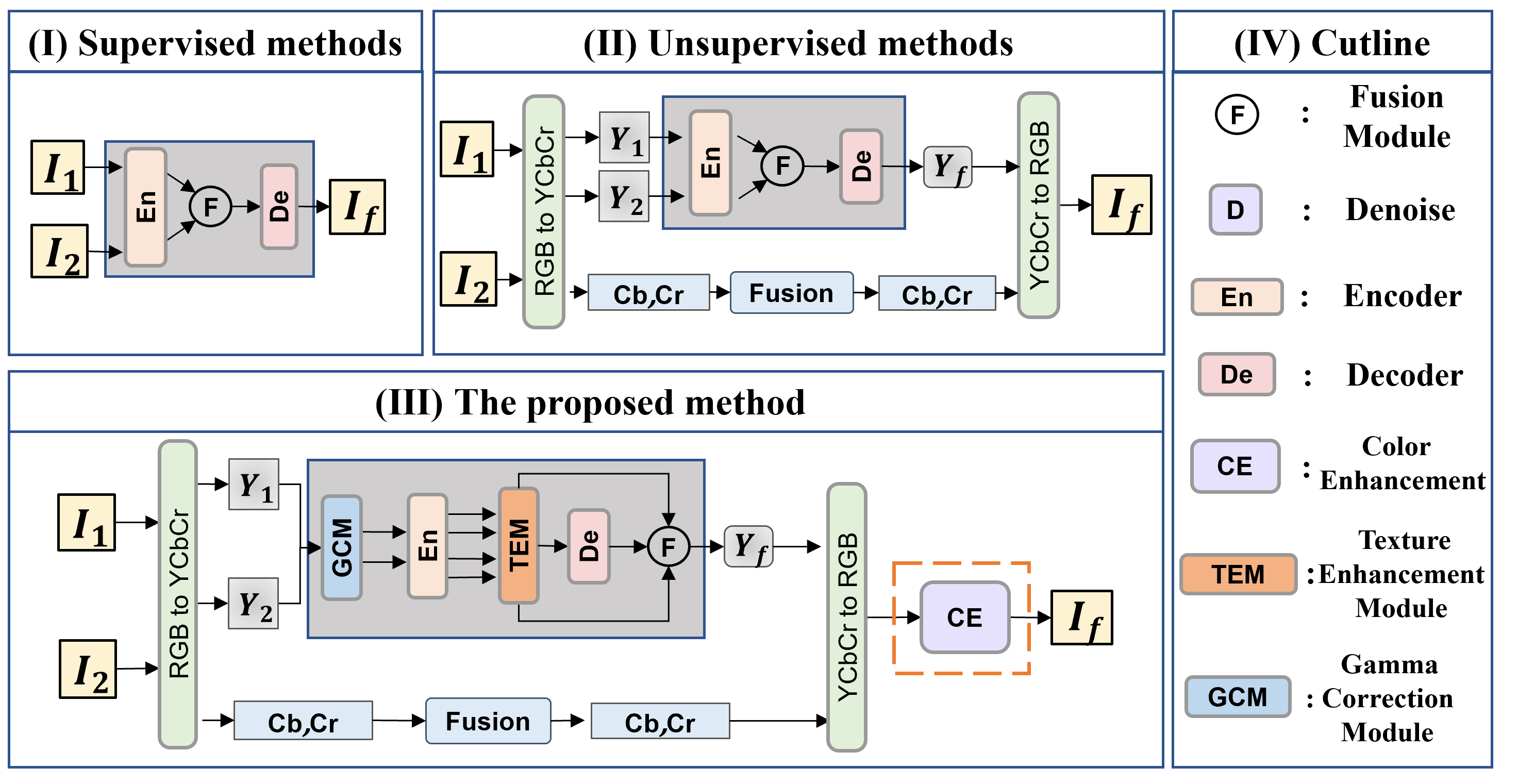}
	\caption{Workflow comparison of our proposed method with existing multi-exposure fusion approaches. 
	}
	\label{fig:overall}
\end{figure}

\section{Introduction}
	Capturing high-quality images of natural scenes using digital cameras is a challenging task, which often results in images containing over-exposed or under-exposed regions, and failing to fully represent the scene's visual information. The reason for these limitations can be attributed to the limited dynamic range of the camera sensor, which cannot cope with the broad dynamic range of natural scenes, despite the use of exposure time and aperture adjustments. The multi-exposure image fusion (MEF) provides an efficient solution to solve the above-mentioned problems by combining multiple low dynamic range (LDR) images with different exposure levels into one high dynamic range (HDR) image~\cite{5762601}.
	
	In recent years, various of multi-exposure image fusion methods have been proposed, and these methods are mainly divided into two categories: traditional methods~\cite{peng2021two,yuan2012automatic} and deep learning methods~\cite{FECNet,ECLNet,Huang_2022_CVPR,liu2020investigating}. In general, traditional multi-exposure image fusion methods include spatial domain-based methods \cite{7351094} and transform domain-based methods \cite{Wang2011ExposureFB}. Spatial domain-based methods directly operate on the pixel values of the input images to achieve fusion, while transform-domain based methods convert the input images into a transform domain, such as frequency domain or wavelet domain, and perform fusion in this transform space. Despite a long history and wide application, traditional multi-exposure image fusion methods still face challenges when dealing with complex scenes with nonlinear exposure response and large illumination changes.

	\begin{figure*}[t]
		\centering
		\includegraphics[width=0.95\linewidth]{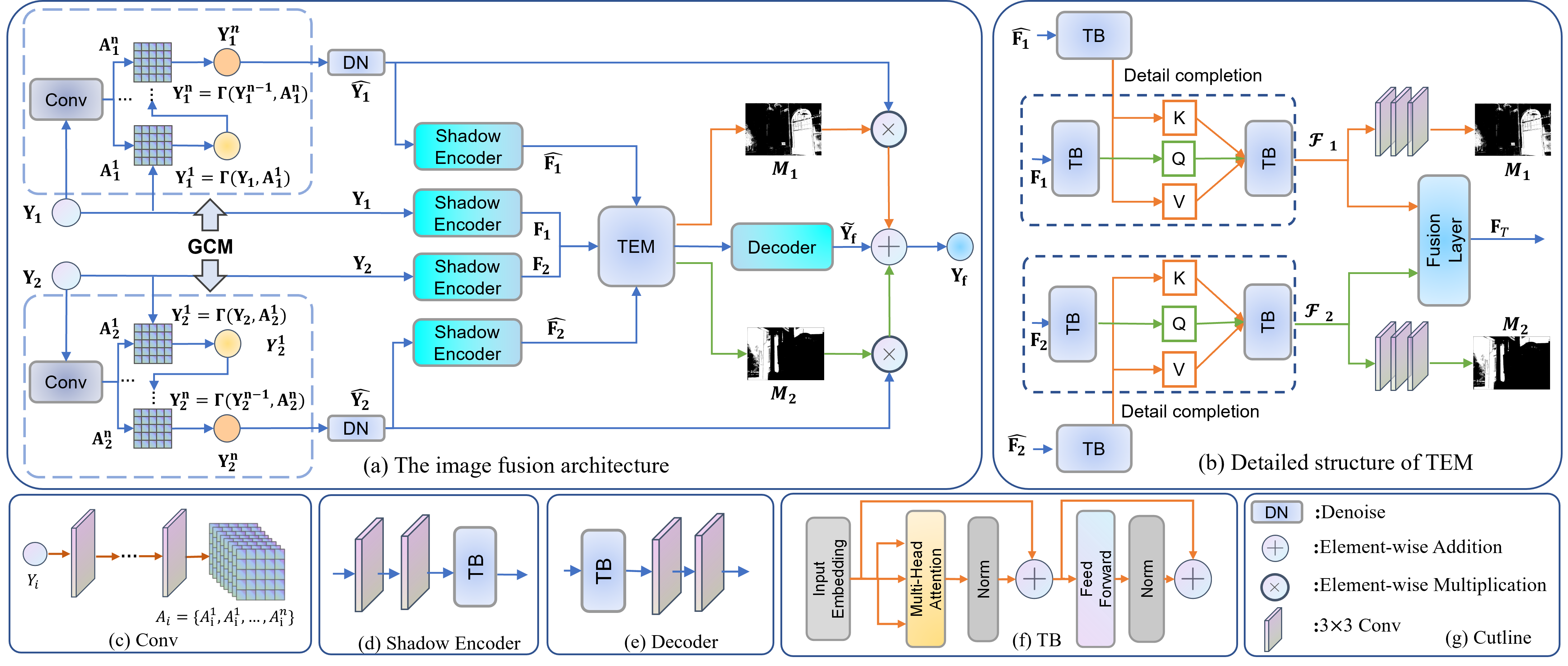}
		\caption{The architecture of the proposed method. The network consists of a gamma correction module (GCM), a denoise module (DN), a texture enhance module (TEM), a shadow encoder, a decoder. TB: transformer block.}
		\label{fig:architecture}
	\end{figure*}

	Deep learning has demonstrated its powerful representation capabilities in computer vision tasks~\cite{liu2023holoco,liu2020bilevel,liu2022learning,mu2021triple}. The initial deep learning-based MEF approaches predominantly utilized conventional CNN architectures, such as VGG and ResNet \cite{9009997,Olmos2019ABI}, to directly learn the mapping between input images and fused images. Nevertheless, these methods frequently suffer from a loss of texture and detail information in the fused images. To mitigate this issue, researchers incorporated attention mechanisms, such as spatial and channel attention, to enhance the fusion performance~\cite{li2022learning,liu2022attention}. 
	In general, deep learning-based MEF approaches have shown promising results and exhibit the potential to become a robust tool for multi-exposure image fusion tasks \cite{li2022learning}.
	
	However, there are still some limitations with these methods: 1) The employed techniques directly process under-exposed and over-exposed images \cite{10106641}, without fully exploiting the information embedded in the source images, thereby leading to the incomplete utilization of critical details \cite{FECNet,ECLNet,Huang_2022_CVPR}. 2) As depicted in Figure~\ref{fig:overall}-(i)(ii), the present methodologies rely on elementary fusion rules, including addition, concatenation and convolution, etc, which may cause the resultant fused images to display noticeable defects in intricate structures or distinct targets. 3) The ground-truth in current dataset is created using the existing methods and is obtained through post-processing. This results in the absence of a genuine ground-truth for multi-exposure image fusion (MEF). To alleviate this limitation, existing methods commonly utilize a technique depicted in Figure~\ref{fig:overall}-(ii). This approach involves converting the RGB color space into YCbCr, subsequently implementing the fusion strategy to fuse solely the Y channel, and utilizing a simplistic linear weighting approach to process the color. Unfortunately, this method may result in the produced image appearing faint and distorted.
	
	To address the above issues, this work propose a Boosting Hierarchical Features for Multi-exposure Image Fusion method (short for BHF-MEF), an unsupervised multi-exposure image fusion architecture with hierarchical feature extraction and deep fusion. 
	To fully leverage latent information embedded within source images, this study proposes a gamma correction module specifically designed to produce two novel images that incorporate obscured details from the originals. Furthermore, an modified transformer block, embracing with self-attention mechanisms, is introduced to optimize the fusion process. Ultimately, a novel color enhancement algorithm is presented to augment image saturation while preserving intricate details. 
	To ensure a fair and comprehensive comparison with other fusion methods, we utilized the latest multi-exposure image fusion benchmark dataset \cite{zhang2021benchmarking} as the test dataset. Furthermore, we incorporated three additional datasets as part of the test set. Subsequently, we compared the performance of our proposed method to ten competitive traditional and deep learning-based methods within the MEF domain. Our approach exhibited superior performance in both subjective and objective evaluations, thereby establishing its efficacy in comparison to existing state-of-the-art approaches. 	
	The main contributions of this paper are as follows:
	\begin{itemize}
		\item To effectively exploit the information present in the source image, we propose a Gamma Correction Module (GCM) that comprises five convolutional layers and an iterative process resulting in enriched images with significant latent detail information. 
%

		\item To fully exploit the detailed information in the source, we address the issue of some information loss during forward propagation by introducing attention-guided detail completion with Texture Enhance Module (TEM). An attention map is also utilized in the process to highlight the salient details.

		
		\item To address the issue of unsupervised fusion method producing faint and distorted results, we propose a color enhancement trick, named CE which modifies the RGB channels using the S and L channels of the HSL color domain, resulting in an image with enhanced color information.
	\end{itemize}

\section{RELATED WORKS}
	In this section, we briefly reviews the representative works of deep learning-based multi-exposure image fusion approaches. 
	In recent years, deep learning has achieved significant breakthroughs in computer vision tasks and has also been successfully applied in the field of MEF~\cite{zhang2021image} which can be broadly categorized into two types: supervised and unsupervised approaches~\cite{xu2020mef,zhang2020ifcnn,xu2020u2fusion,liu2022learning,LowLightZhang,liu2021retinex}. Some methods convert the MEF task into a supervised optimization problem. For example, Xu et al. proposed MEF-GAN~\cite{xu2020mef} which leverages generative adversarial networks. Zhang et al. developed IFCNN~\cite{zhang2020ifcnn} which uses two branches to extract features from each source image and fuses them element-wise. However, since ground truth is not available for MEF, the performance of supervised approaches is inherently limited by the existing methods. Therefore, unsupervised deep methods have become the mainstream research direction for MEF. For instance, methods such as U2Fusion \cite{xu2020u2fusion}, SDNet~\cite{zhang2021sdnet}, and FusionDN~\cite{xu2020fusiondn} calculate the gradient and intensity distances between the fused image and the multi-exposed source images, enabling them to preserve scene textures and adjust the illumination distribution.
	
	Notably, most unsupervised MEF methods do not handle color information directly. Specifically, U2Fusion, DeepFuse, SDNet, and other similar methods only operate on the Y channel, while the Cb and Cr channels are usually fused using a simple linear formula:
	\begin{equation}\label{equ:CbCr}
		C_f = \frac{C_1(\vert C_1-\tau\vert)+C_2(\vert C_2-\tau\vert)}{\vert C_1-\tau\vert+\vert C_2-\tau\vert}
	\end{equation}
	where $C_1$ and $C_2$ donote the Cb (or Cr) channels of input image pair, and $C_f$ is the corresponding fused chrominance channel. $\tau$ usually is set as the median of dynamic range, e.g., 128. The final color fused image is obtained by transforming the YCbCr into RGB space. However, this kind of compromise operation will may result in the produced image appearing faint and distorted.

	\begin{figure*}[htbp] 
		\centering
		\begin{center}
			\begin{tabular}{c@{\extracolsep{0.1mm}}c@{\extracolsep{0.1mm}}c@{\extracolsep{0.1mm}}c@{\extracolsep{0.1mm}}c@{\extracolsep{0.1mm}}c}		
				\includegraphics[width=0.166\textwidth]{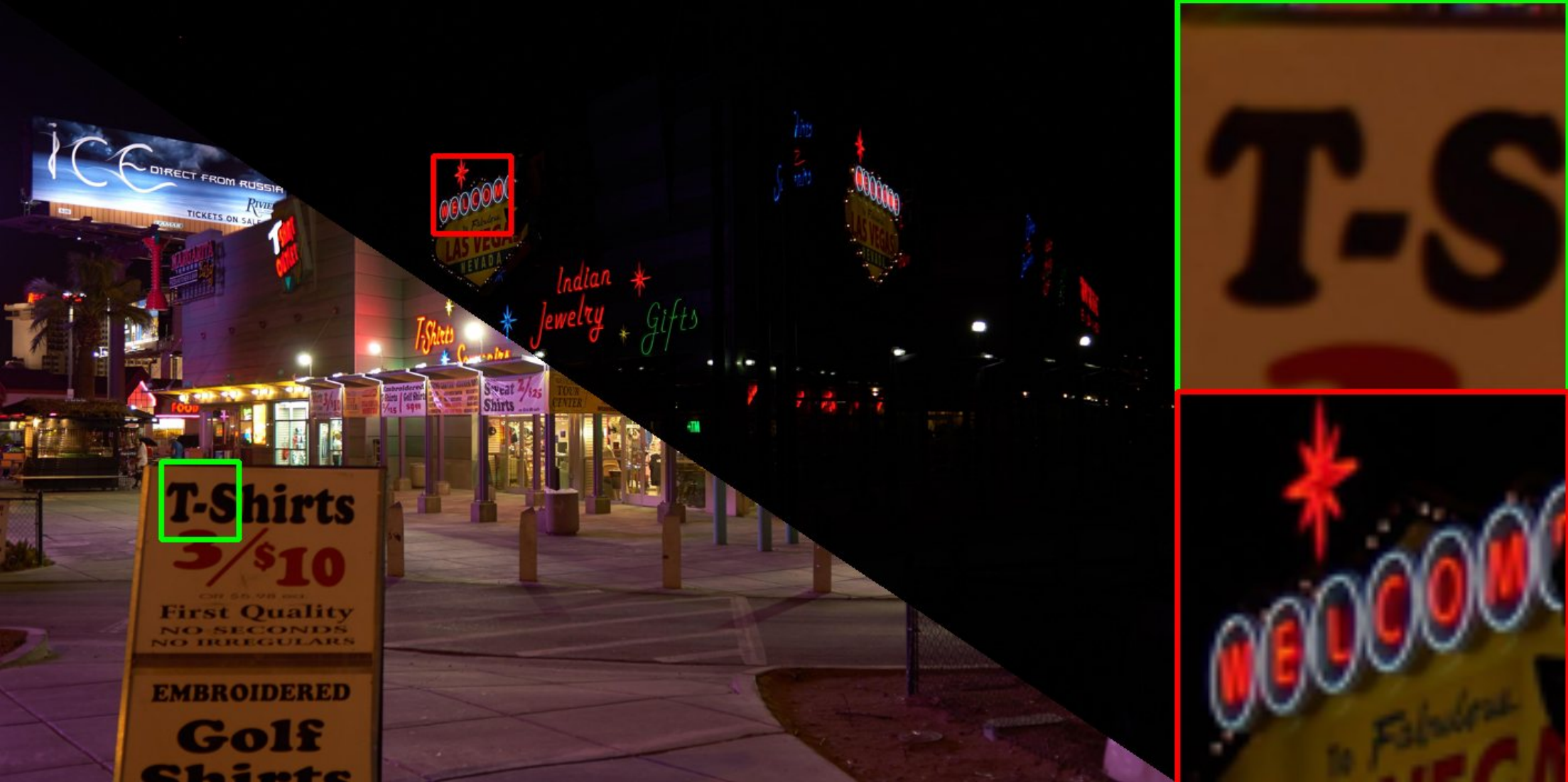}&
				\includegraphics[width=0.166\textwidth]{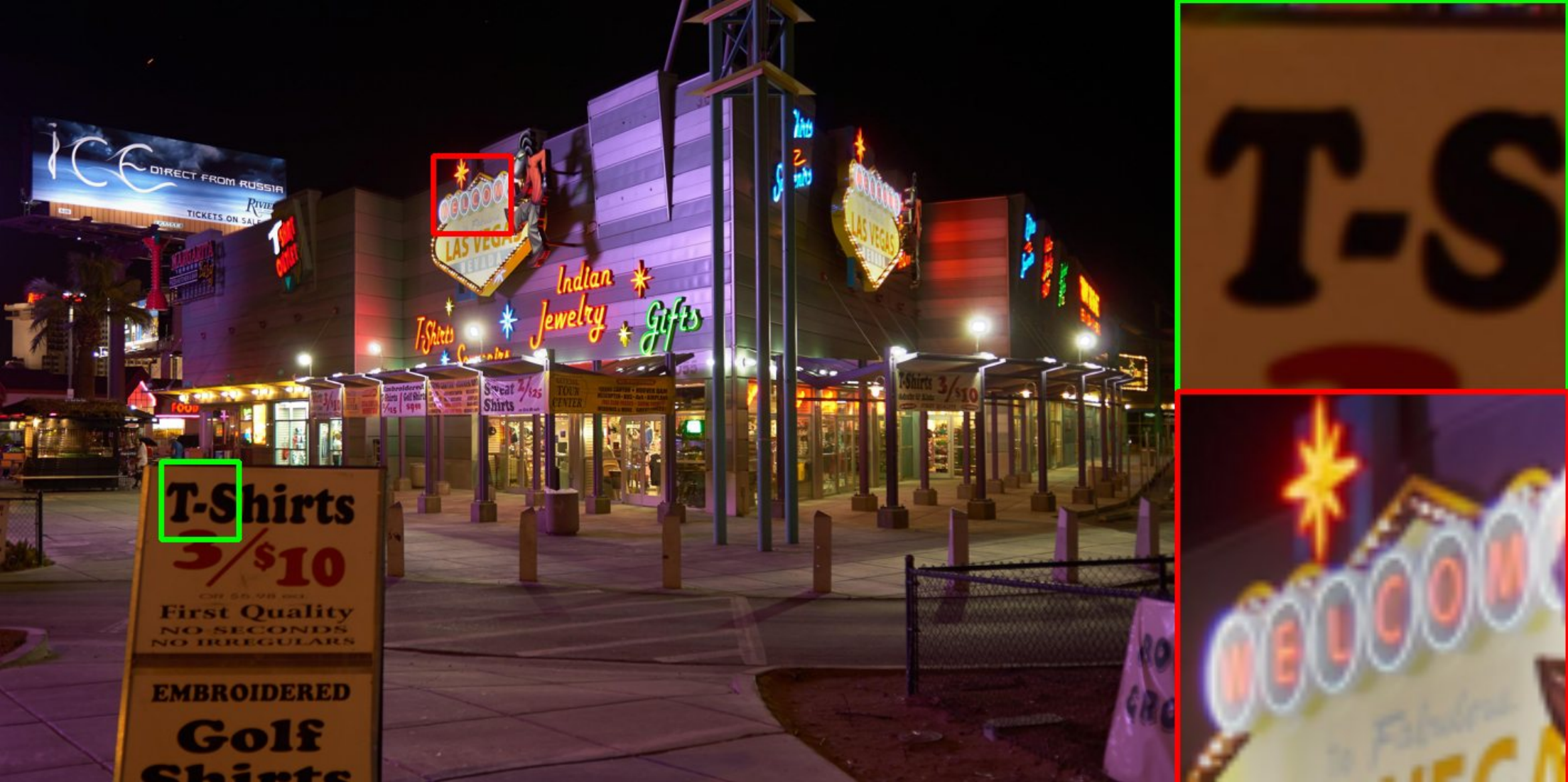}&
				\includegraphics[width=0.166\textwidth]{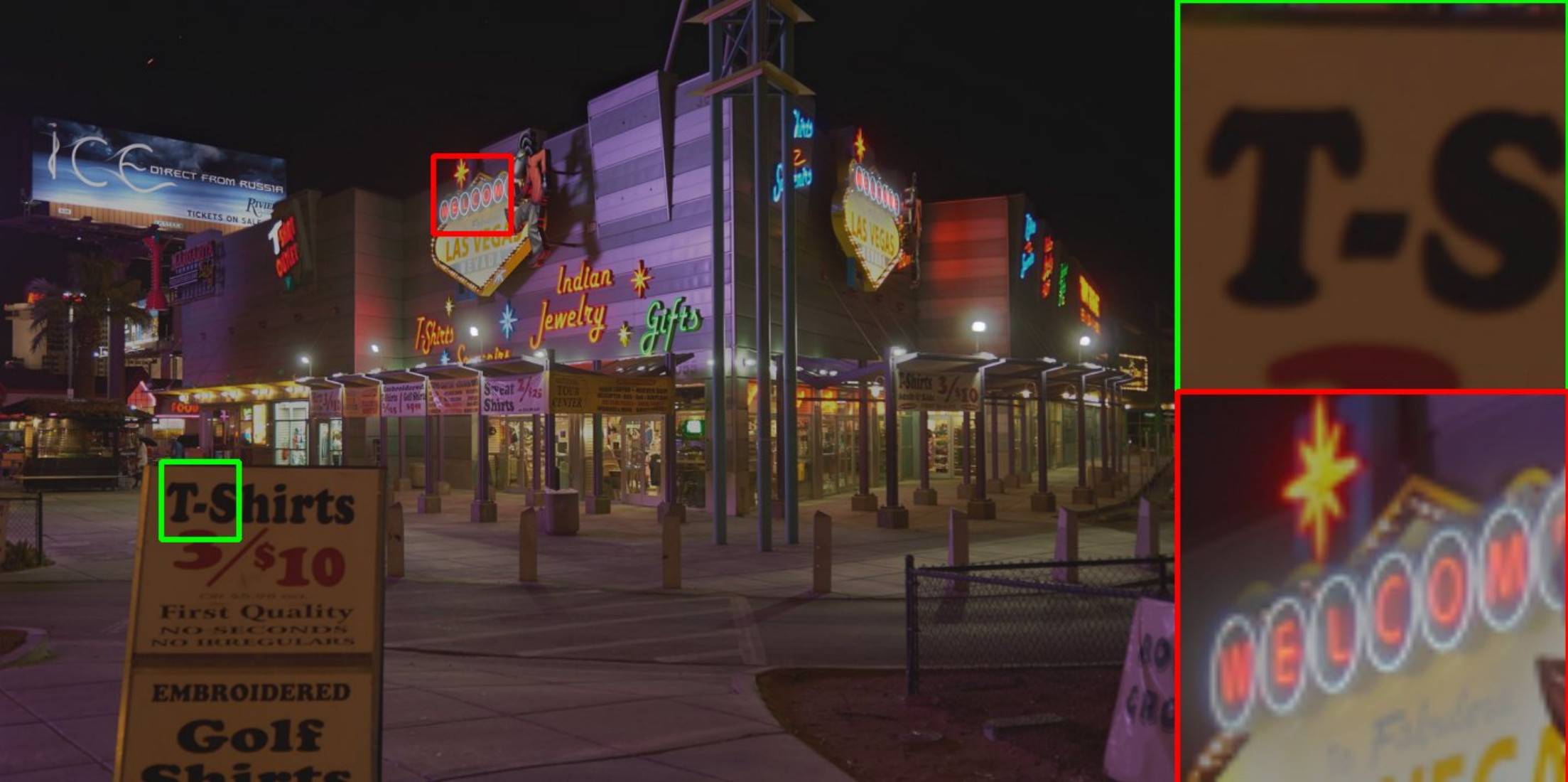}&
				\includegraphics[width=0.166\textwidth]{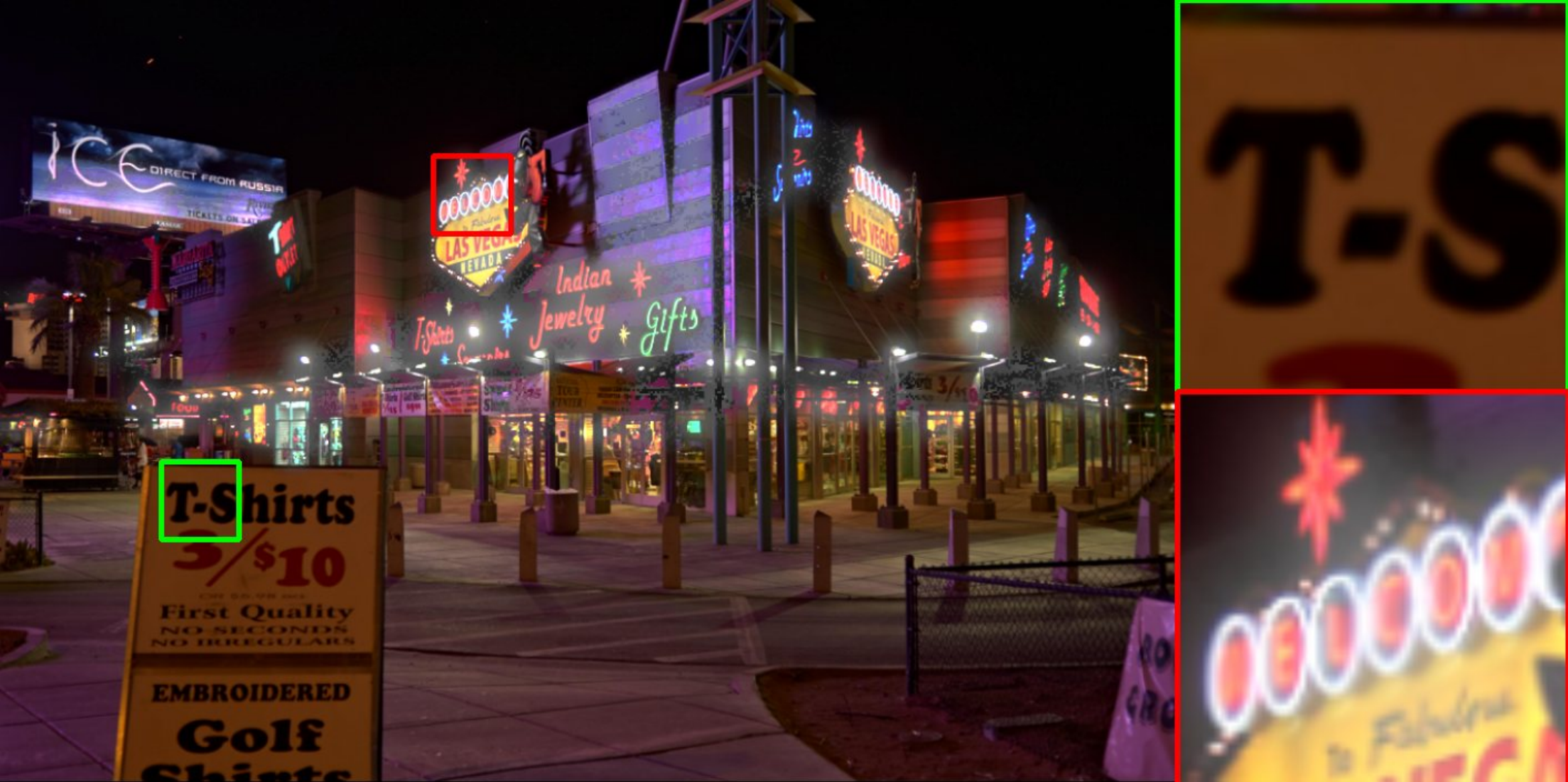}&
				\includegraphics[width=0.166\textwidth]{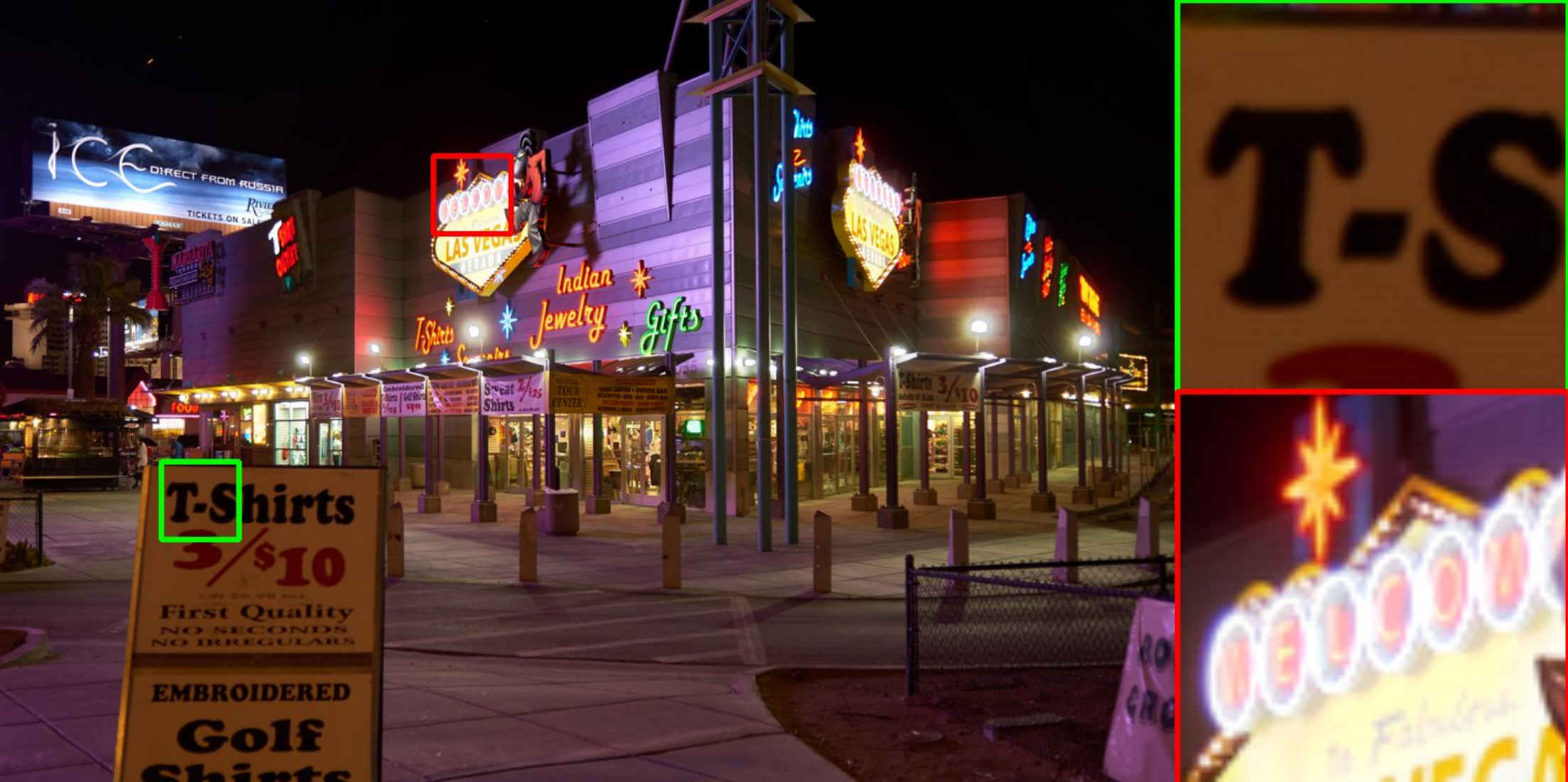}&
				\includegraphics[width=0.166\textwidth]{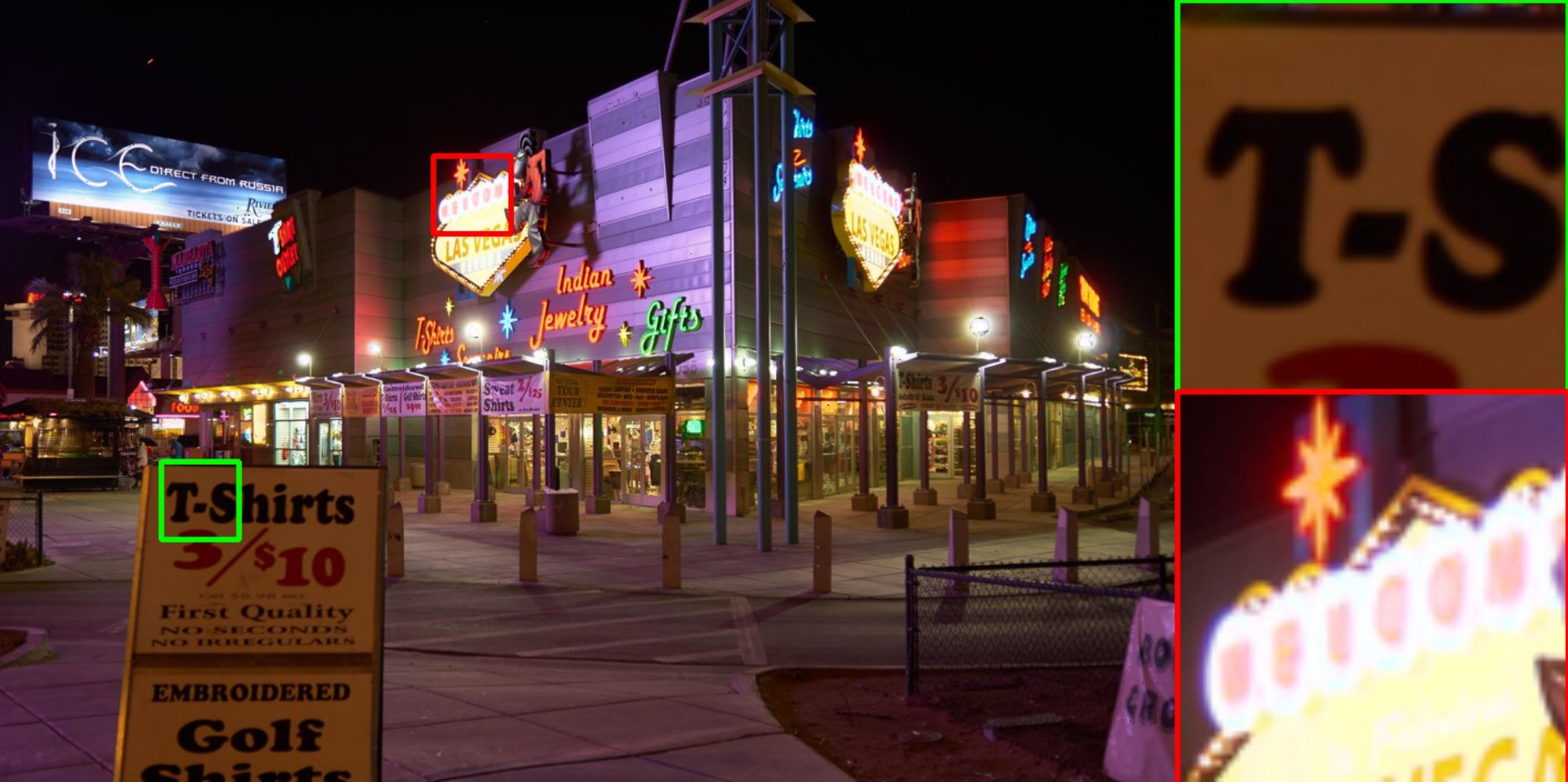}\\
				
				\includegraphics[width=0.166\textwidth]{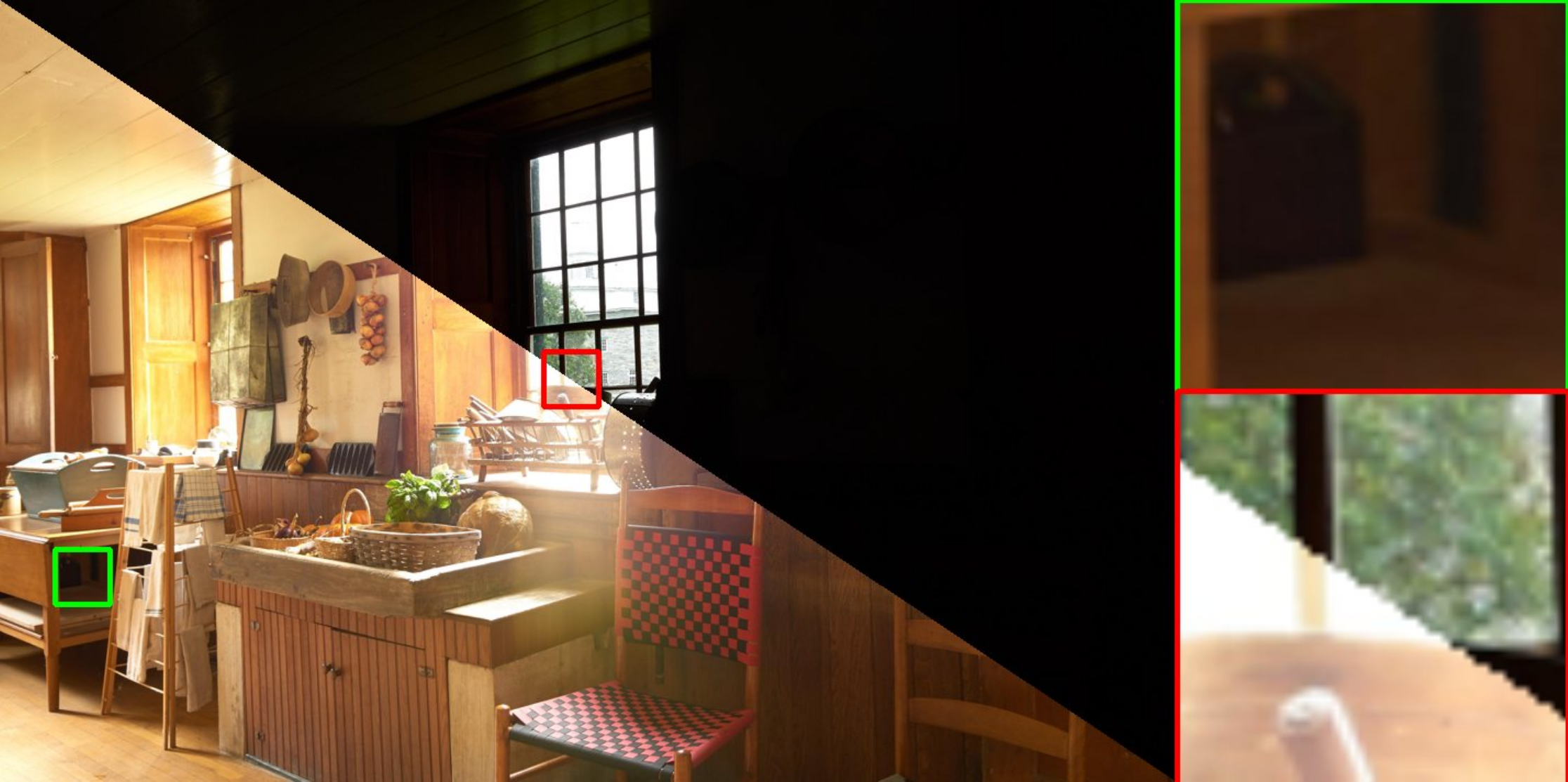}&
				\includegraphics[width=0.166\textwidth]{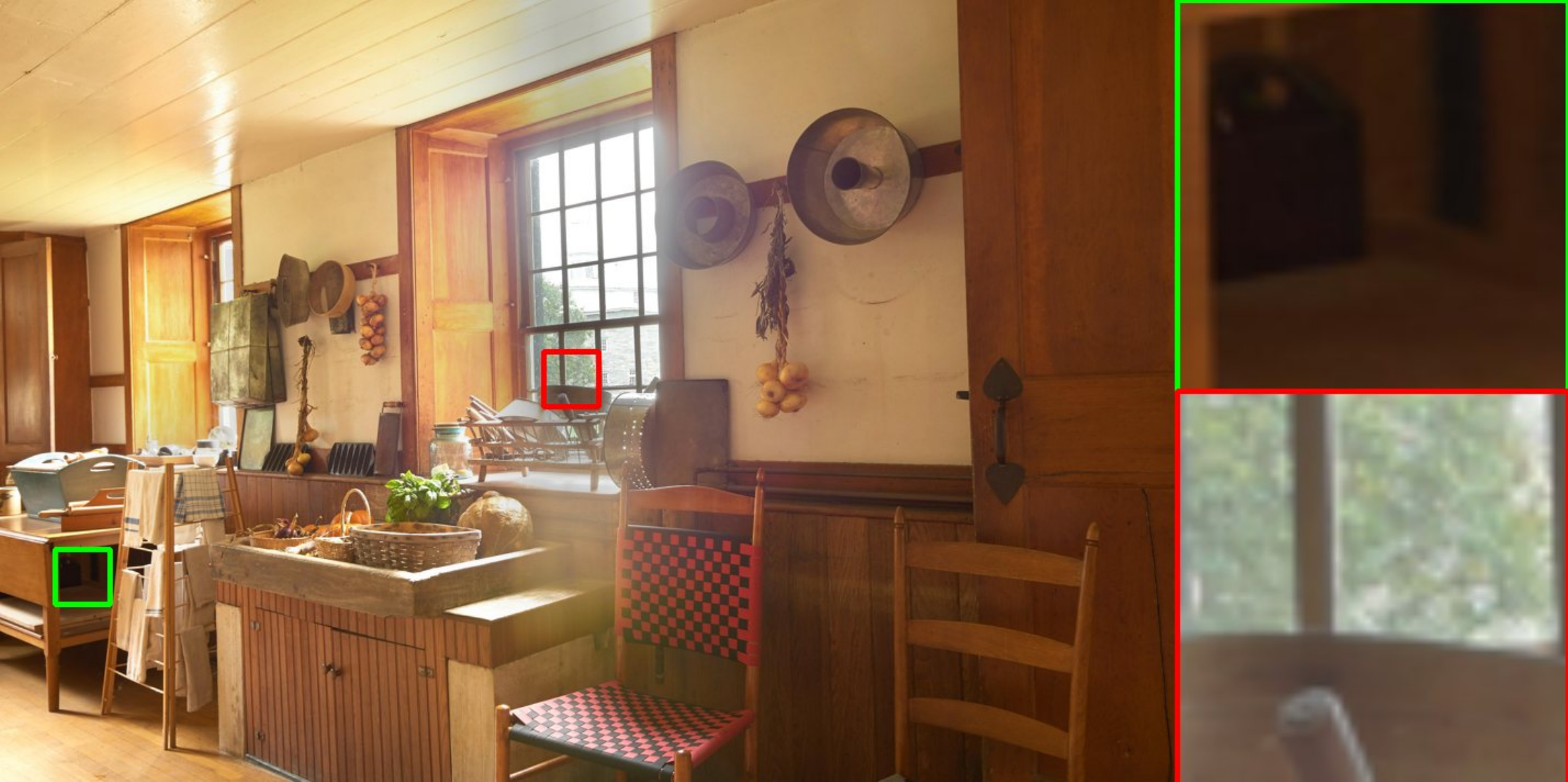}&
				\includegraphics[width=0.166\textwidth]{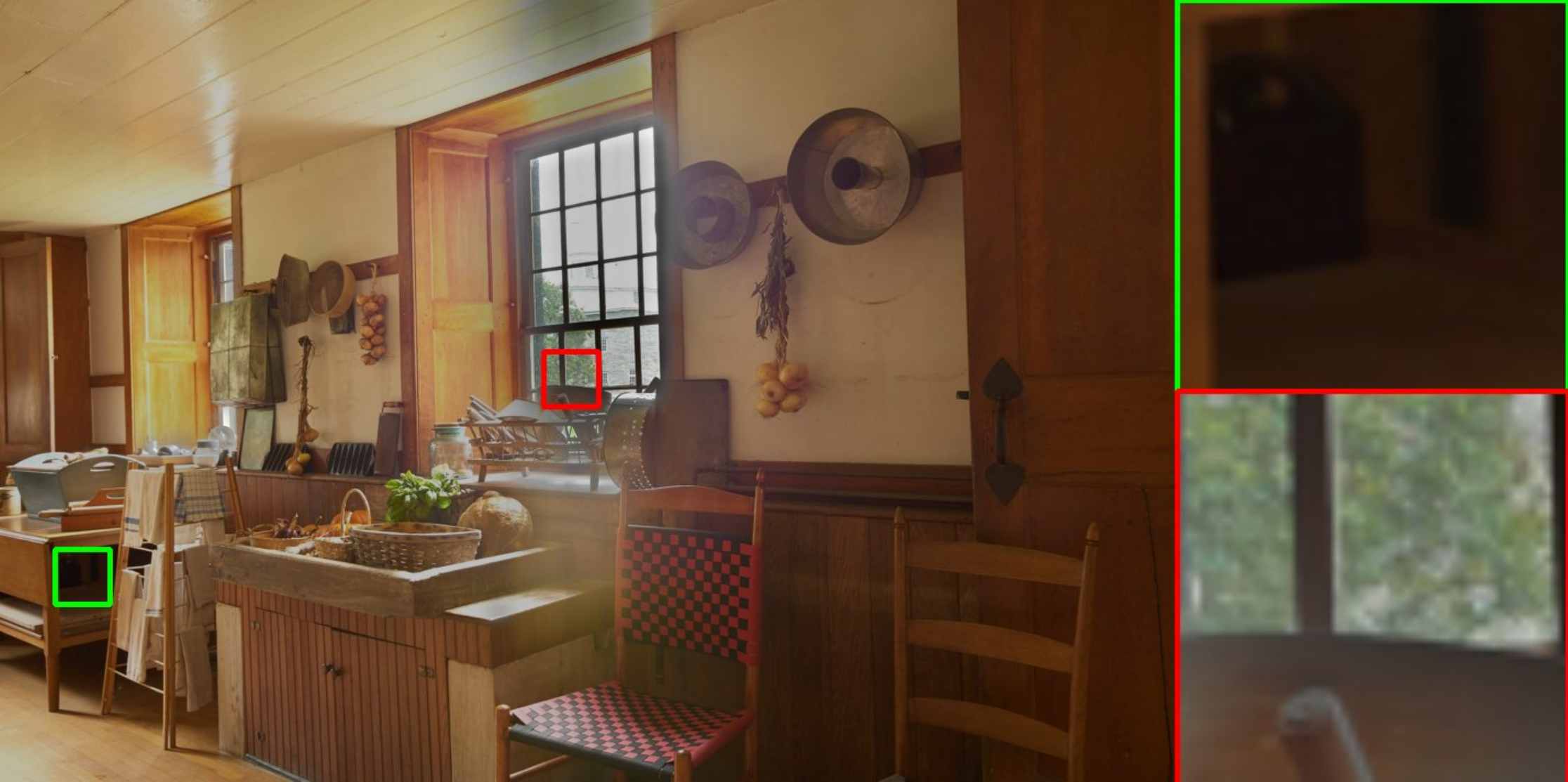}&
				\includegraphics[width=0.166\textwidth]{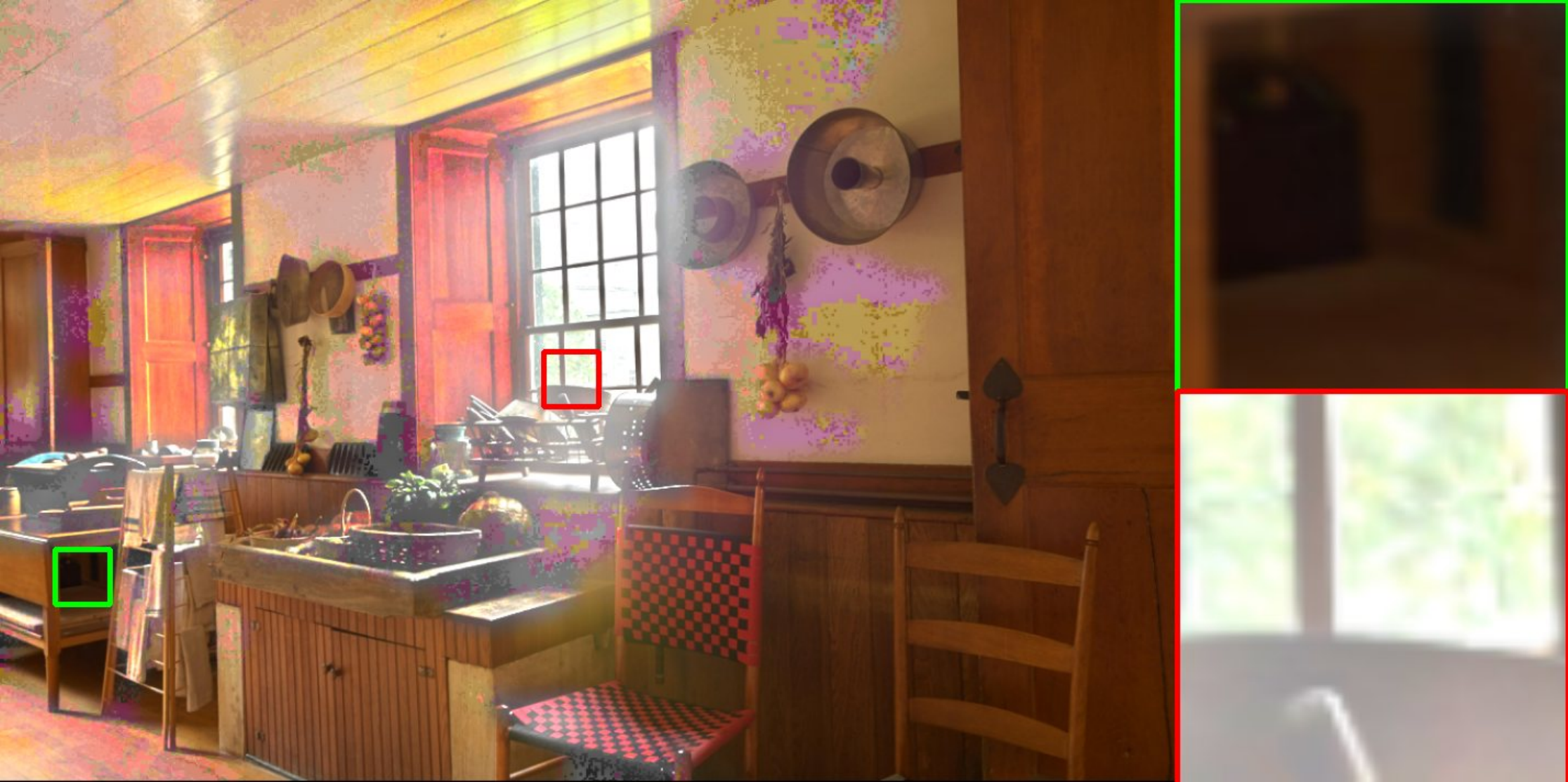}&
				\includegraphics[width=0.166\textwidth]{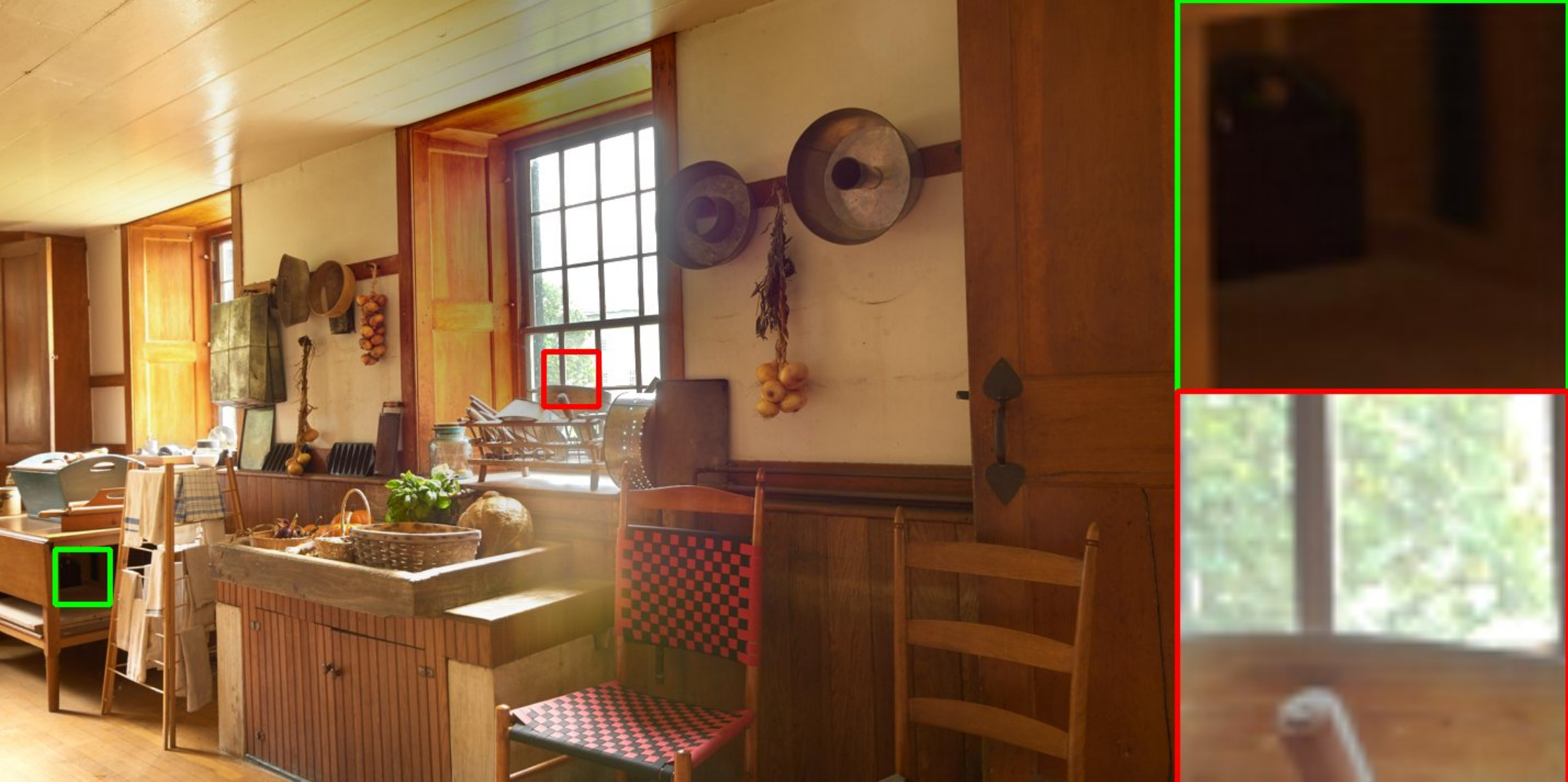}&
				\includegraphics[width=0.166\textwidth]{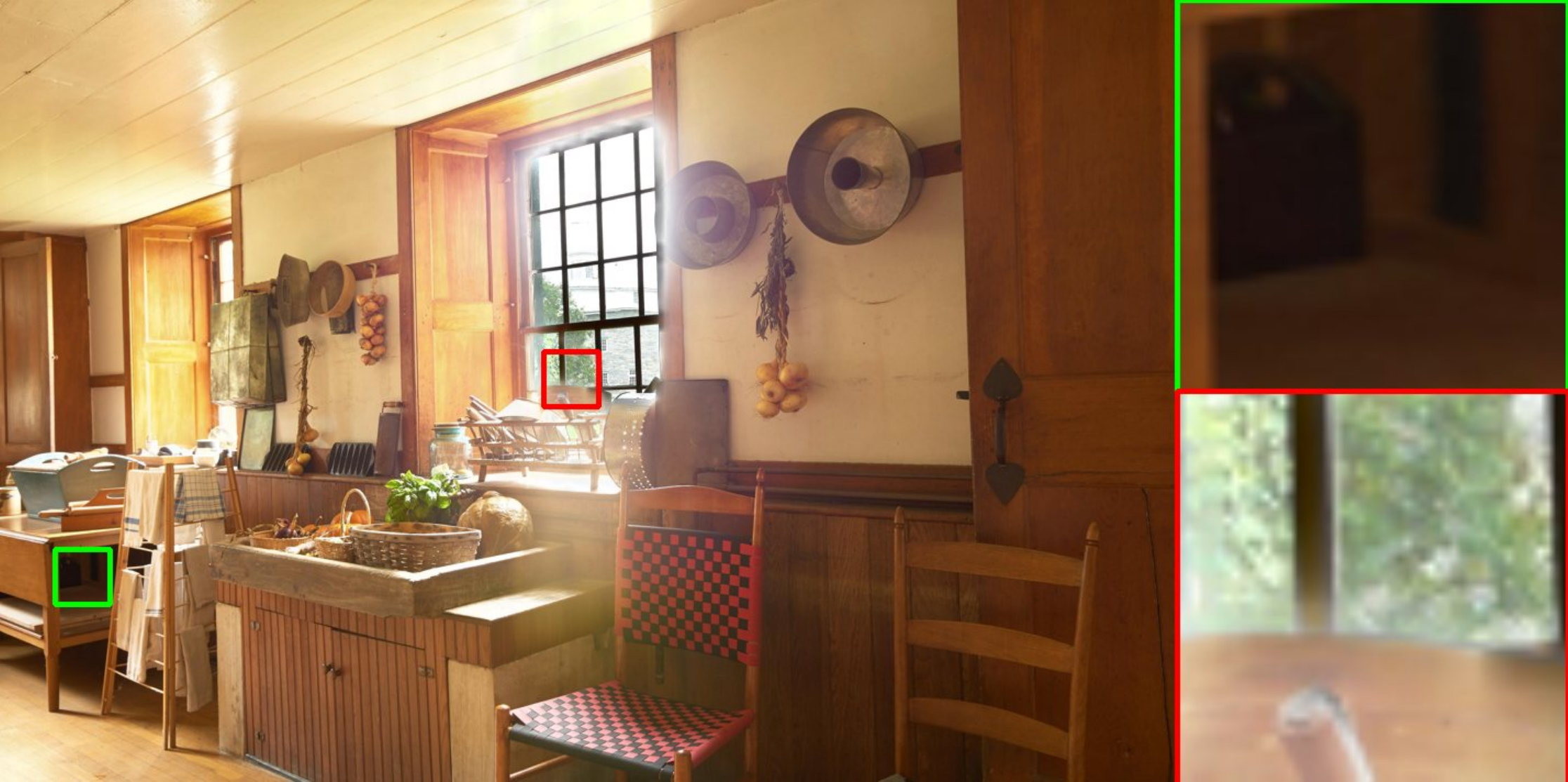}\\
				
				\includegraphics[width=0.166\textwidth]{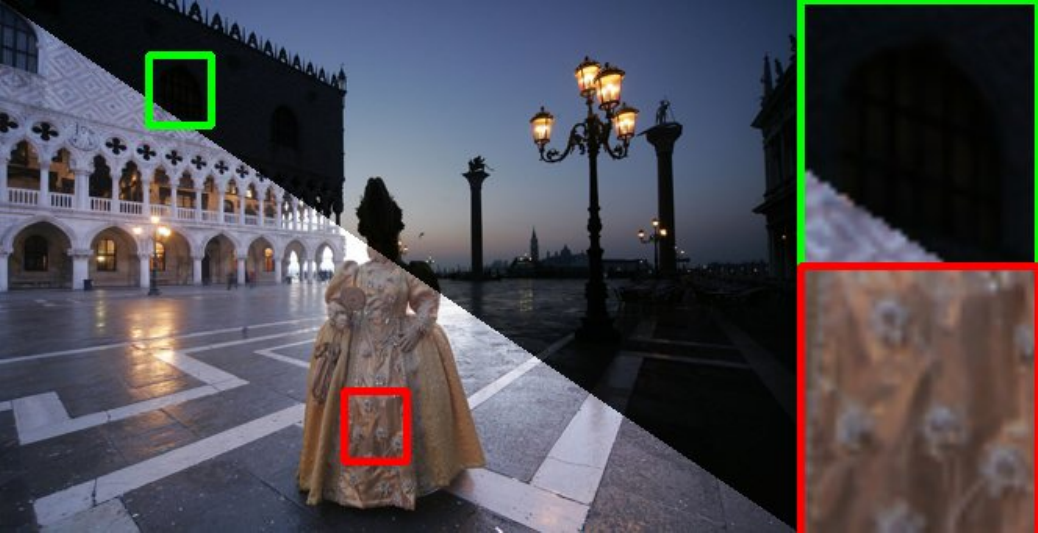}&
				\includegraphics[width=0.166\textwidth]{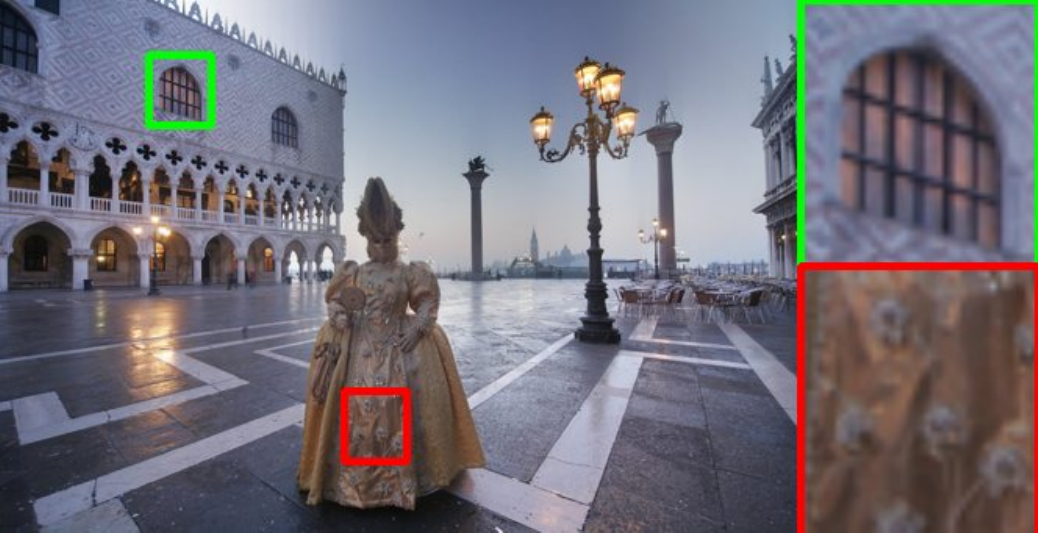}&
				\includegraphics[width=0.166\textwidth]{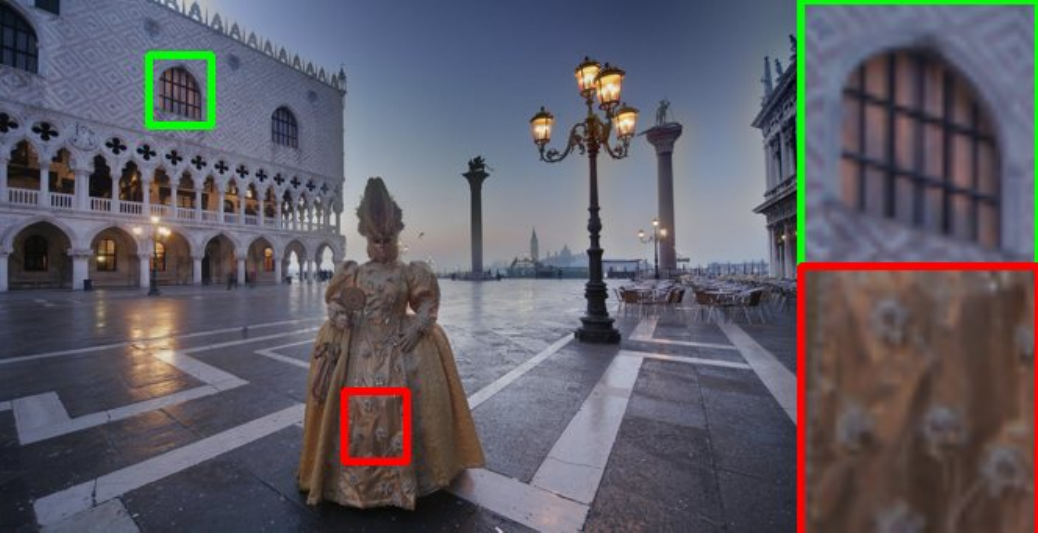}&
				\includegraphics[width=0.166\textwidth]{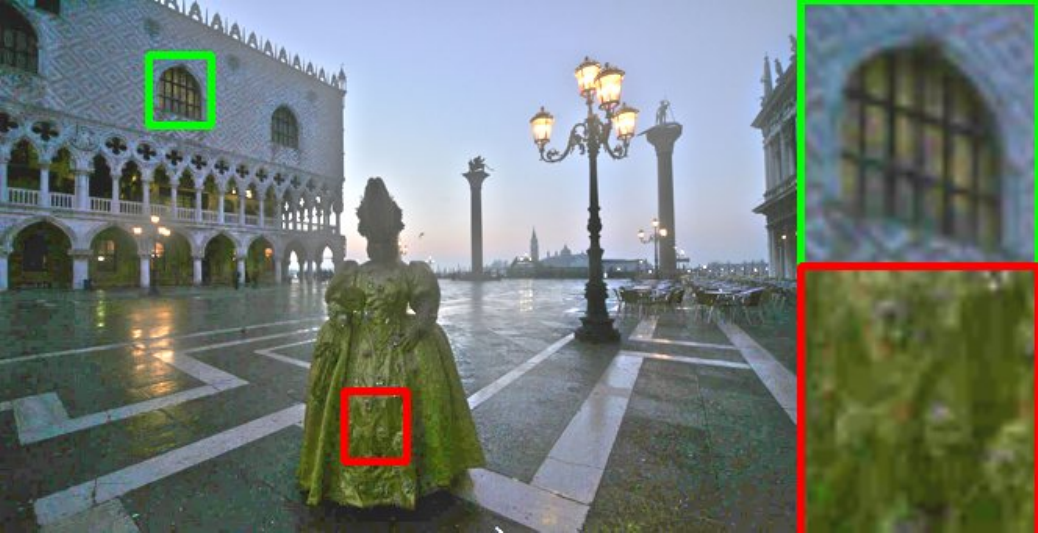}&
				\includegraphics[width=0.166\textwidth]{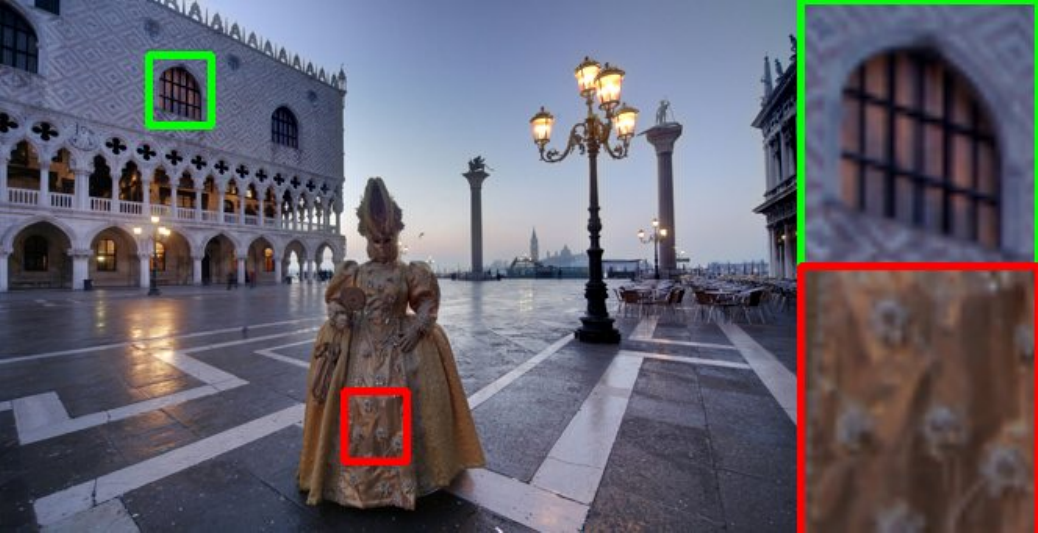}&
				\includegraphics[width=0.166\textwidth]{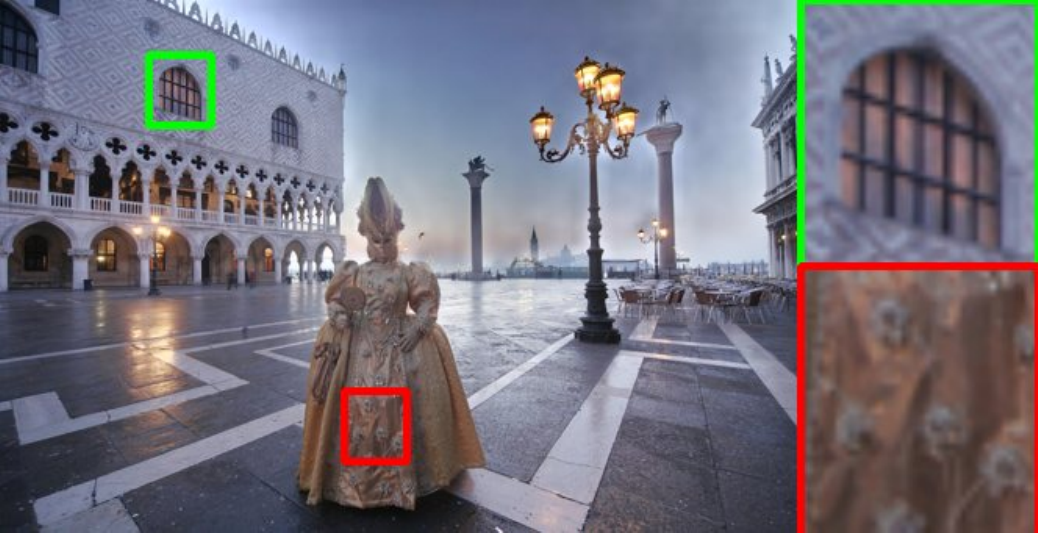}\\
				
				\includegraphics[width=0.166\textwidth]{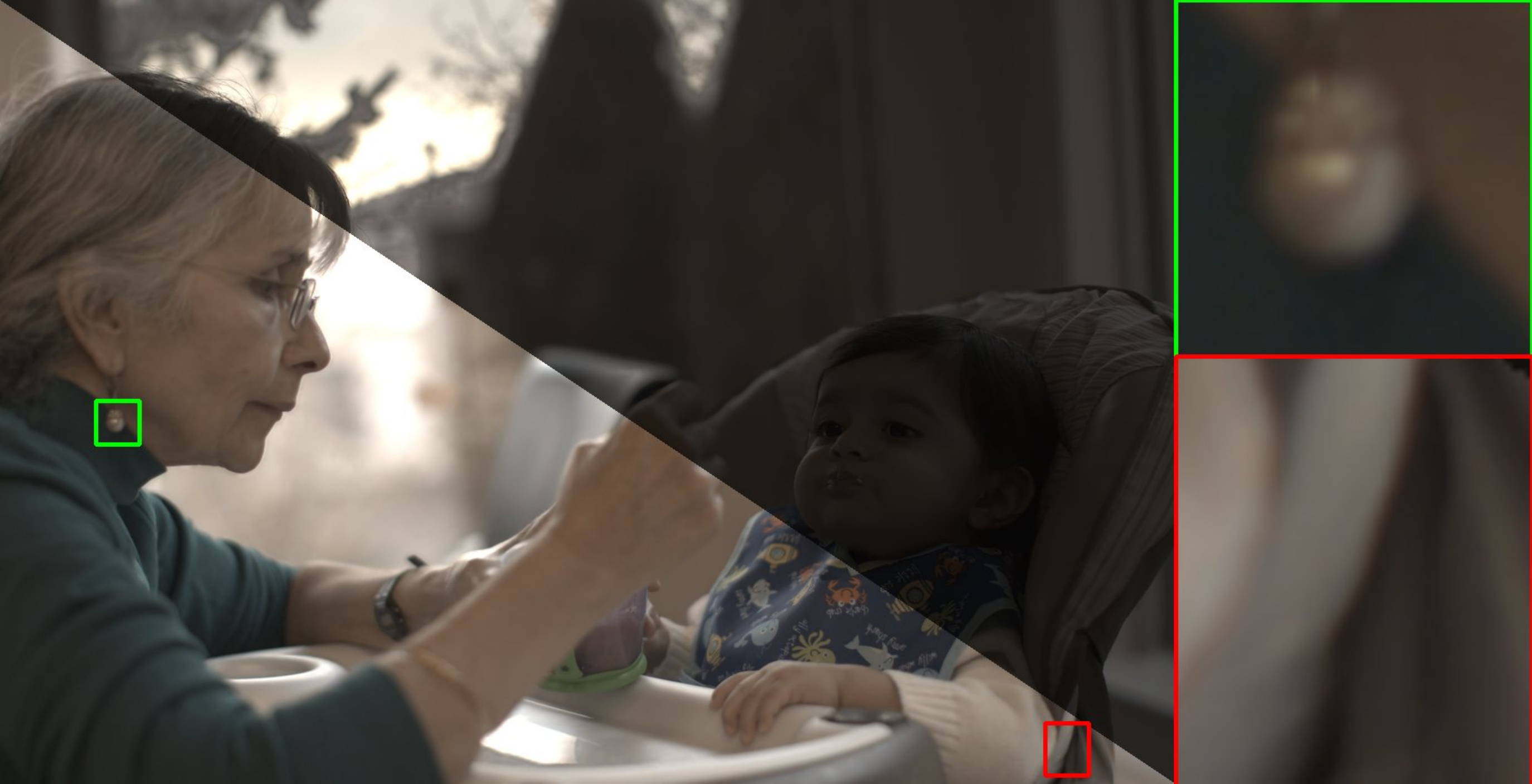}&
				\includegraphics[width=0.166\textwidth]{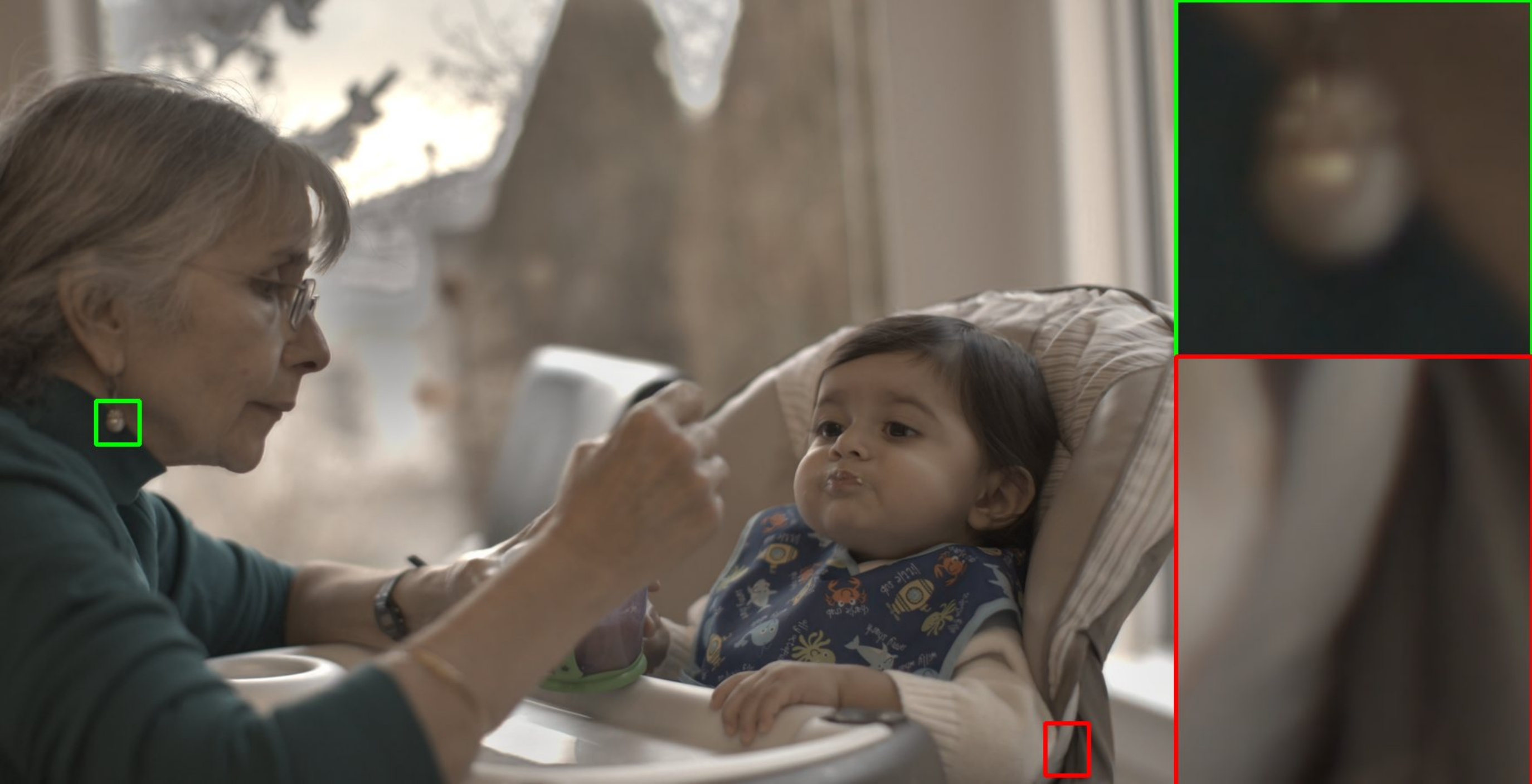}&
				\includegraphics[width=0.166\textwidth]{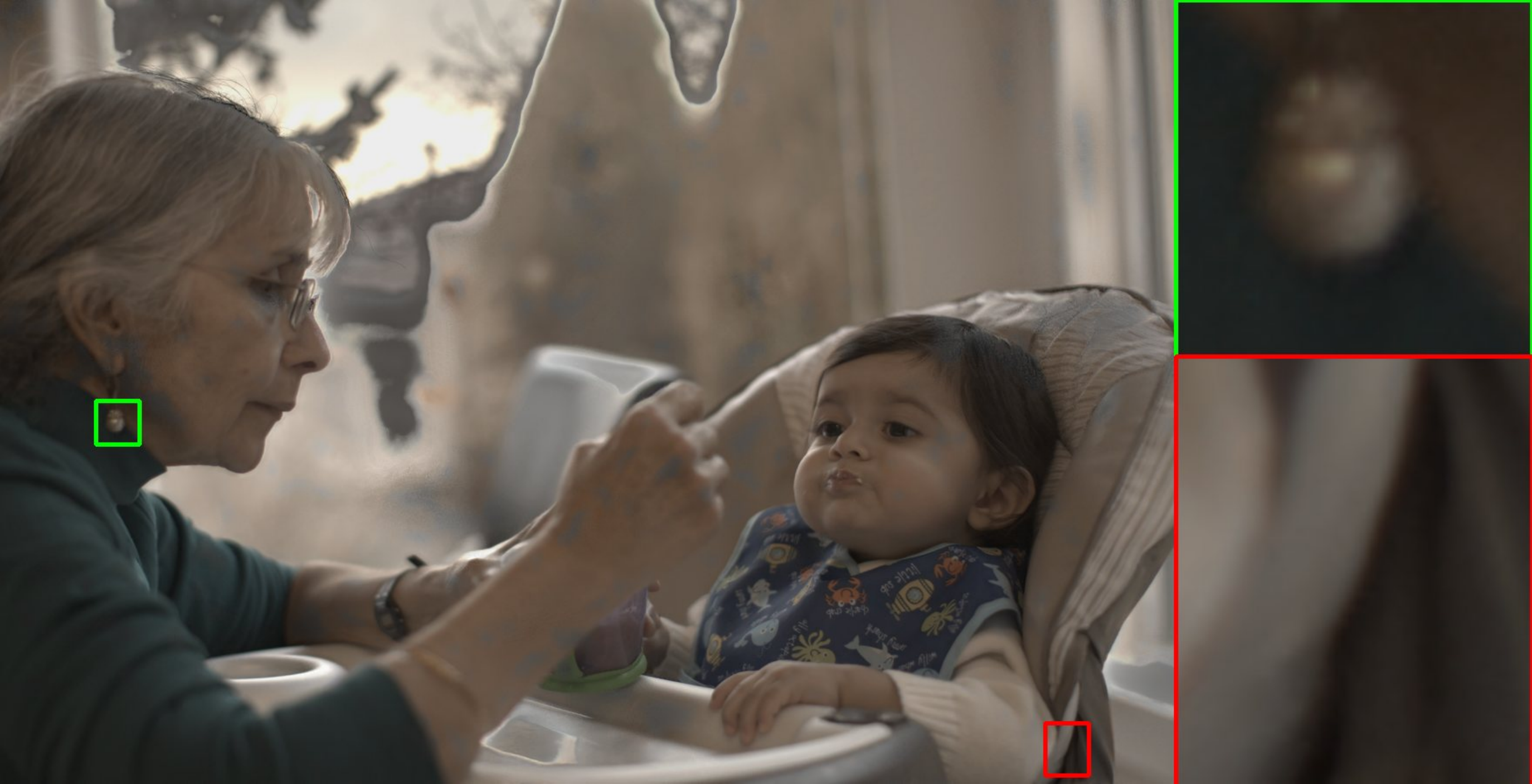}&
				\includegraphics[width=0.166\textwidth]{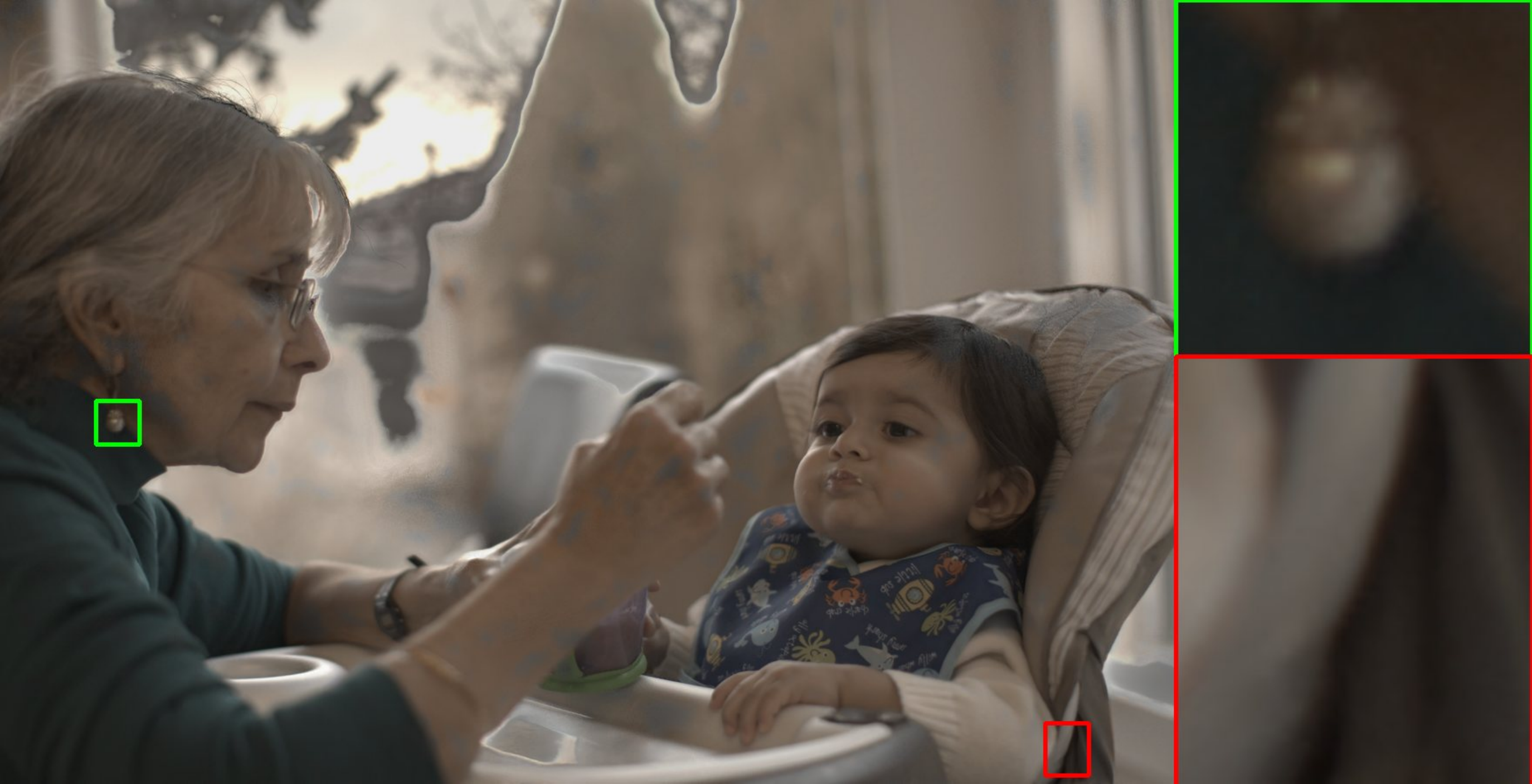}&
				\includegraphics[width=0.166\textwidth]{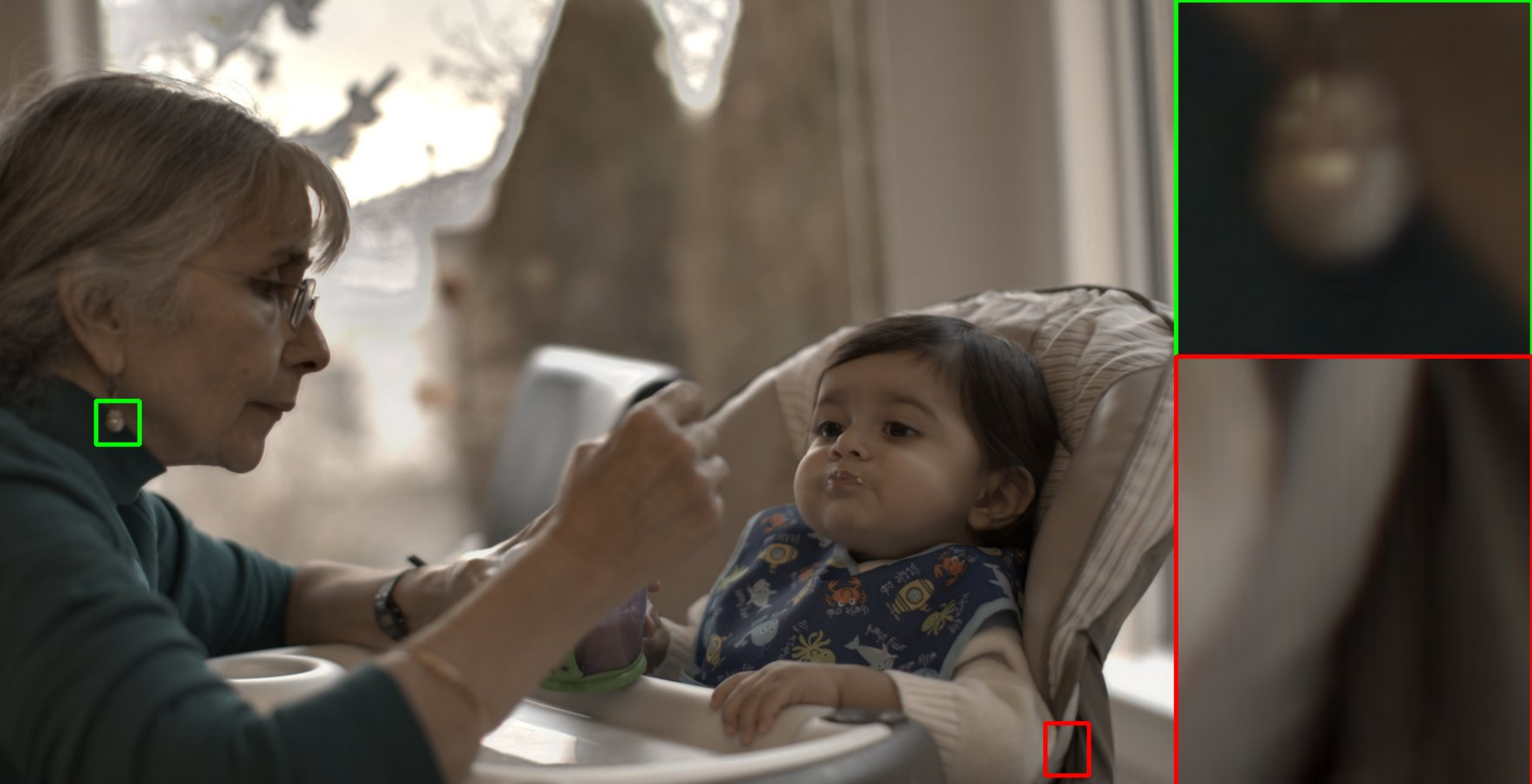}&
				\includegraphics[width=0.166\textwidth]{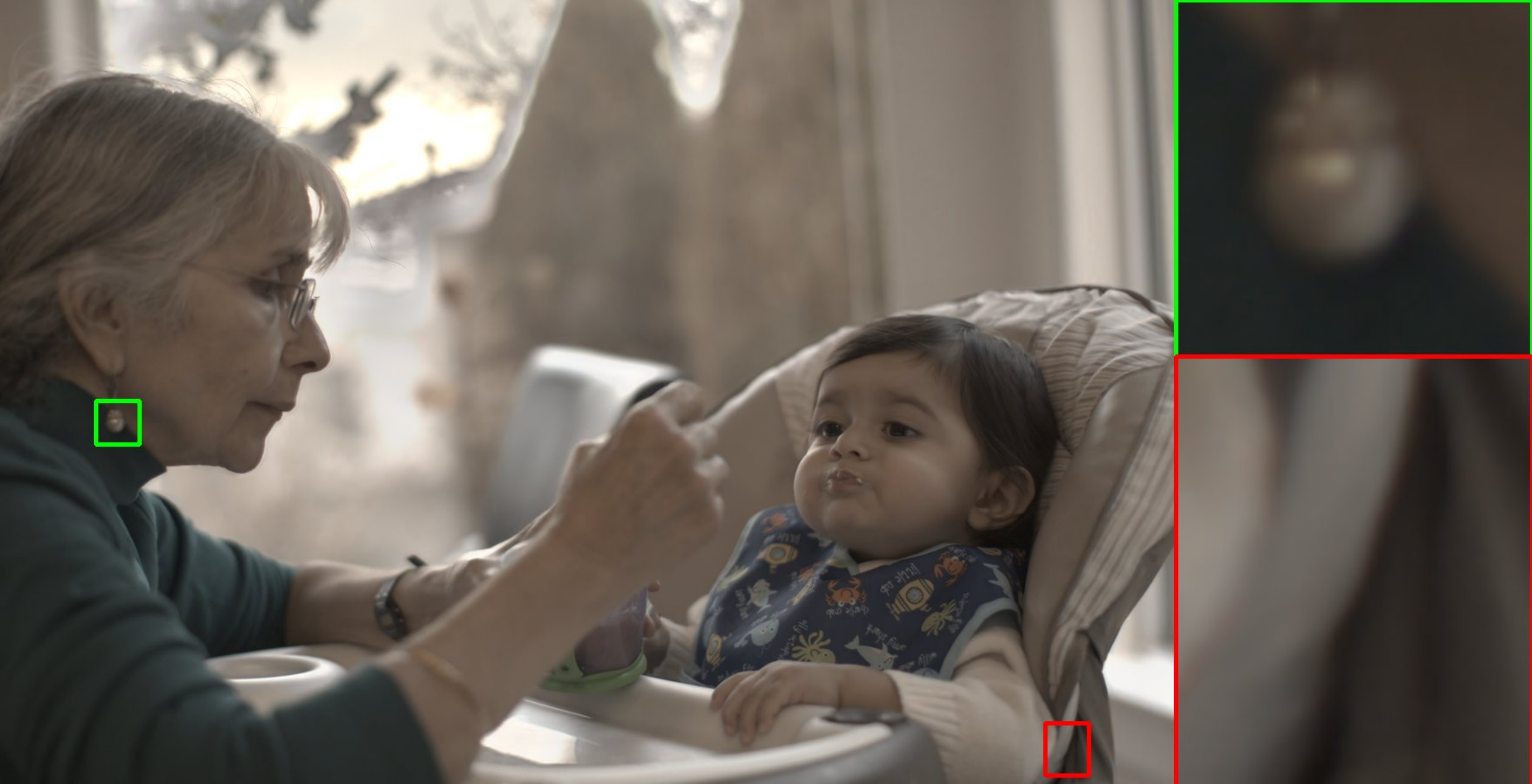}\\
				
				\includegraphics[width=0.166\textwidth]{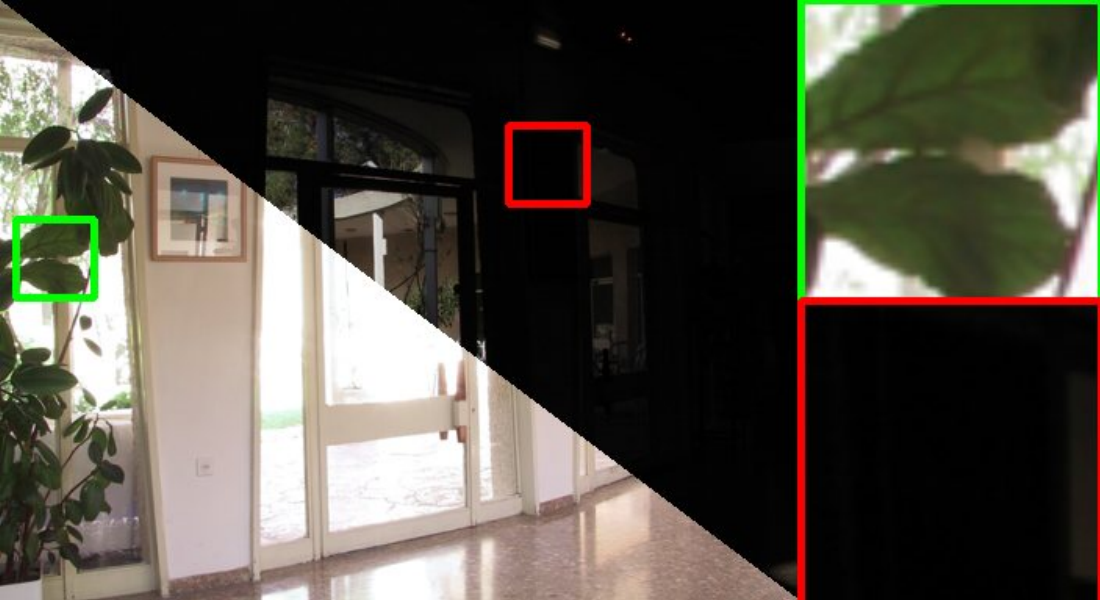}&
				\includegraphics[width=0.166\textwidth]{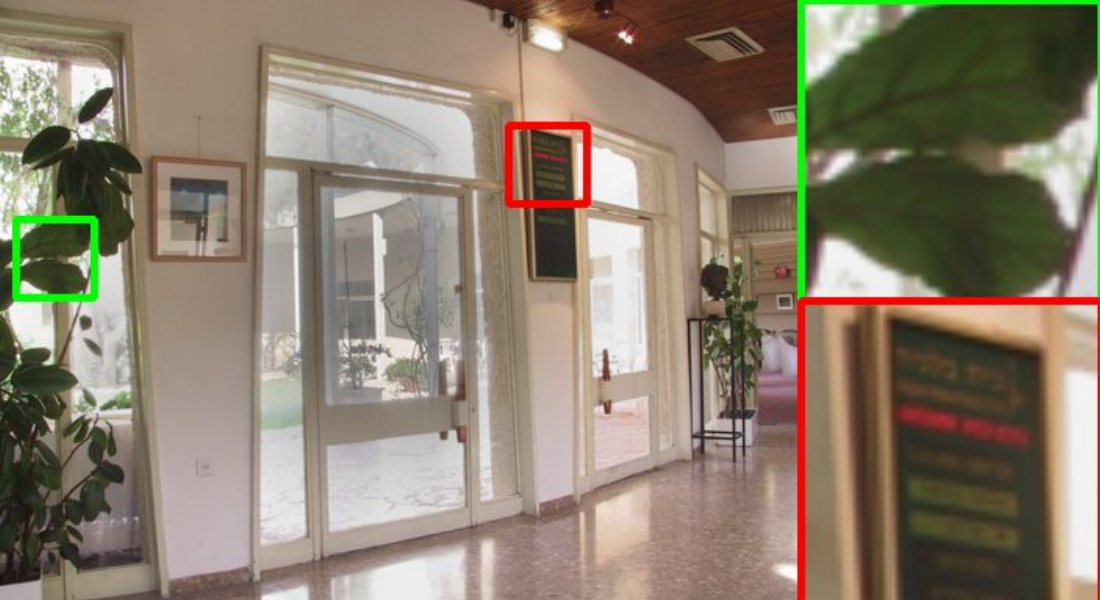}&
				\includegraphics[width=0.166\textwidth]{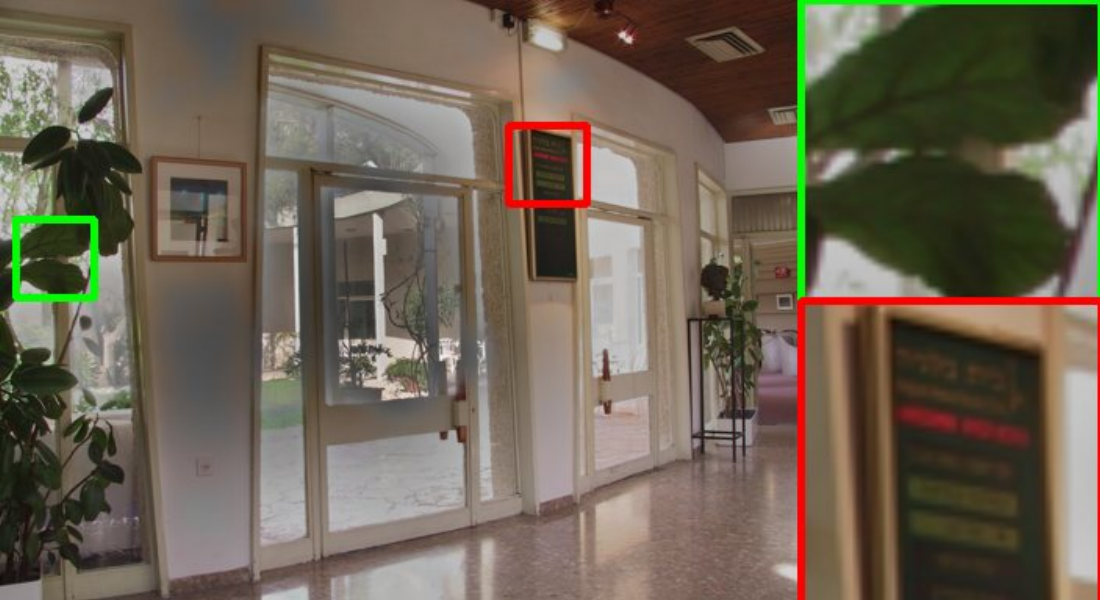}&
				\includegraphics[width=0.166\textwidth]{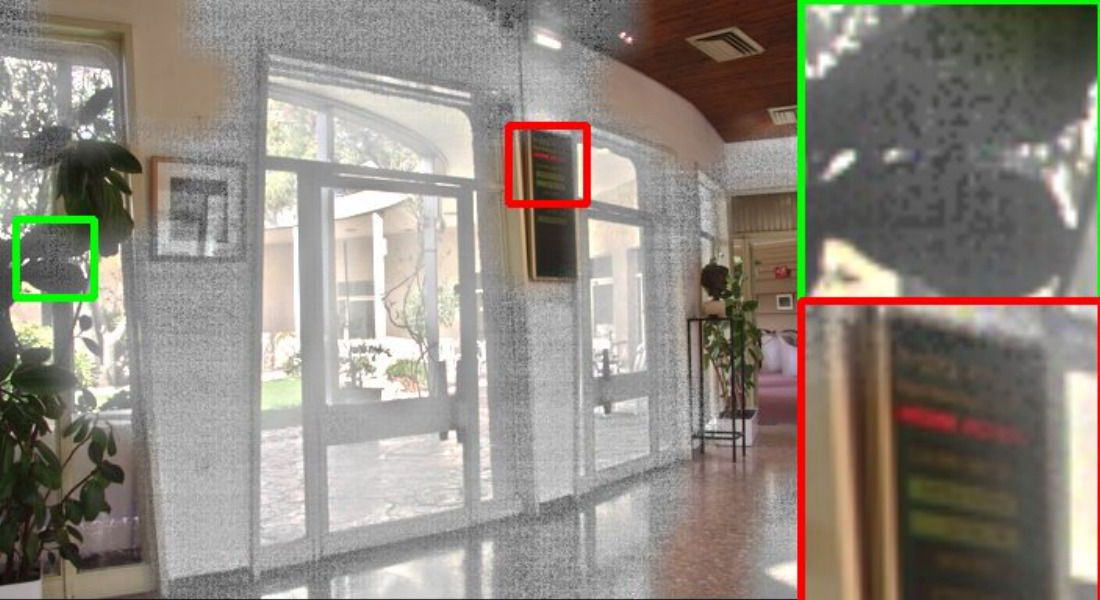}&
				\includegraphics[width=0.166\textwidth]{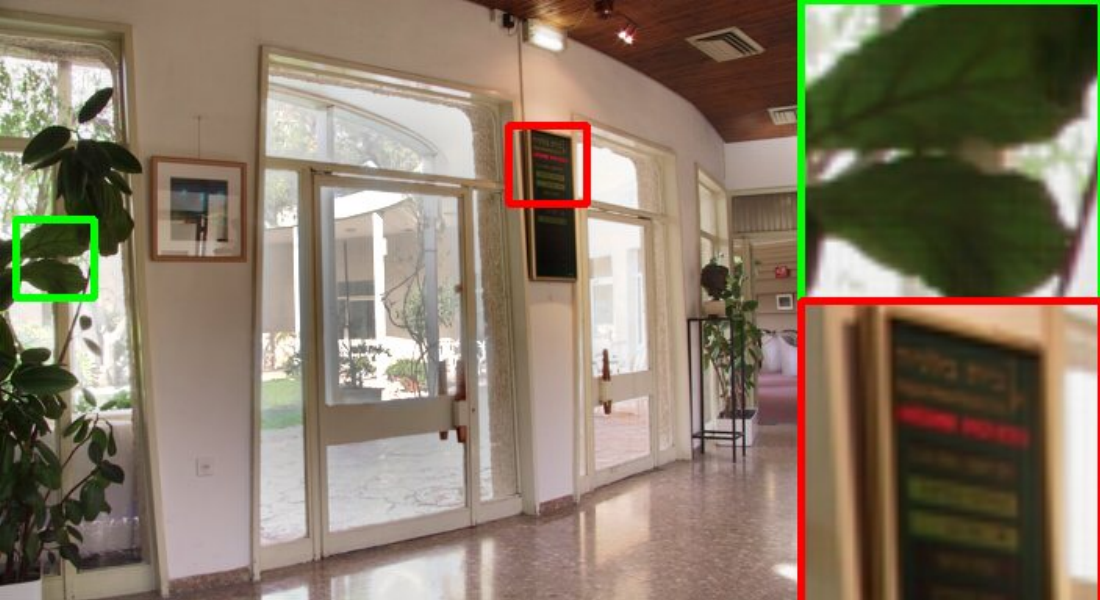}&
				\includegraphics[width=0.166\textwidth]{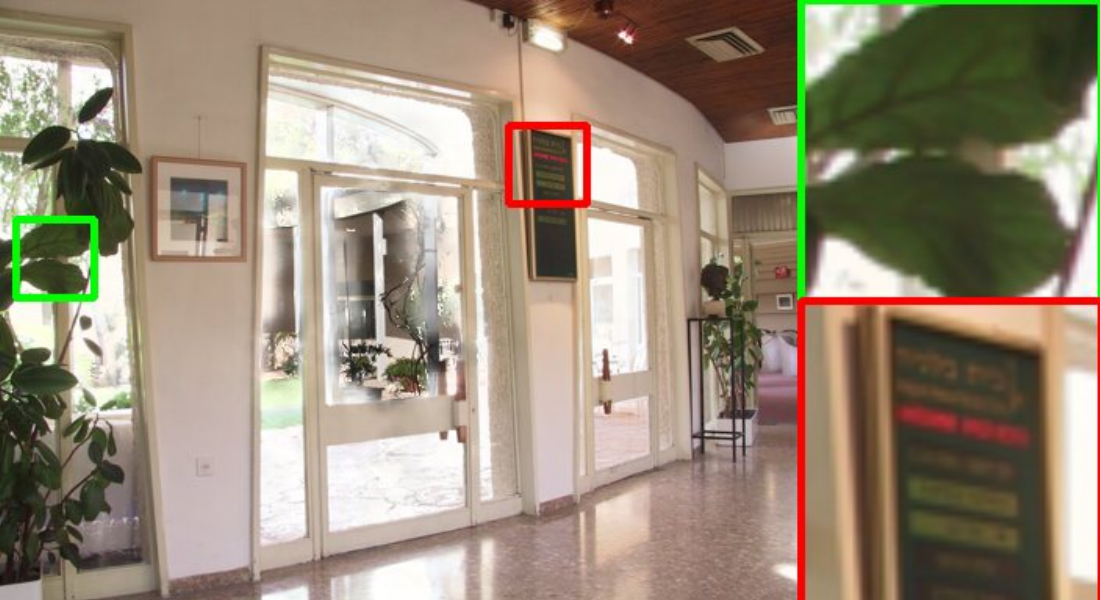}\\
				
				Inputs&  (a) DSIFT&  (b) GFF&  (c) SPDMEF&  (d) FMMEF&  (e) MEFNet\\
				
				\includegraphics[width=0.166\textwidth]{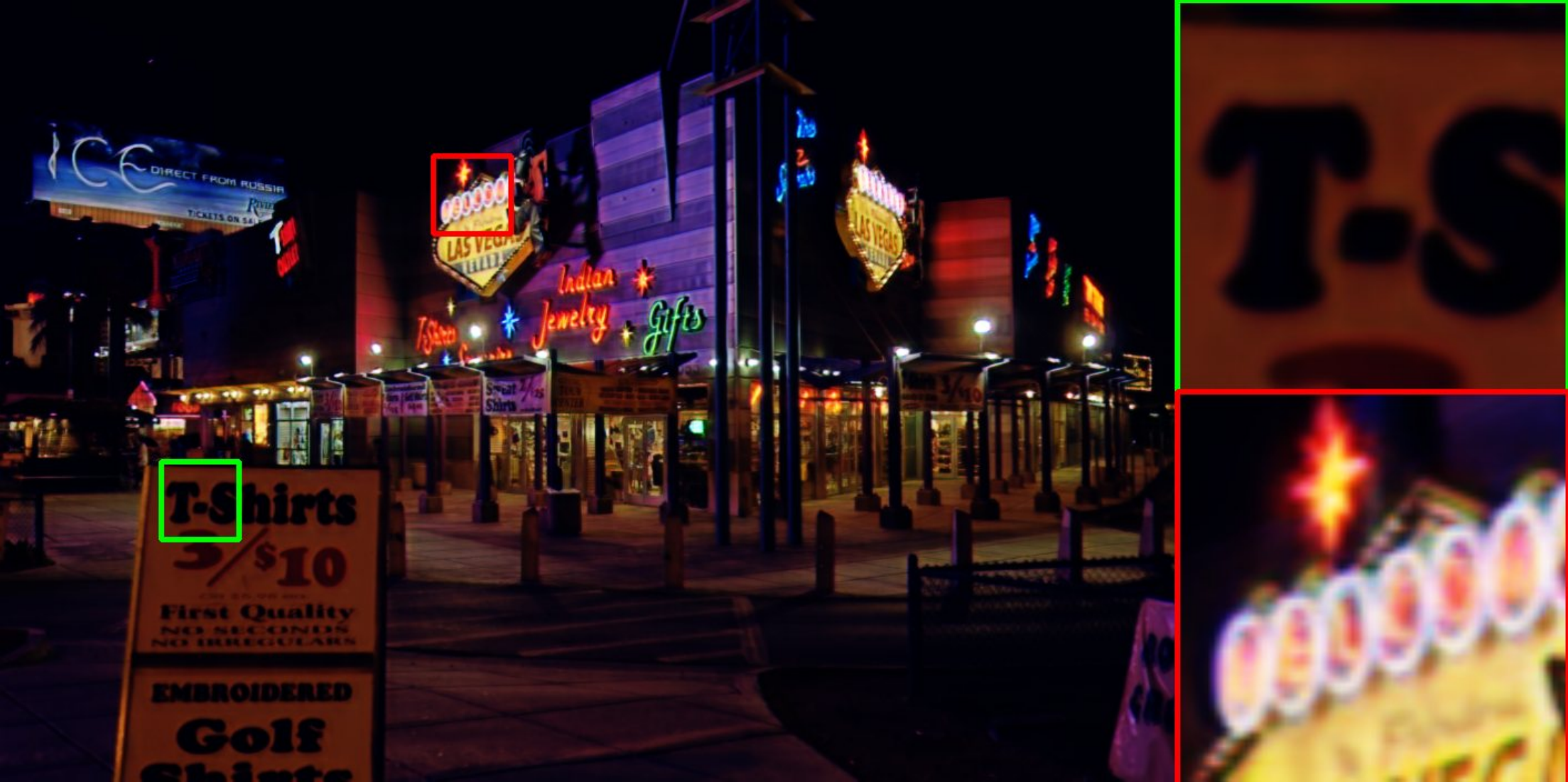}&
				\includegraphics[width=0.166\textwidth]{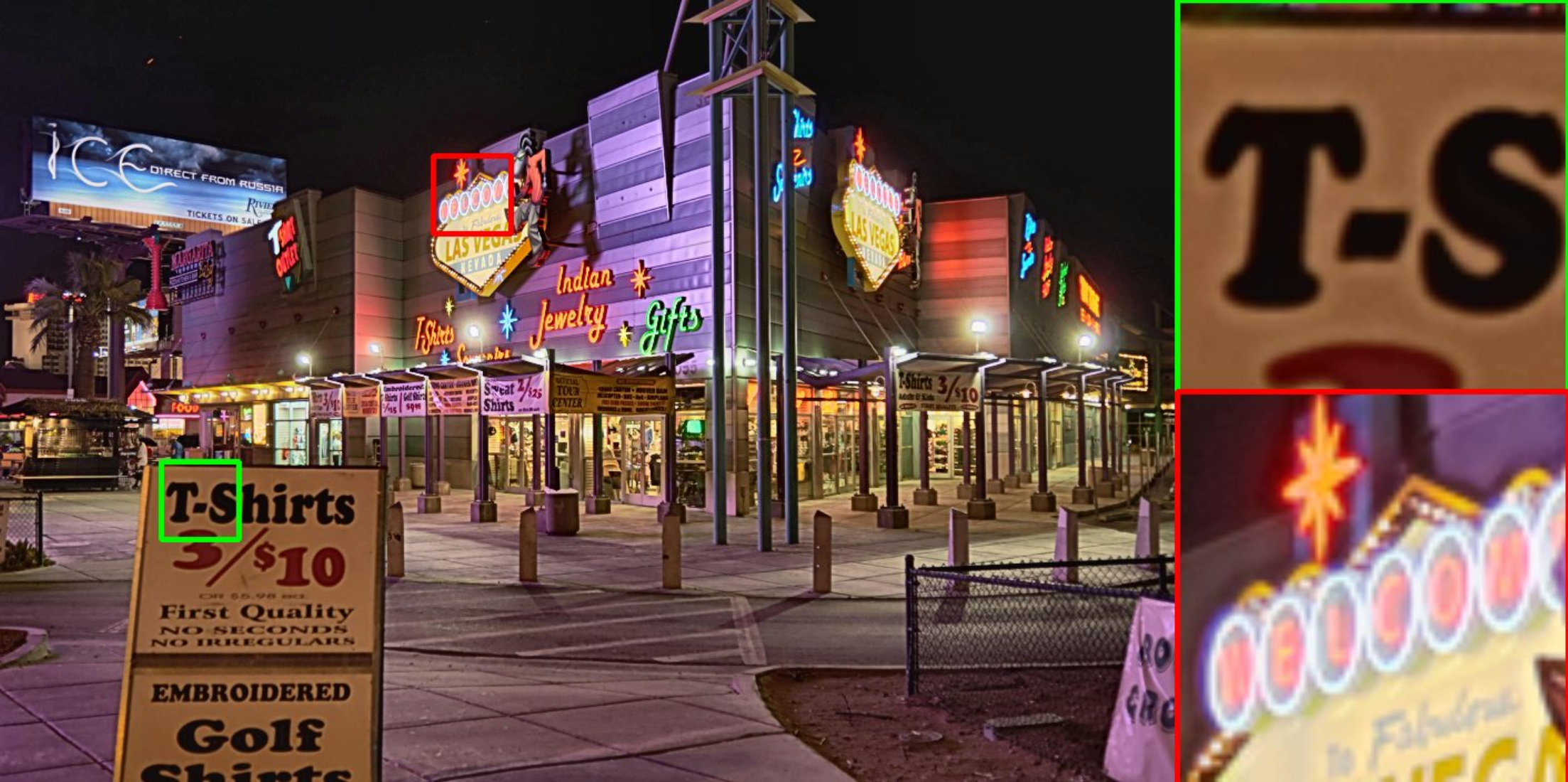}&
				\includegraphics[width=0.166\textwidth]{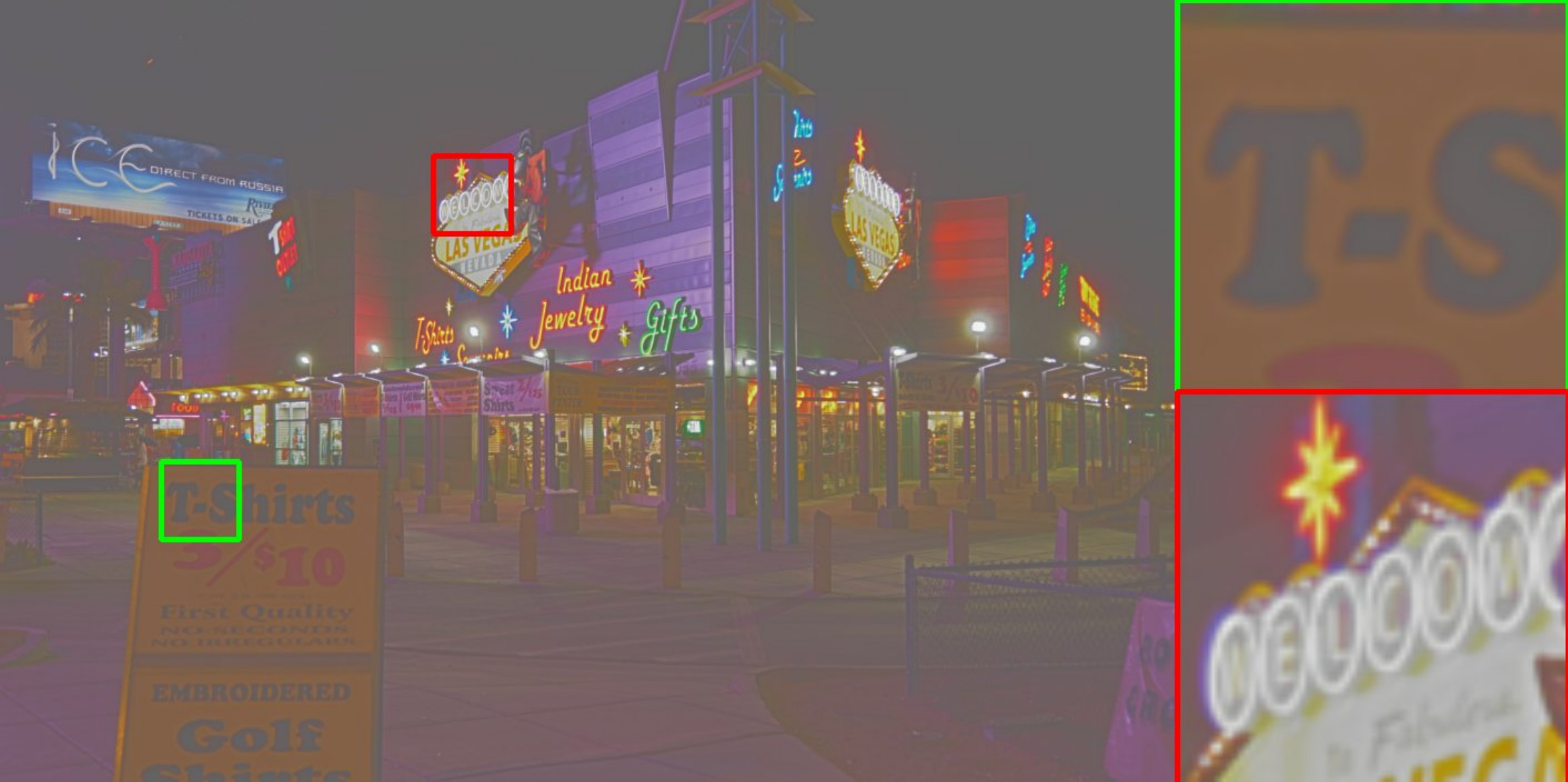}&
				\includegraphics[width=0.166\textwidth]{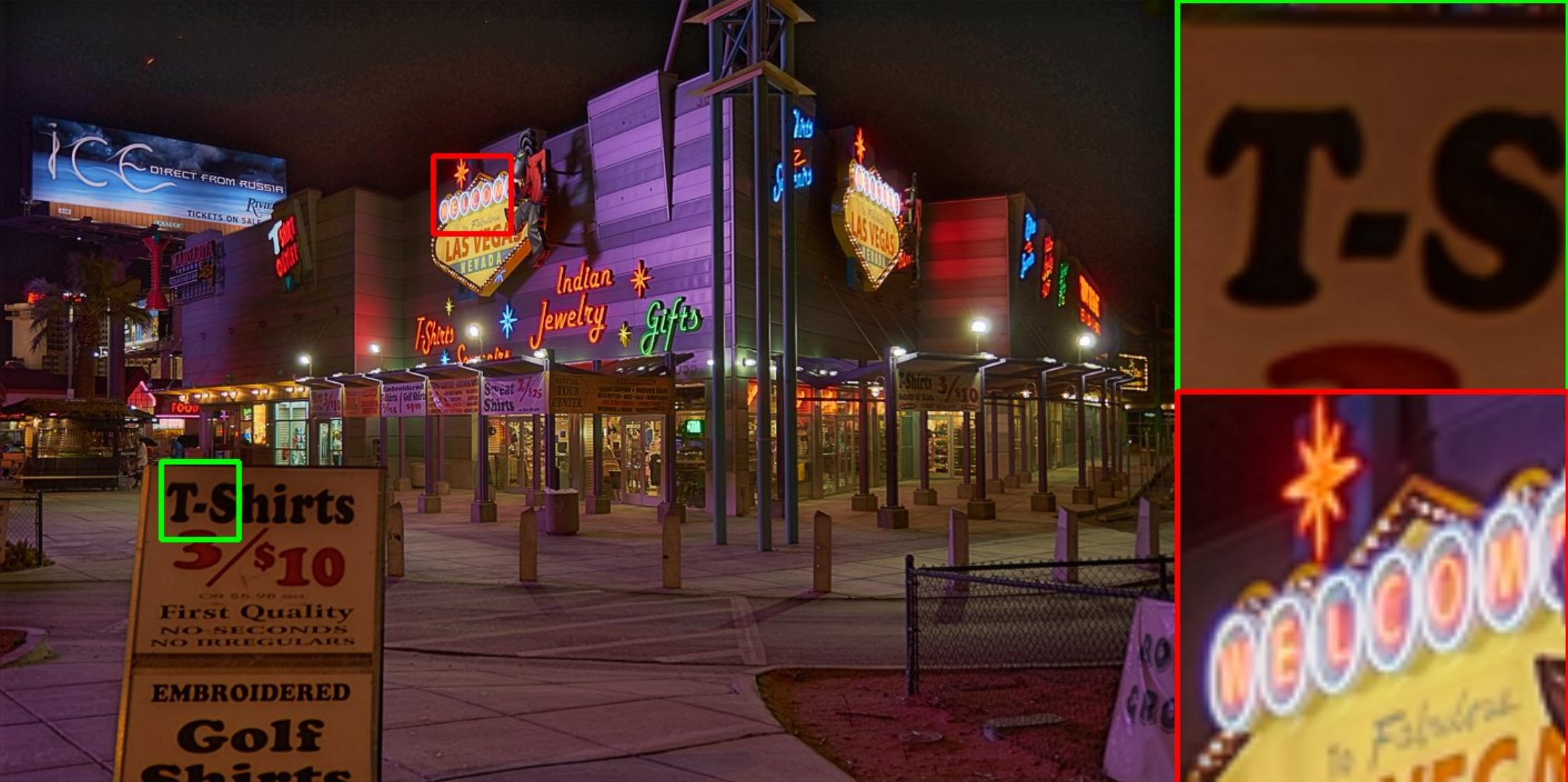}&
				\includegraphics[width=0.166\textwidth]{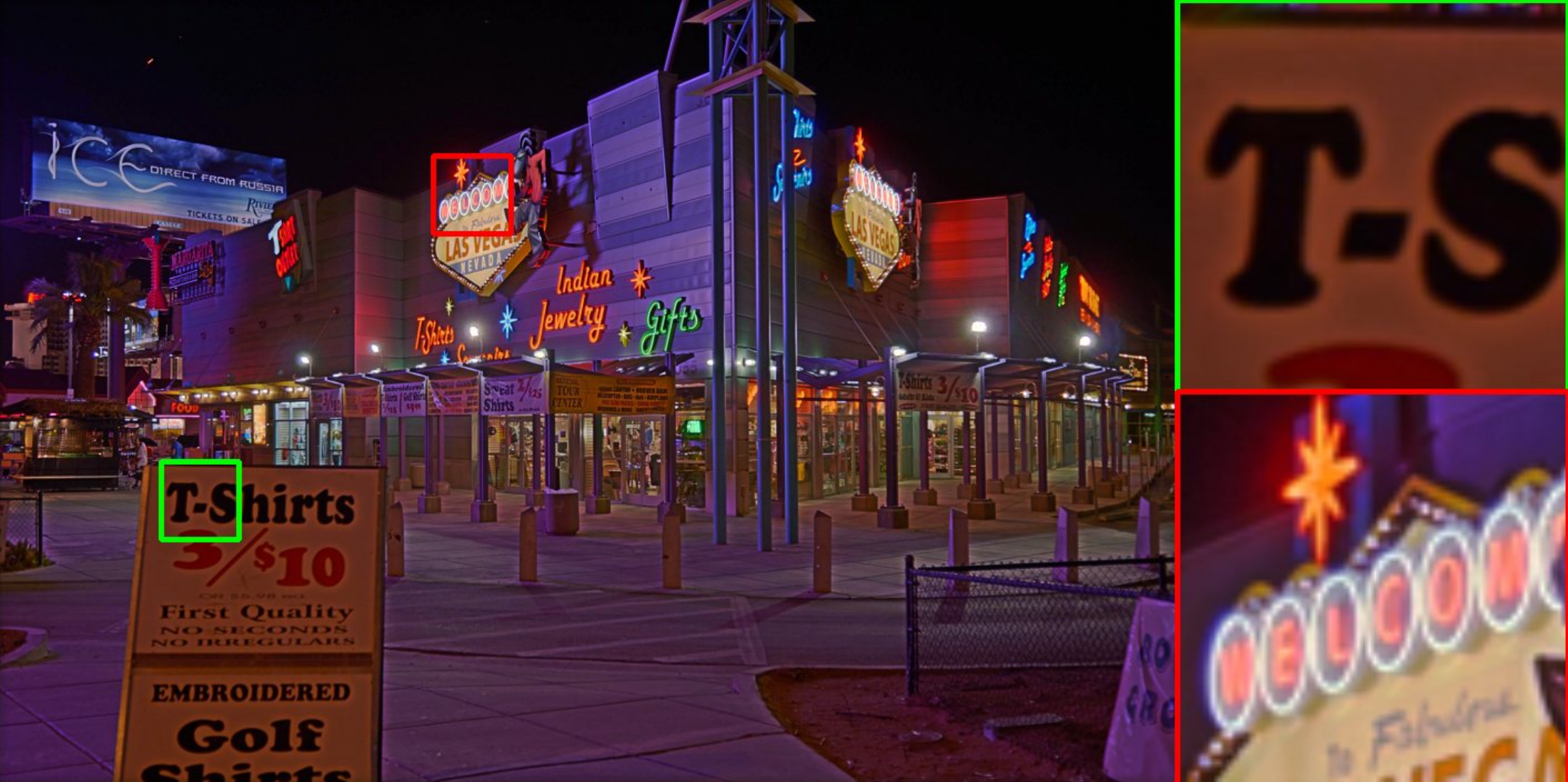}&
				\includegraphics[width=0.166\textwidth]{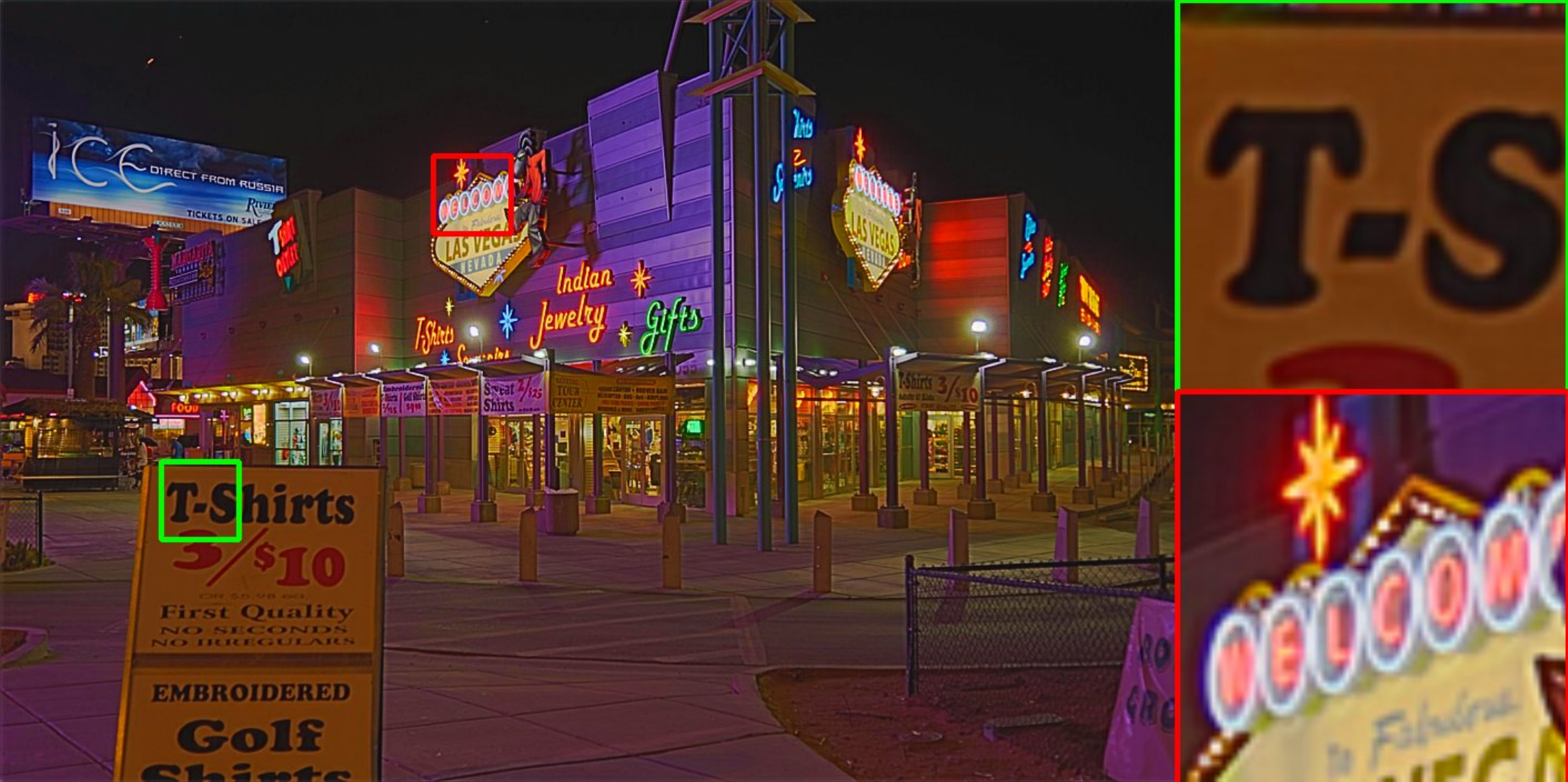}\\
				
				\includegraphics[width=0.166\textwidth]{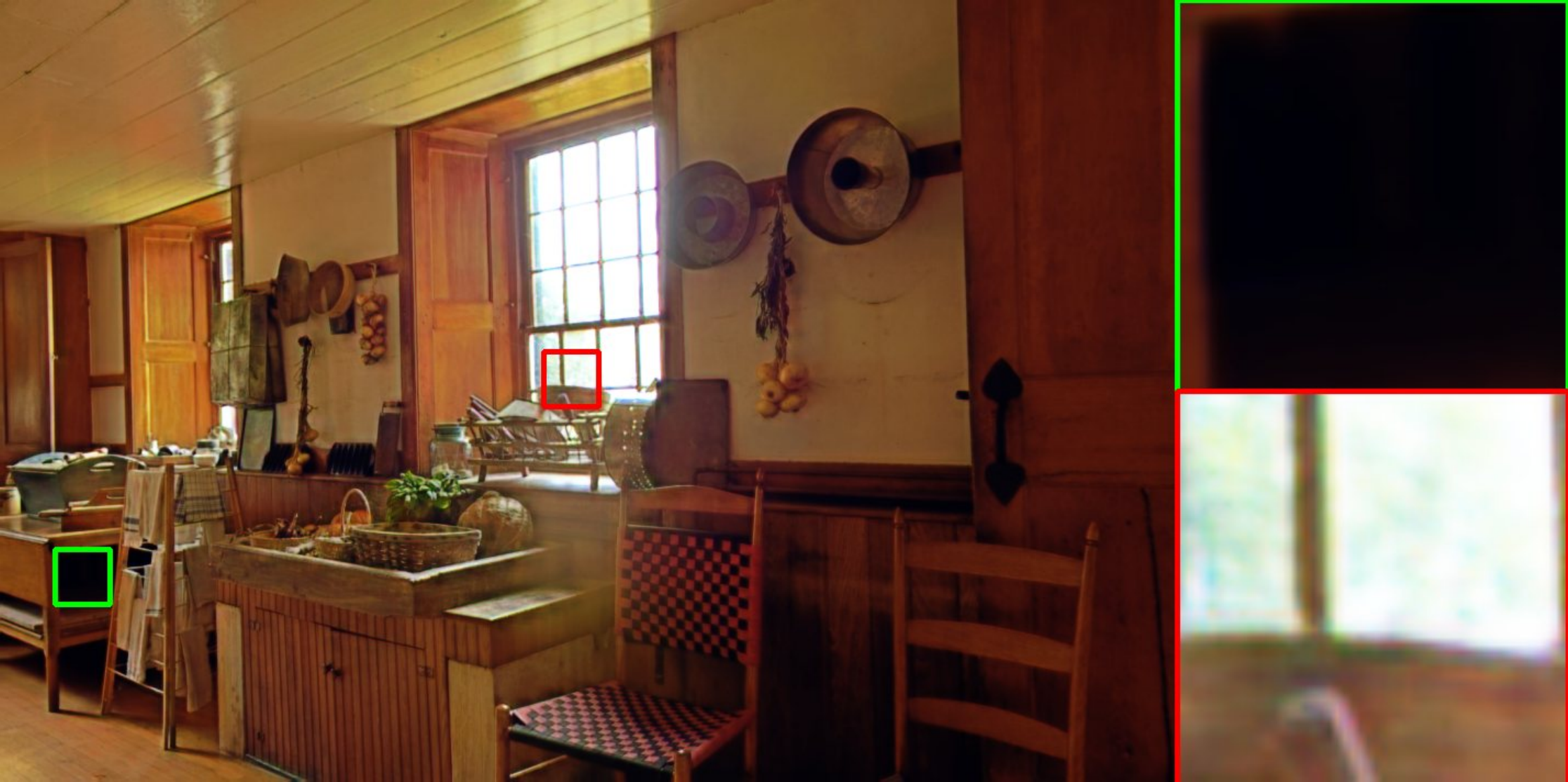}&
				\includegraphics[width=0.166\textwidth]{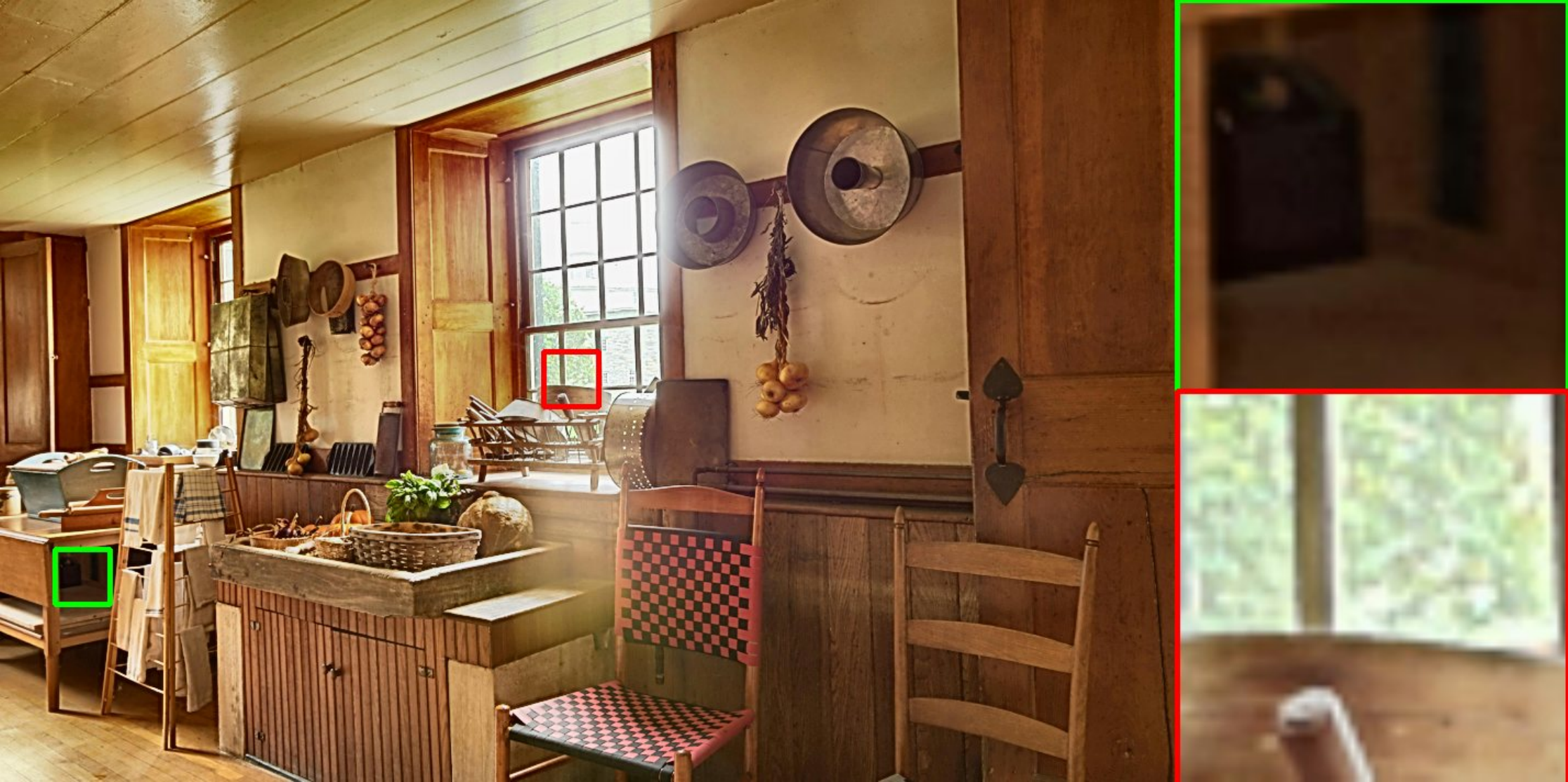}&
				\includegraphics[width=0.166\textwidth]{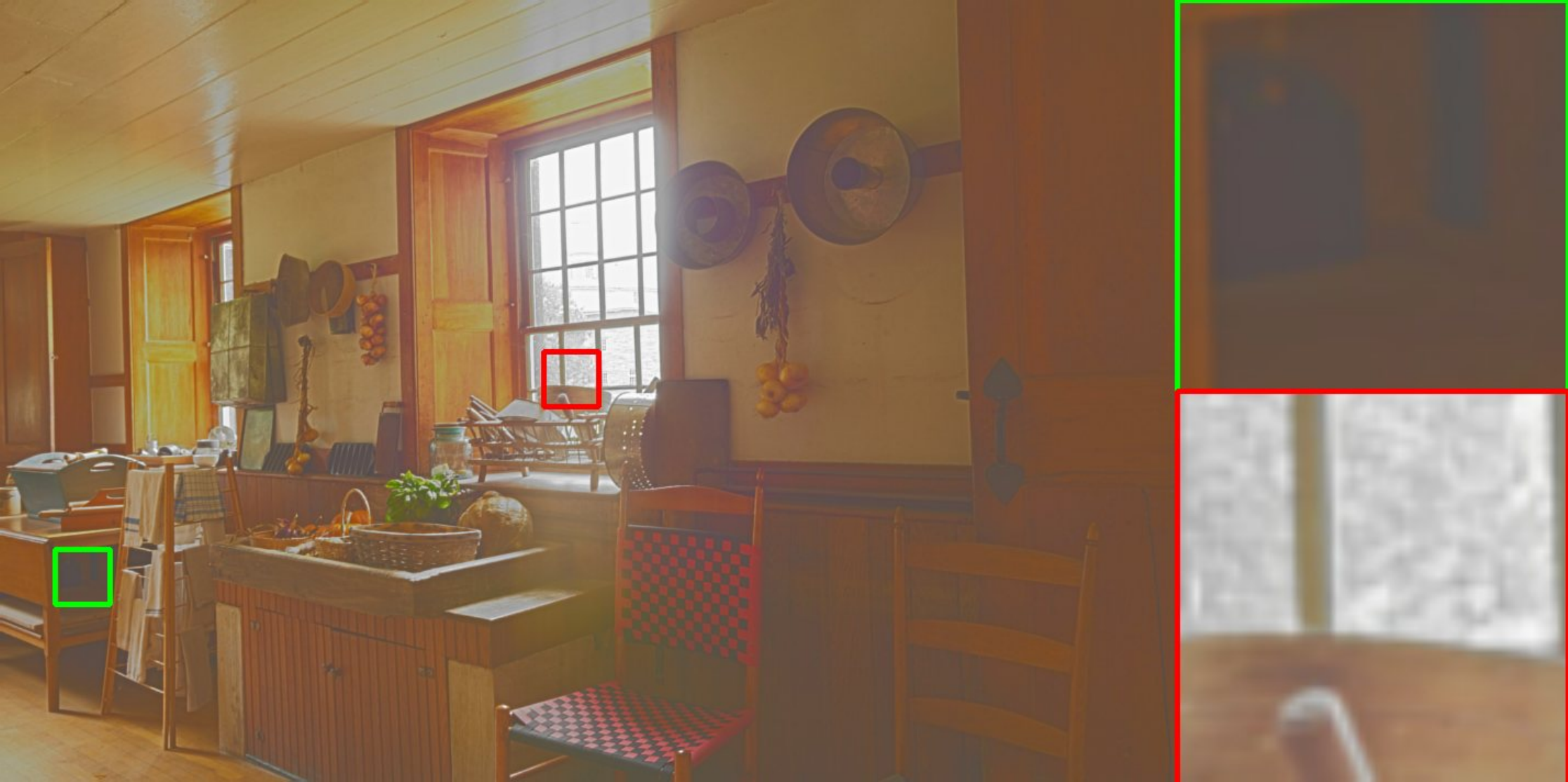}&
				\includegraphics[width=0.166\textwidth]{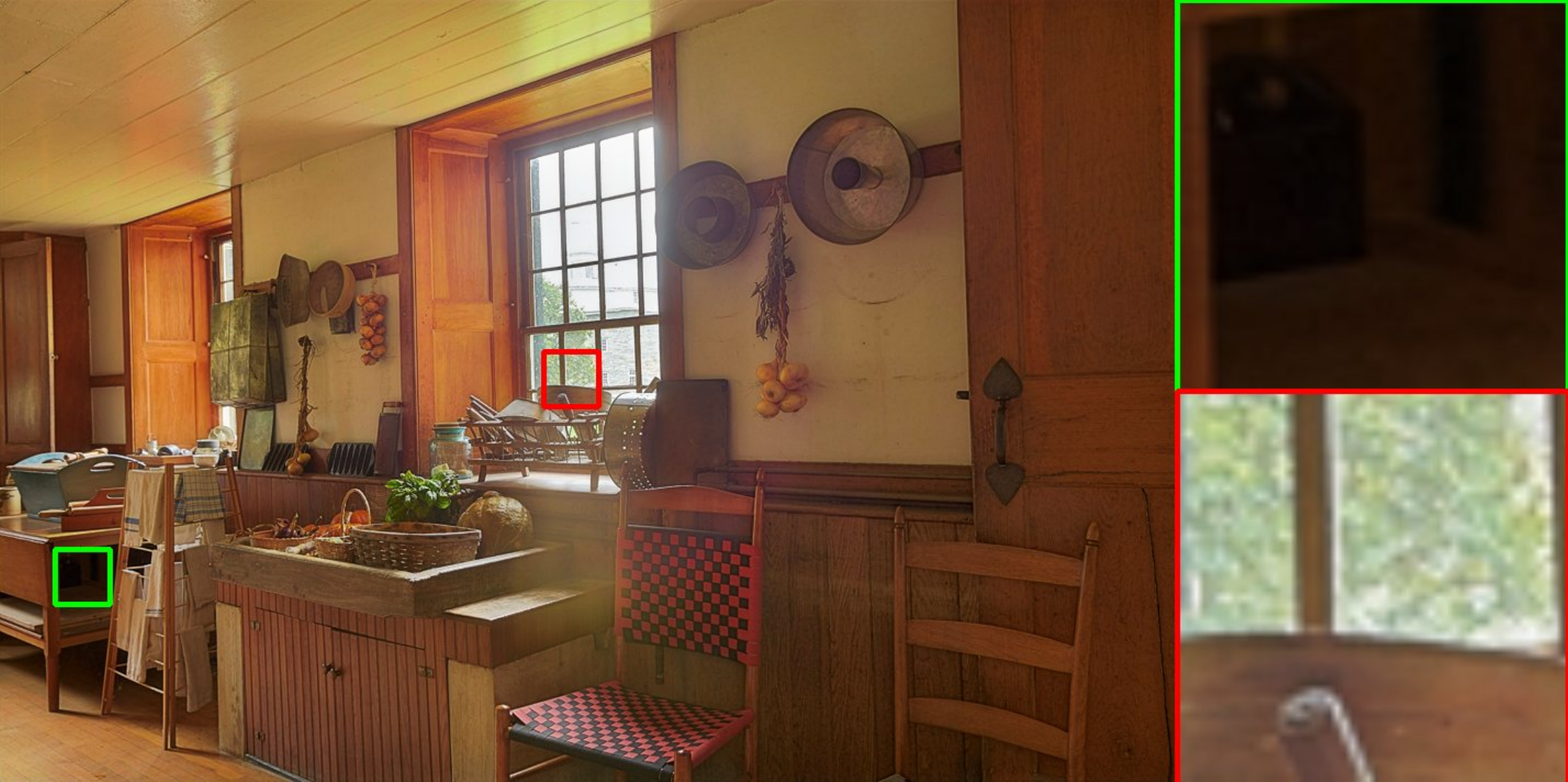}&
				\includegraphics[width=0.166\textwidth]{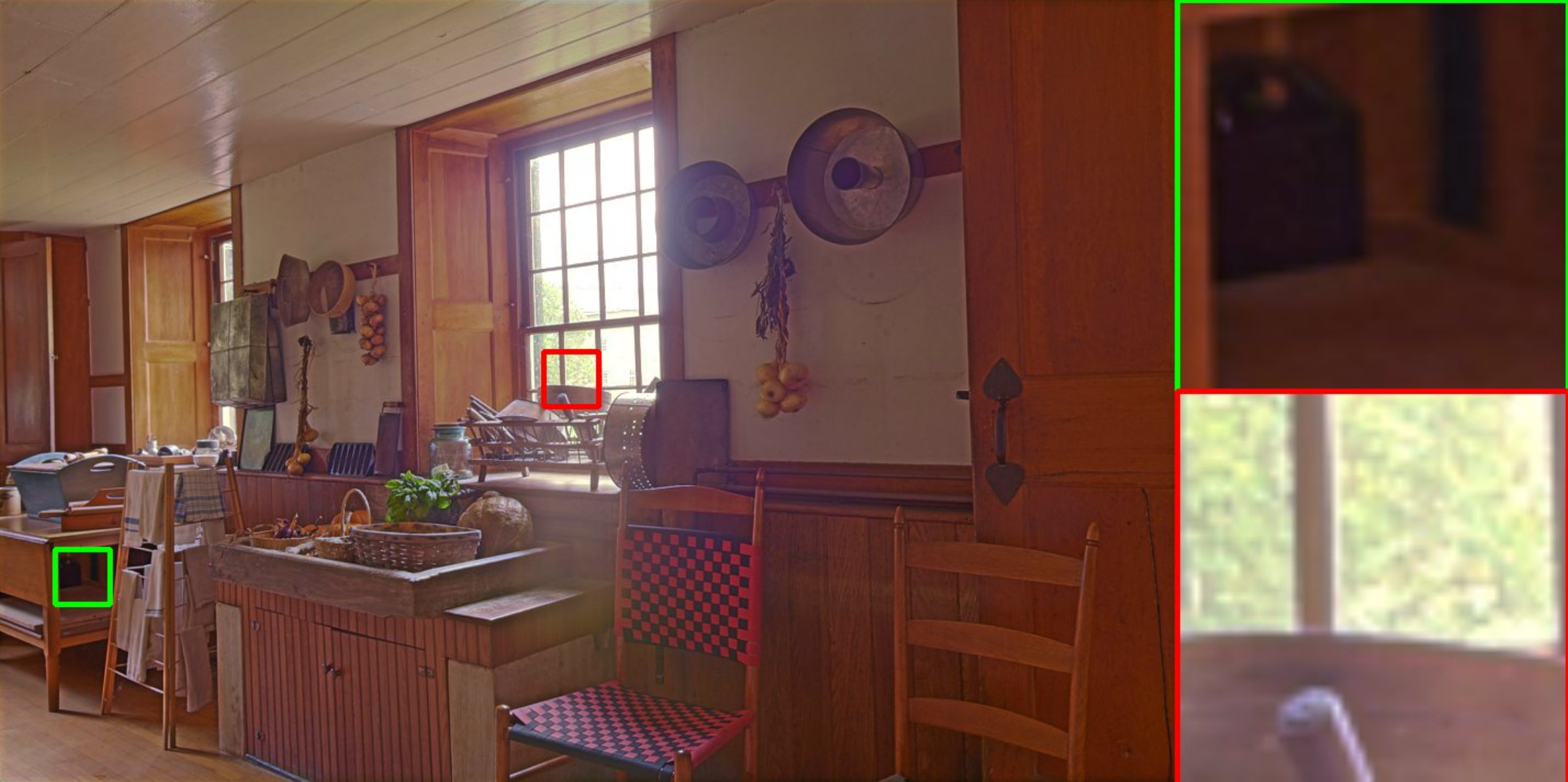}&
				\includegraphics[width=0.166\textwidth]{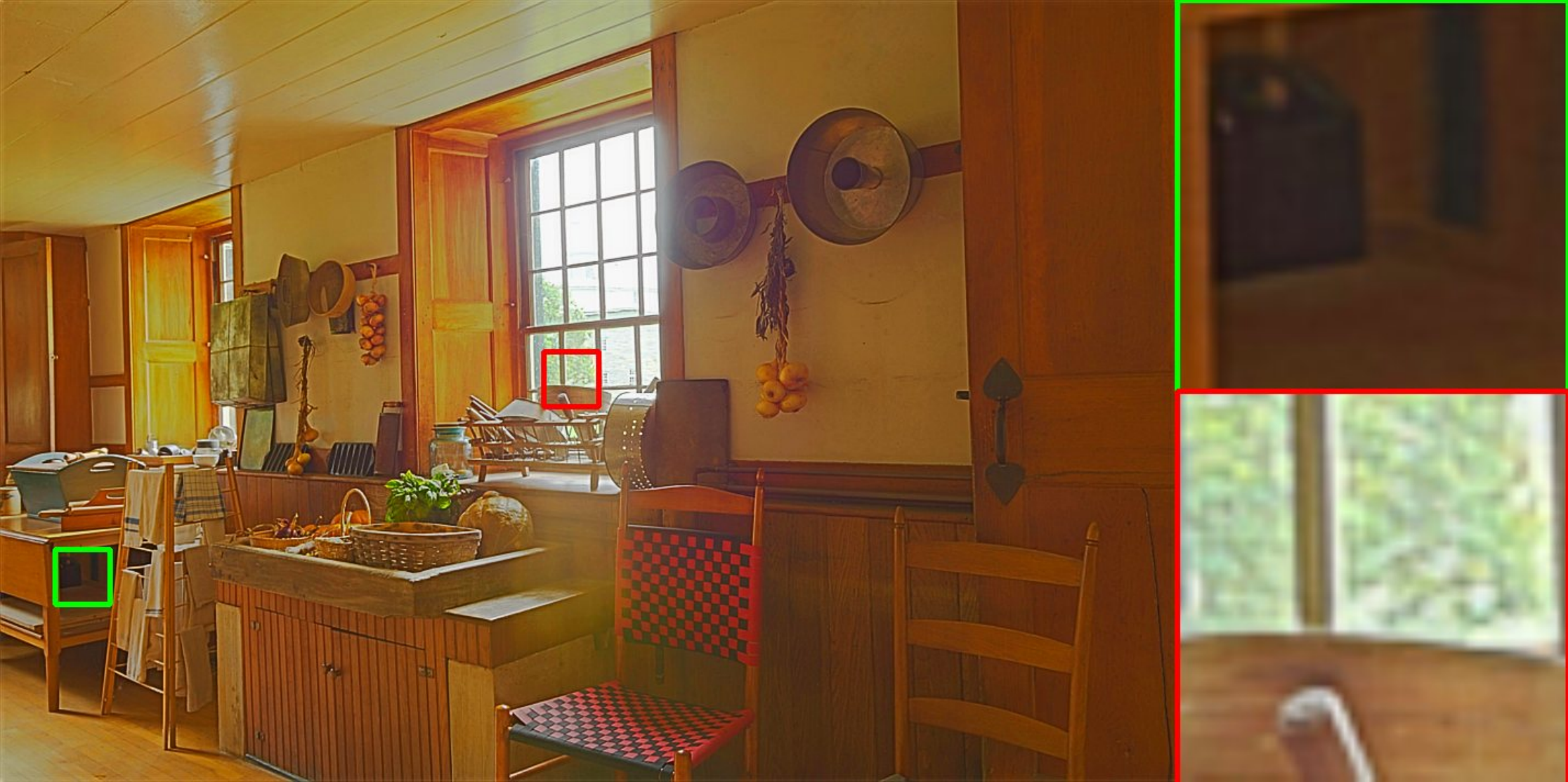}\\
				
				\includegraphics[width=0.166\textwidth]{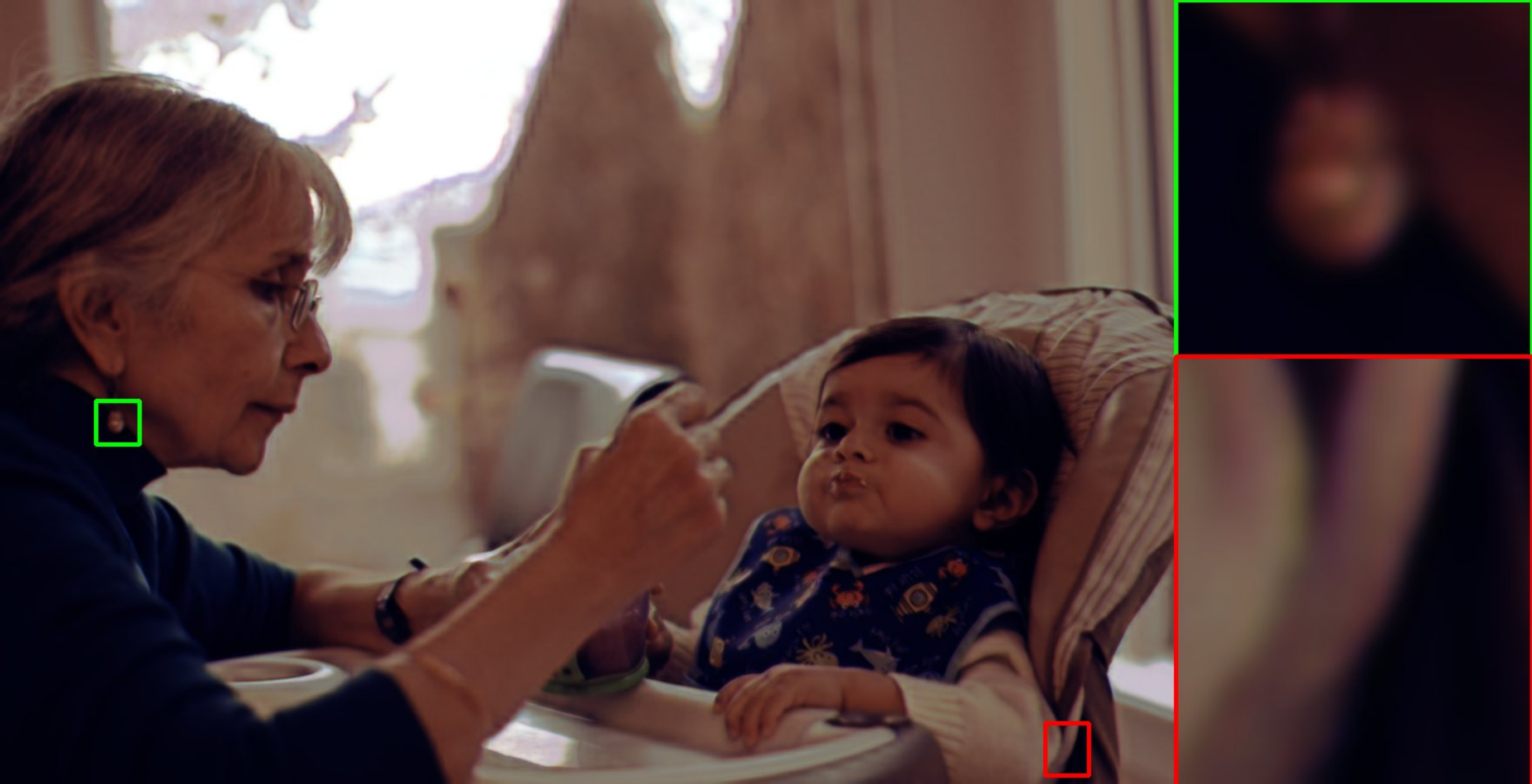}&
				\includegraphics[width=0.166\textwidth]{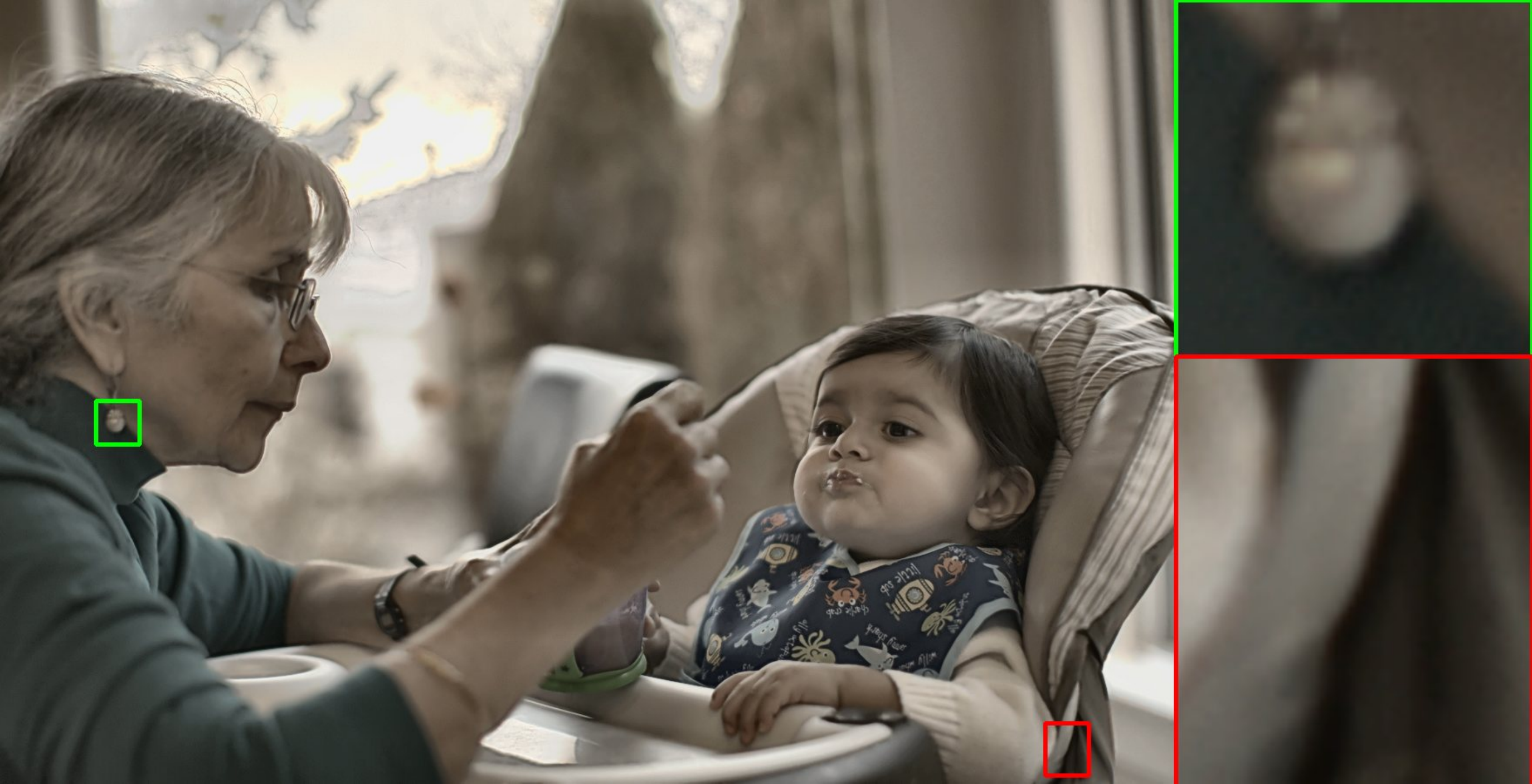}&
				\includegraphics[width=0.166\textwidth]{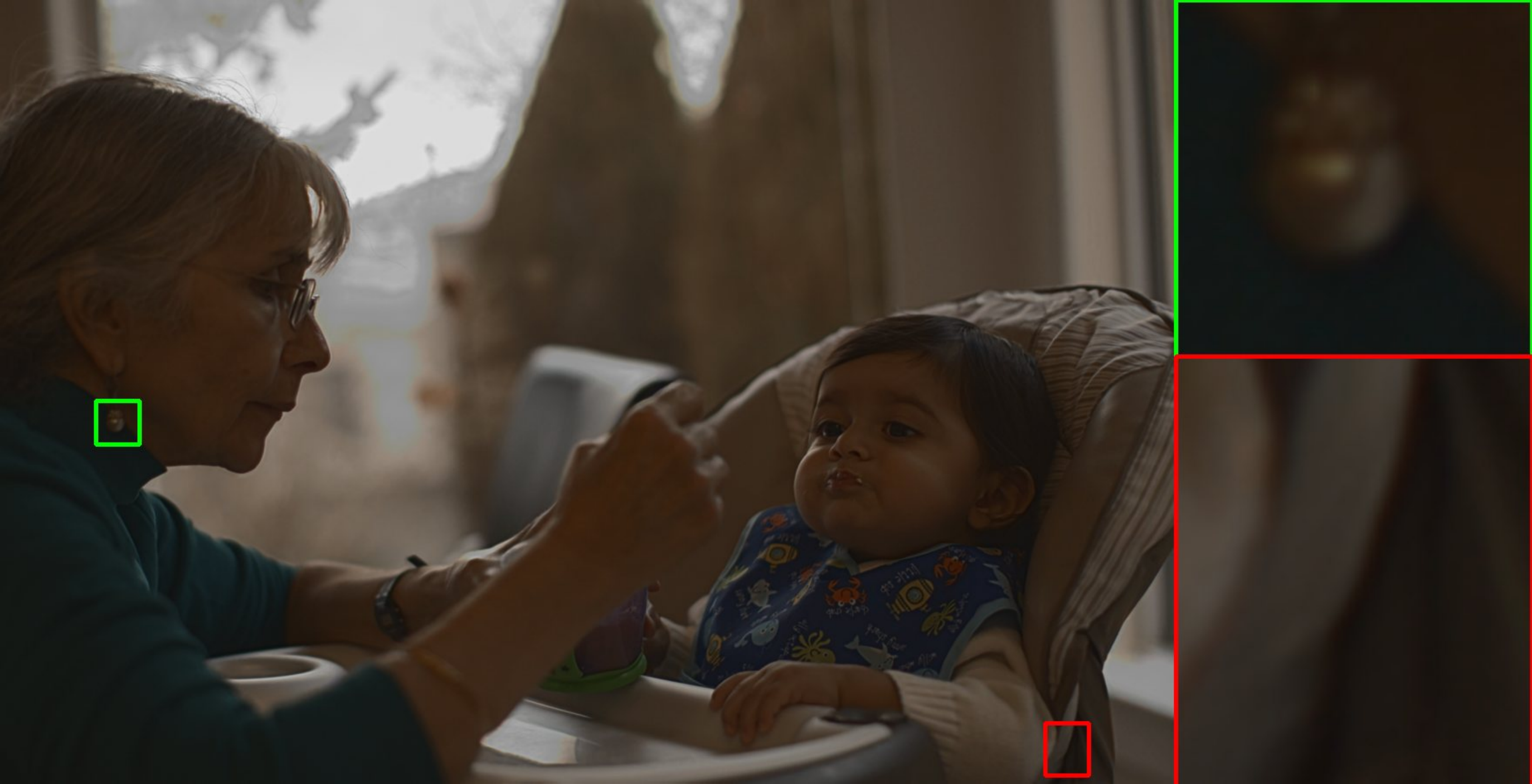}&
				\includegraphics[width=0.166\textwidth]{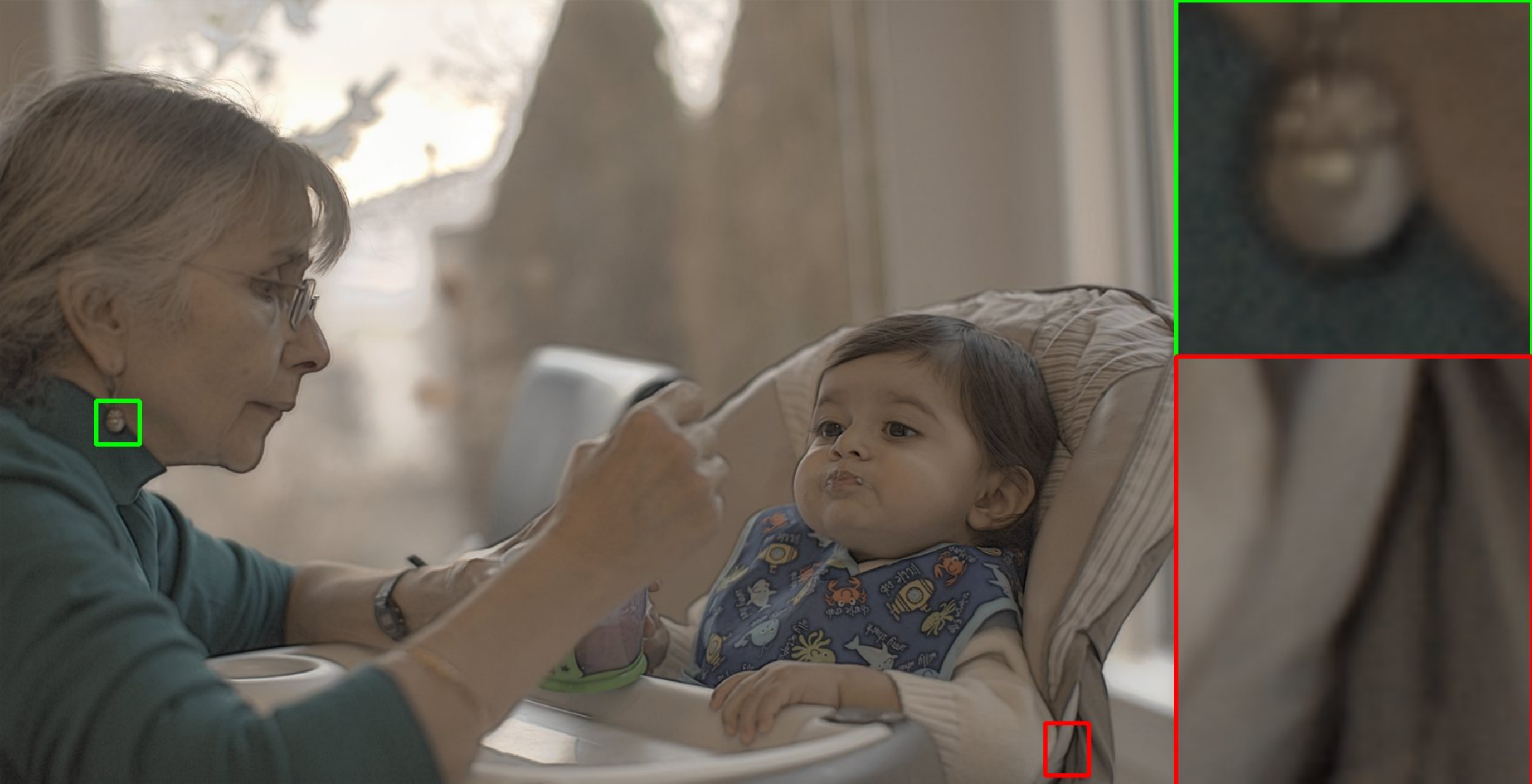}&
				\includegraphics[width=0.166\textwidth]{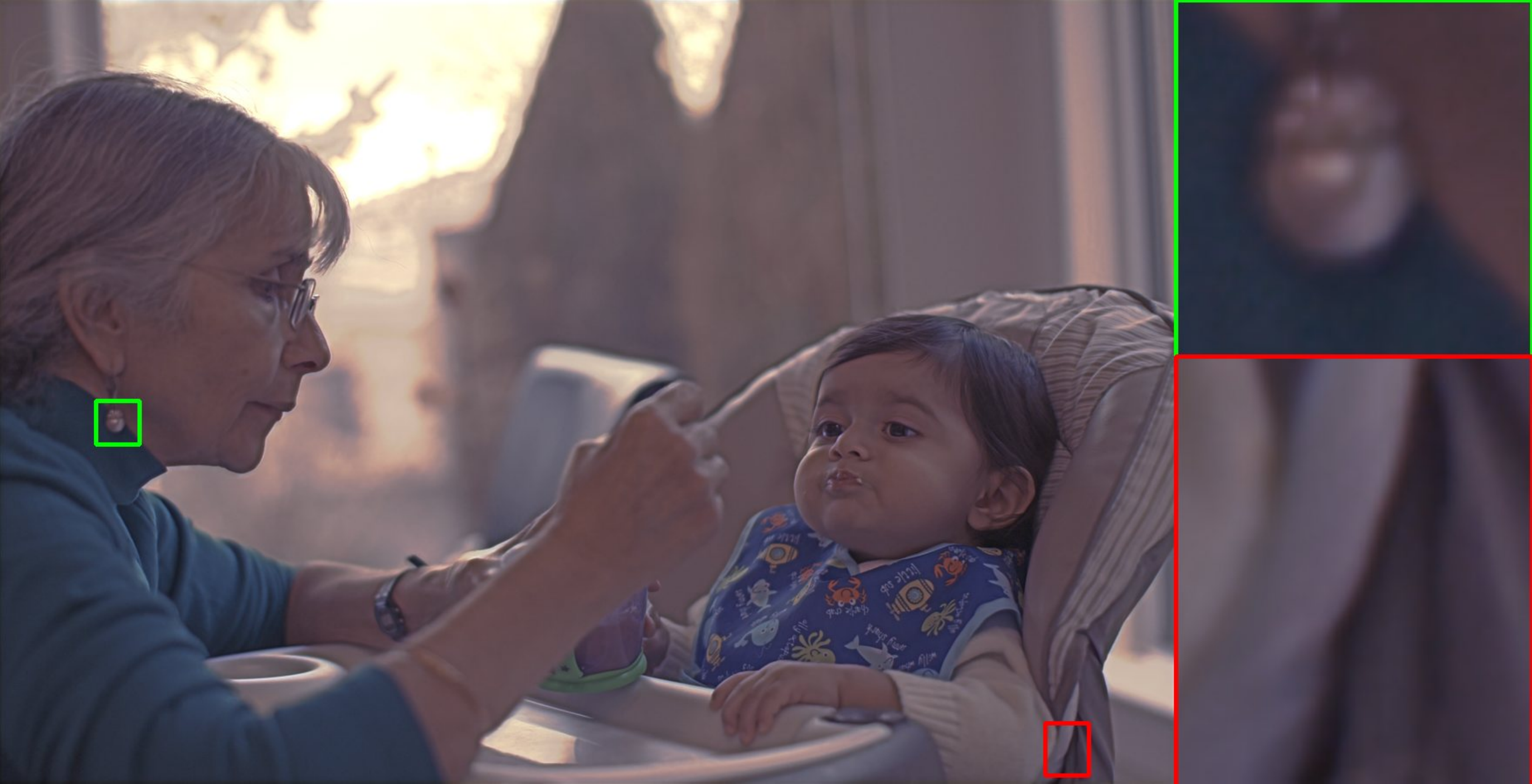}&
				\includegraphics[width=0.166\textwidth]{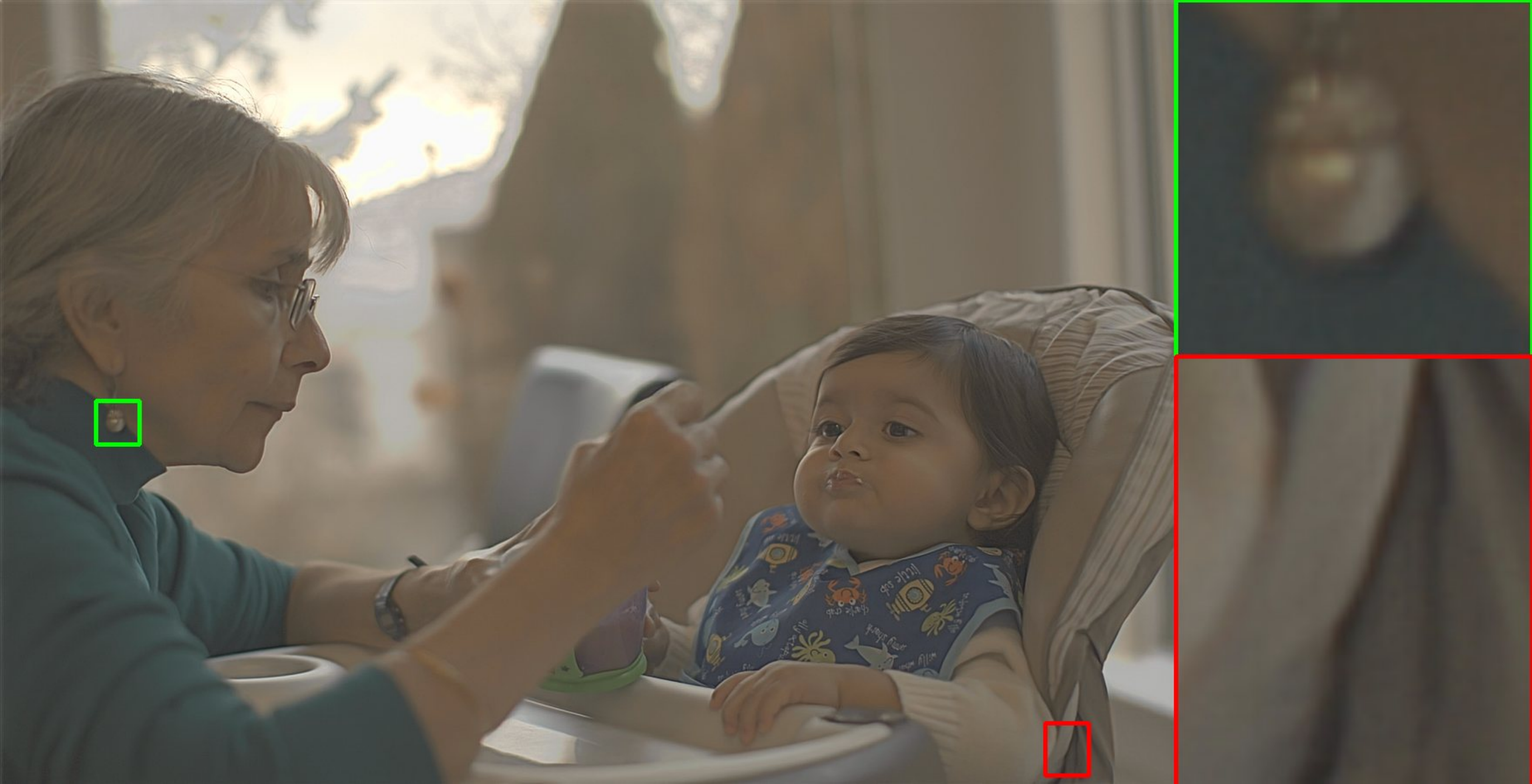}\\
				
				\includegraphics[width=0.166\textwidth]{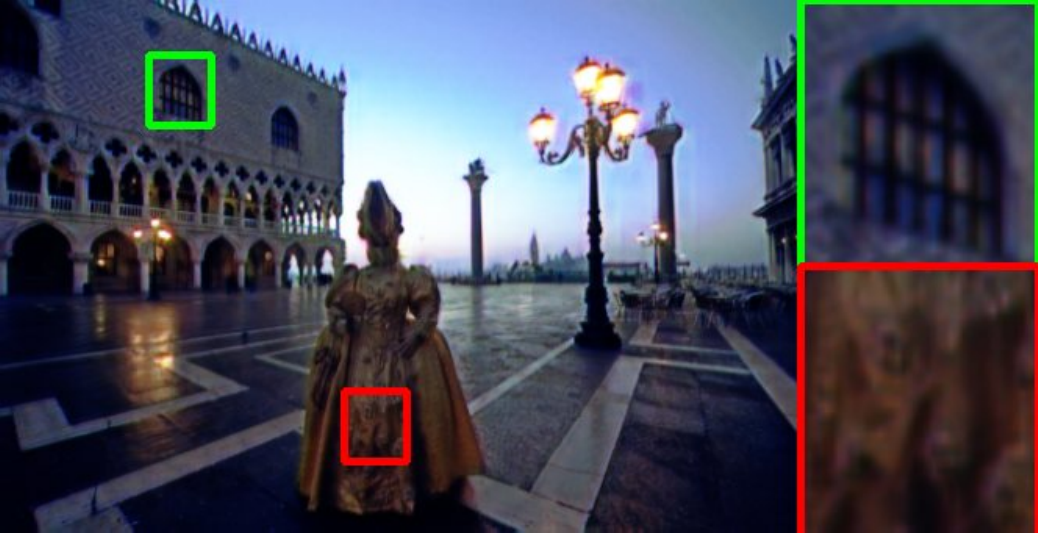}&
				\includegraphics[width=0.166\textwidth]{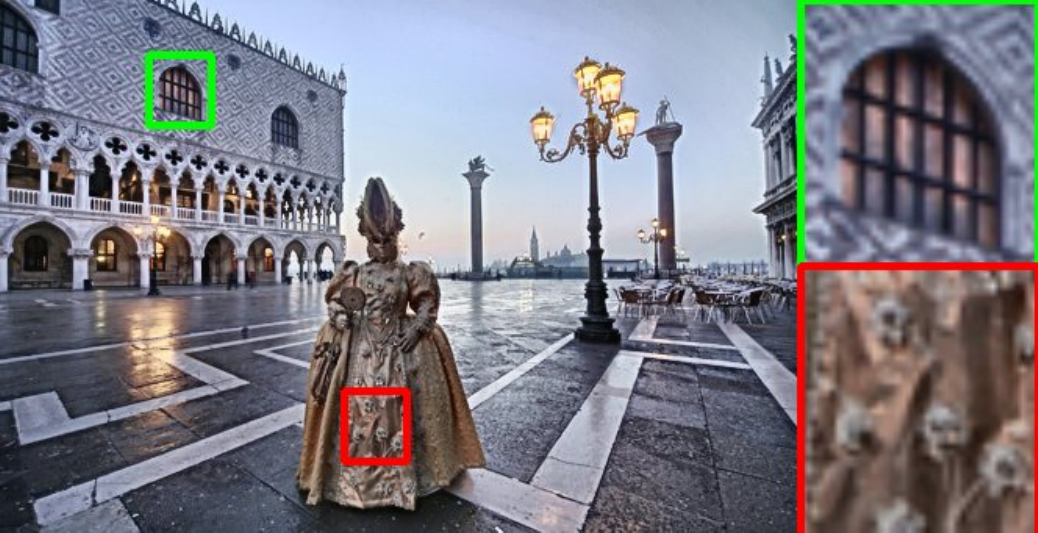}&
				\includegraphics[width=0.166\textwidth]{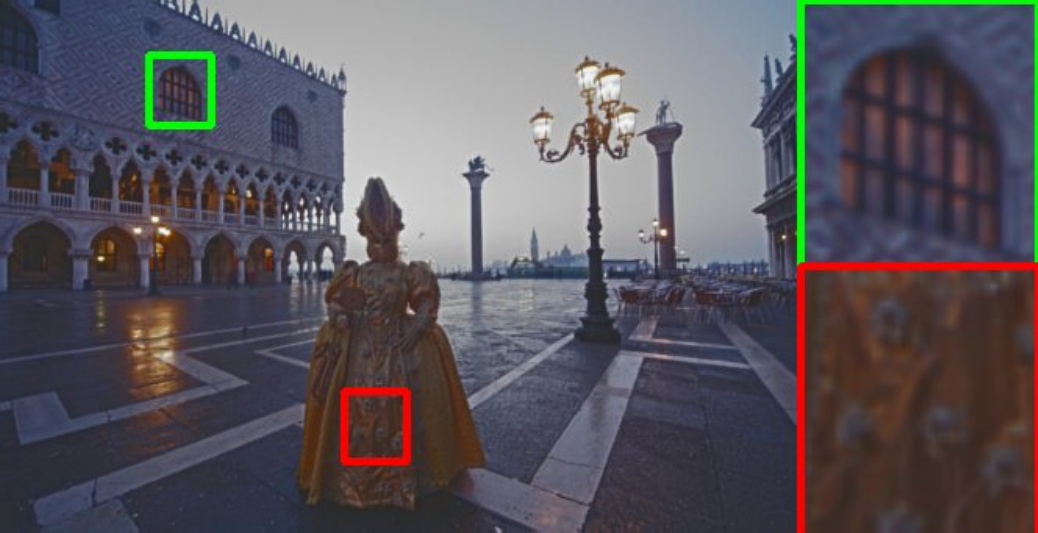}&
				\includegraphics[width=0.166\textwidth]{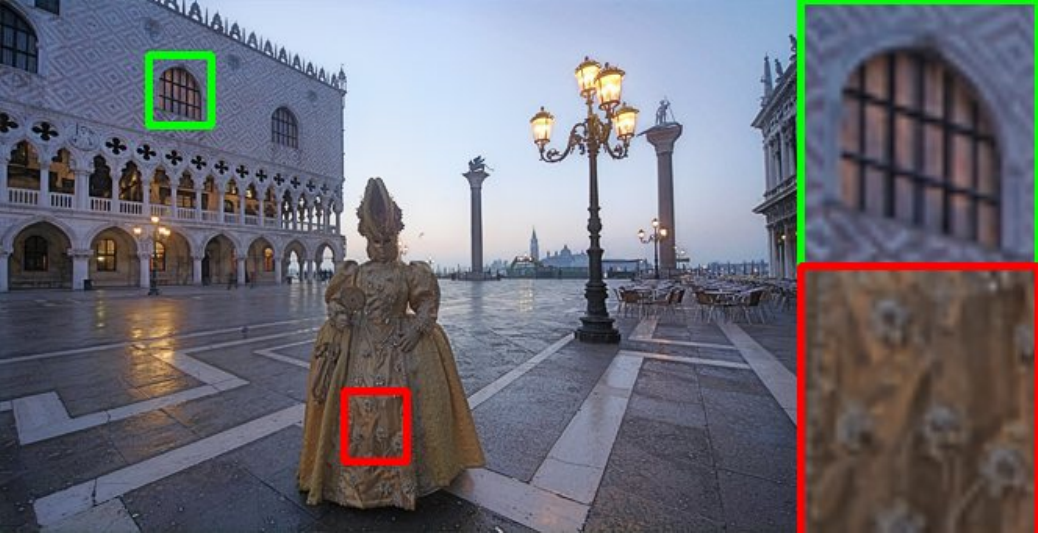}&
				\includegraphics[width=0.166\textwidth]{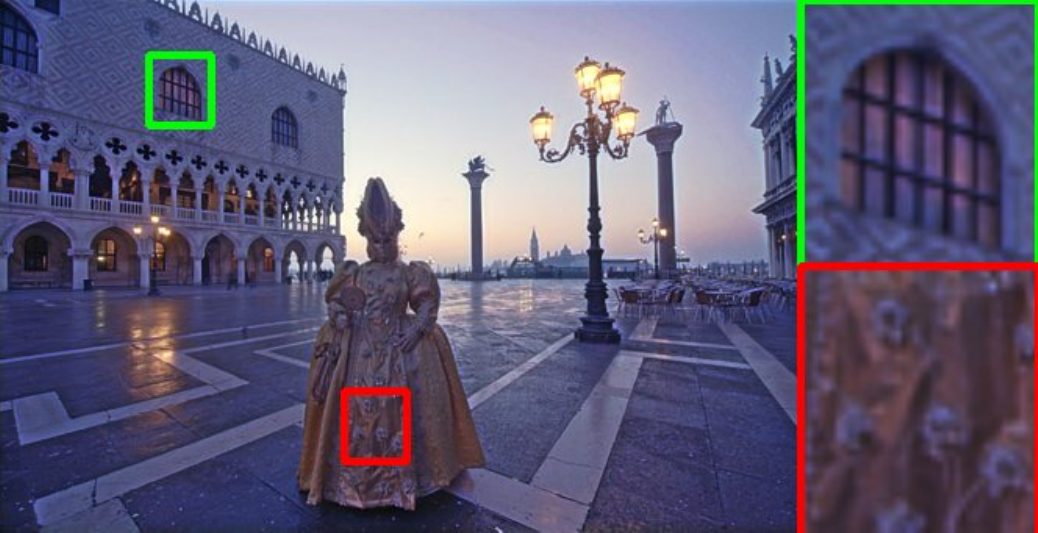}&
				\includegraphics[width=0.166\textwidth]{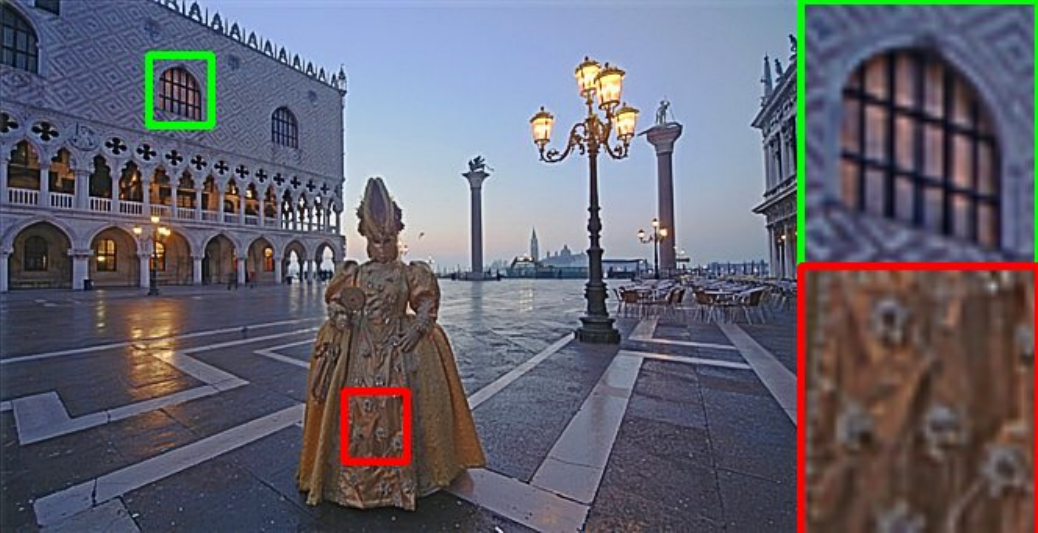}\\
				
				\includegraphics[width=0.166\textwidth]{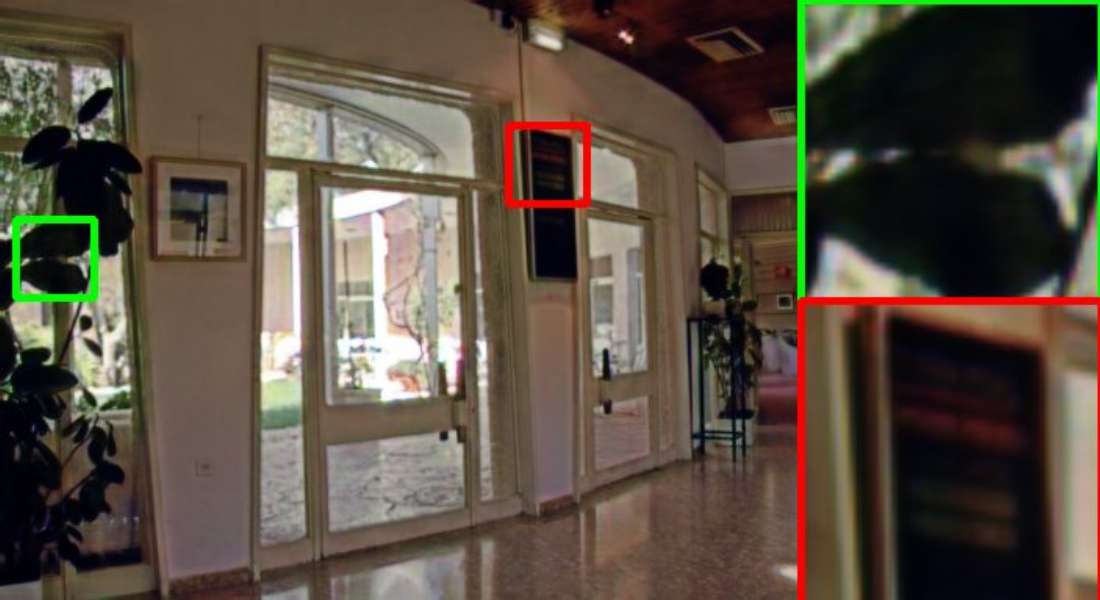}&
				\includegraphics[width=0.166\textwidth]{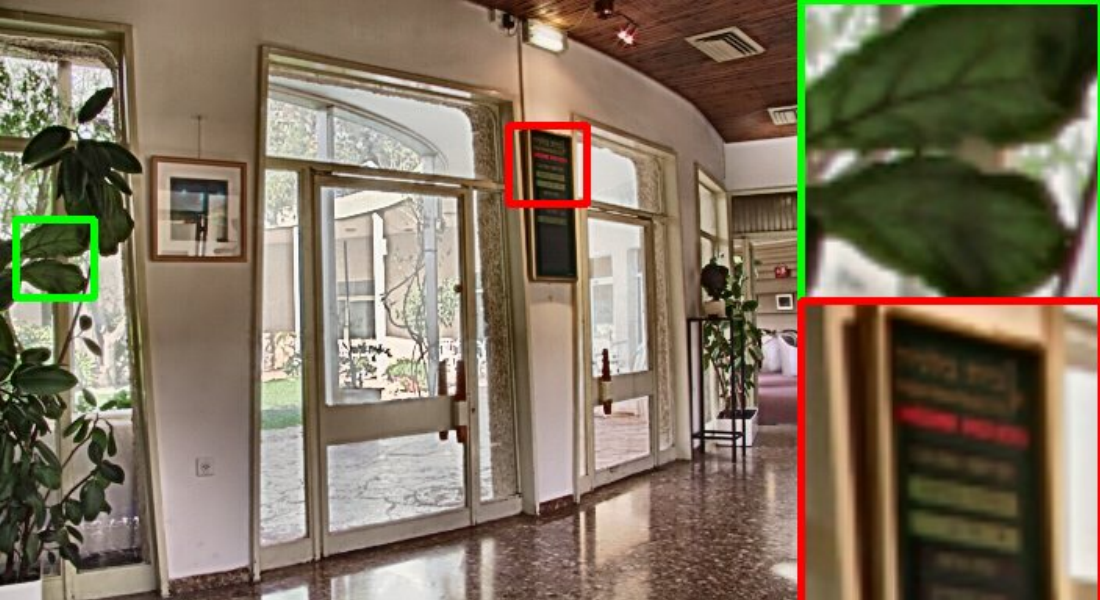}&
				\includegraphics[width=0.166\textwidth]{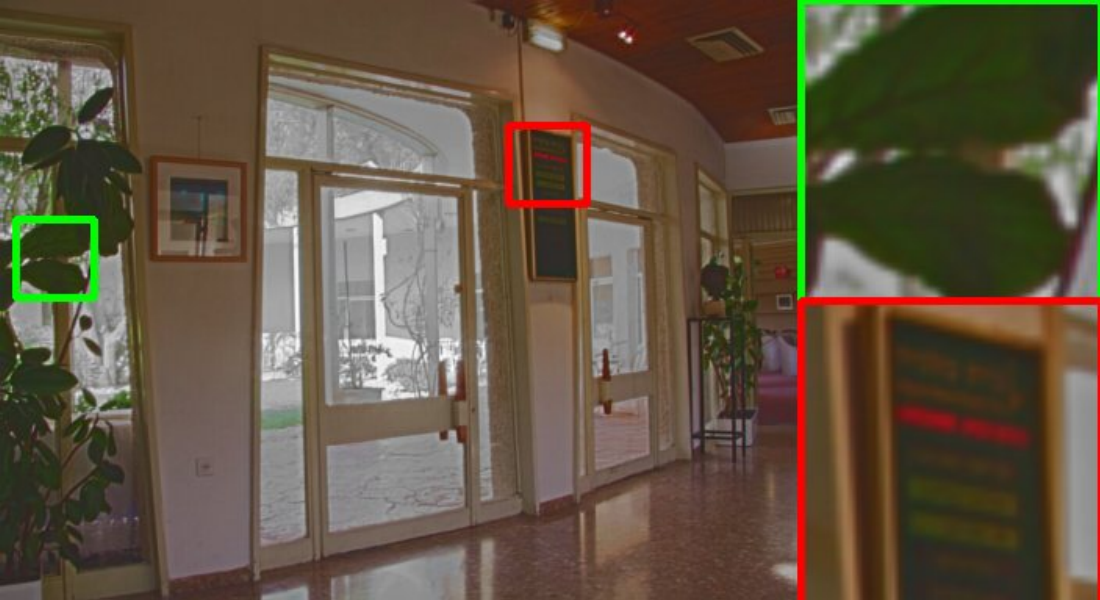}&
				\includegraphics[width=0.166\textwidth]{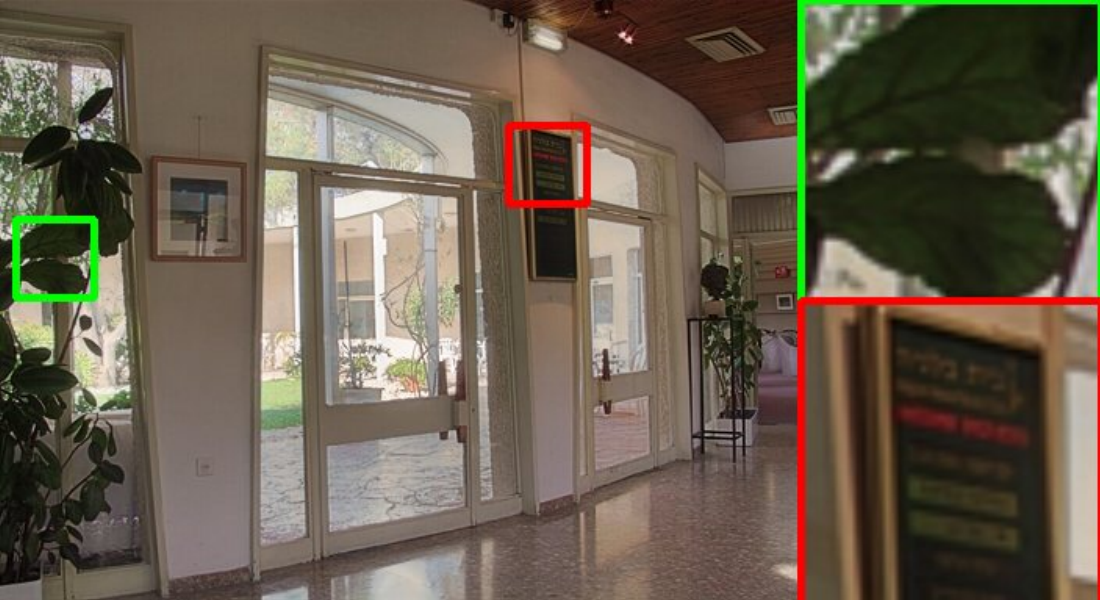}&
				\includegraphics[width=0.166\textwidth]{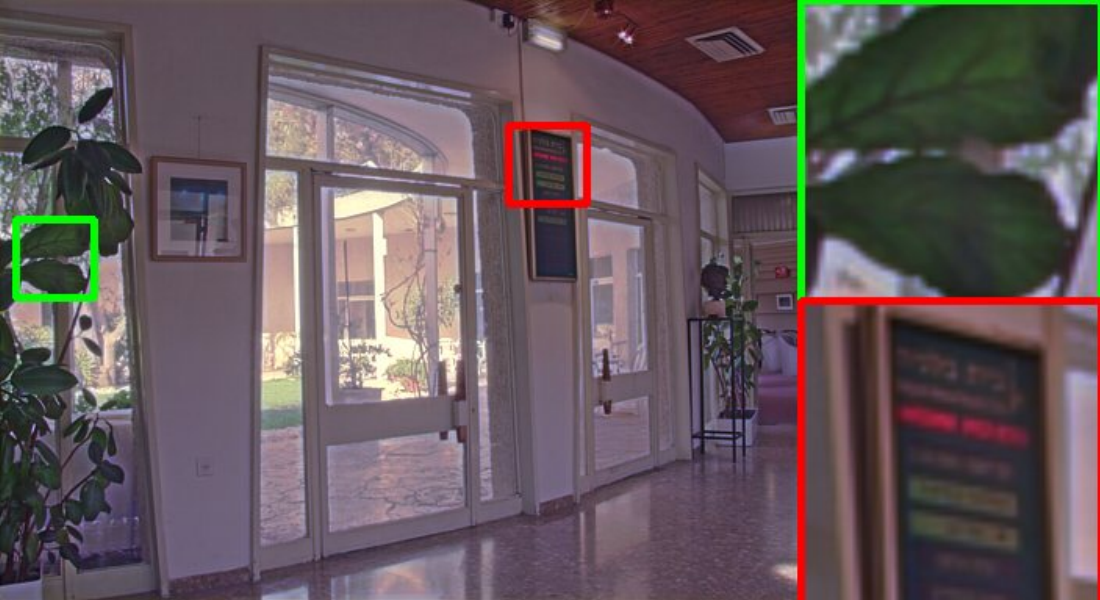}&
				\includegraphics[width=0.166\textwidth]{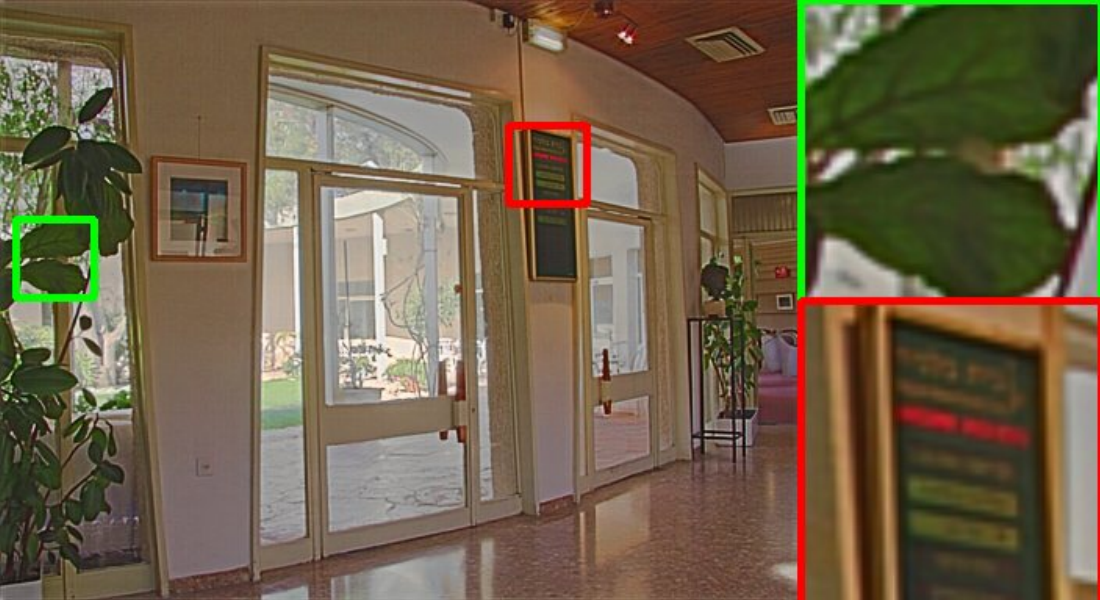}\\
				
				(f) MEFGAN&  (g) IFCNN&  (h) U2Fusion&  (i) DPEMEF&  (j) AGAL&  (k) Ours\\
			\end{tabular}
		\end{center}
		\caption{Visual comparison of our method with other state-of-the-art methods on datasets \cite{cai2018learning,zhang2021benchmarking,gallo2009artifact,ma2015perceptual}. Our method performs the best visual quality in terms of natural color and details.}
		\label{fig:SICE}
	\end{figure*}

\section{METHOD}
	
	Multi-exposure image fusion task aims to combine and merge multiple input images, such as $\mathbf{Y}_1$ and $\mathbf{Y}_2$, into a single output image $\mathbf{Y}_f$ that contains all the relevant information from the input images. Our proposed method, as depicted in Figure~\ref{fig:architecture}, comprises of four modules: a Gamma Correction Module (GCM) for exploring the latent information in source images, a shadow encoder for feature extraction, a Texture Enhancement Module (TEM) for full details completion, and a decoder. Additionally, a Color Enhancement (CE) algorithm is included to enhance the pale result and generate images with rich colors. 
	This work develops an unsupervised multi-exposure image fusion architecture with boosting the hierarchical features. First, the two source images are initially subjected to a GCM to produce two novel images that incorporate obscured details from the originals. The four resultant images are then subjected to feature extraction, followed by deep feature fusion using an modified transformer block. After decoder, the final image is subsequently passed through a color-related algorithm CE to augment its color information. 
	In the following subsections, we will provide a detailed description of these modules and loss function.

	\begin{figure*}[htbp]
		\centering
		\includegraphics[width=\linewidth]{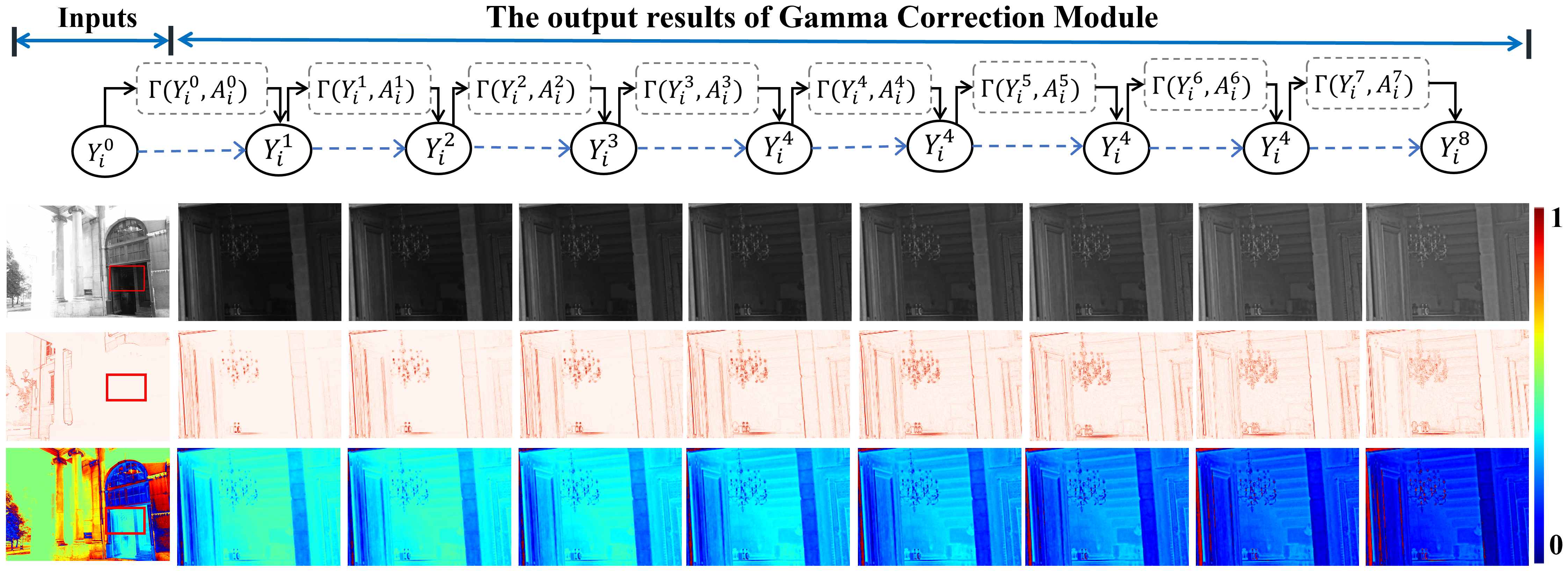}
		\caption{Visual display of GCM output results in important areas of source images. It includes three types of images: the GCM iteration result, the gradient image, and the image representing the difference between the original pixel values and 128.}
		\label{fig:GCM}
	\end{figure*}

	\subsection{Gamma Correction Module(GCM)}
		The proposed Gamma Correction Module (GCM) aims to exploit latent information in source images to improve the fusion process for completing details. The GCM architecture, illustrated in Figure~\ref{fig:architecture}-(a), comprises multiple convolutional layers and a iterative process, with the number of layeres set to 5 in this study. Drawing inspiration from recent works~\cite{kumar2021improved,guo2020zero}, we propose a gamma correction operater $\Gamma$ to enhance underexposed and overexposed images, expressed as:
		\begin{equation}
			\begin{array}{l}
				\mathbf{Y}_i^n=\Gamma( \mathbf{Y}_i^{n-1},\mathbf{A}_i^n) = \mathbf{Y}_i^{n-1} +  \bm{\beta},\\
				\bm{\beta} = \mathbf{Y}_i^{n-1}-(\mathbf{Y}_i^{n-1})^{\mathbf{A}_i^n}, n = 1,\cdots, N,\ i= 1,2,
			\end{array}
			\label{equation:GC}
		\end{equation}
		where $\mathbf{Y}_i$ denotes the Y channel of over/under exposure image and $\mathbf{Y}_i^n$ represents the enhanced image obtained by the $n$-th iteration of $\mathbf{Y}_i$. The operations are performed pixel-wise and each pixel of $\mathbf{Y}_i$ is normalized to [0, 1]. The output of the last convolutional layer, $\mathbf{A}_i^n \in [0.8,1.4]$, controls the exposure level and the last convolutional layer outputs $N$ channels with the same size as $\mathbf{Y}_i$. 
		Figure~\ref{fig:GCM} shows a series of images enhanced by GCM, demonstrating that the previously latent details in both under and overexposed images are gradually revealed, leading to an improvement in overall image quality.

	\subsection{Denoise}
		In low-light regions, noise can often be present and it is known to be amplified during the fusion process~\cite{crighton1975basic}. Furthermore, since the proposed GCM corrects the exposure of the image, it is likely to cause noise amplification, therefore, denoising the enhanced image becomes imperative. 
		In this work, we employ FFDNet~\cite{zhang2018ffdnet} as our denoising module, and detailed structure can be found in~\cite{zhang2018ffdnet}.
		The denoising operation can be mathematically formulated as follows:
		\begin{equation}
			\widehat{\mathbf{Y}}_i=\mathcal{D}(\mathbf{Y}_i^N),\ i= 1,2,
		\end{equation}
		where $N$ denotes the number of GCM iterations, $\mathbf{Y}_i^N$ represents the result of $N$ GCM iterations. And $\mathcal{D}$ denotes the denoise operation. Over-exposed image $\mathbf{Y}_1^N$ and under-exposed image $\mathbf{Y}_2^N$ are processed separately to obtain $\widehat{\mathbf{Y}}_1$ and $\widehat{\mathbf{Y}}_2$, respectively. These result are then input into the Shadow Encoder together as a set $R:=\{\widehat{\mathbf{Y}}_1, \mathbf{Y}_1, \mathbf{Y}_2, \widehat{\mathbf{Y}}_2\}.$

	\subsection{Shadow Encoder and Decoder}
		Shadow Encoder comprises two convolutional layers and a transformer block. The convolutional layeres are adopted at early visual processing and extracting local semantic information, resulting in more stable optimization and improved performance~\cite{xiao2021early}. Transformer block, on the other hand, is capable of extracting global information and establishing long-distance connections. 
		The operation of Shadow Encoder can be mathematically formulated as:
		\begin{equation}
			\mathbf{F}_i=\mathcal{T}(\mathsf{conv}(\mathsf{conv}(\mathbf{Y}_i))),\  \widehat{\mathbf{F}}_i=\mathcal{T}(\mathsf{conv}(\mathsf{conv}(\widehat{\mathbf{Y}}_i))),\ i = 1,2,		
		\end{equation}
		where $\mathcal{T}$ denotes the operator of transformer block (i.e., TB). 
		
		The structure of Decoder is similar to the above Shadow Encoder. It consists of a transformer block and two convolutional layers. For the specific structure, please see Figure~\ref{fig:architecture}-(e). The operation of Decoder can be mathematically formulated as follows:
		\begin{equation}
			\widetilde{\mathbf{Y}}_f=\mathsf{conv}(\mathsf{conv}(\mathcal{T}(\mathbf{F}_T)))
		\end{equation}
		where $\mathbf{F}_T$ denotes the output feature of fusion layer in TEM. 

	\subsection{Texture Enhancement Module}
		The Texture Enhancement Module is designed to fully supplement the details in fusion process, as shown in Figure \ref{fig:architecture}-(b). To effectively utilize $\widehat{\mathbf{F}}_{1}$ and $\widehat{\mathbf{F}}_{2}$ in supplementing the details of $\mathbf{F}_{1}$ and $\mathbf{F}_{2}$, we propose an TEM, inspired by~\cite{ma2022swinfusion}, which can be formulated as:
		\begin{equation}
			\begin{array}{c}
				\{{\mathbf{\widetilde{K}},\mathbf{\widetilde{V}}}\}={\widehat{\mathbf{F}}_i \mathbf{W}^{\mathbf{\widetilde{K}}},\widehat{\mathbf{F}}_i \mathbf{W}^{\mathbf{\widetilde{V}}}}, \ 
				\{{\mathbf{Q}}\} = {\mathbf{F}_i\mathbf{W}^{\mathbf{Q}}}, \\
				{\mathsf{Attention}(\mathbf{Q},\mathbf{\widetilde{K}},\mathbf{\widetilde{V}})}=\mathsf{softmax}(\frac{\mathbf{Q}\mathbf{\widetilde{K}^\top}}{\sqrt{d_{k}}}+\mathbf{B})\mathbf{\widetilde{V}},
			\end{array}
			\label{equation:att}
		\end{equation}
		where $\mathbf{W}^{\mathbf{Q}},\mathbf{W}^{\mathbf{\widetilde{K}}},\mathbf{W}^{\mathbf{\widetilde{V}}}\in\mathbb{R}^{d\times d}$, and $\mathbf{F}_i\in\mathbb{R}^{n\times d}$ represents the features of under/overexposed image, while $\widehat{\mathbf{F}}_i\in\mathbb{R}^{n\times d}$ represents the features of corresponding enhanced image. $N$ and $d$ denote the number of tokens and the dimension of one token, respectively.  		
		After detail completion, we obtain two features, $\bm{\mathcal{F}_1}$ and $\bm{\mathcal{F}_2}$. Then, with the fusion layer, we obtain $\mathbf{F}_T$. Additionally, to address the loss of high-frequency information often caused by forward propagation in fusion networks~\cite{hou2021generative,zhao2021efficient}, we propose a simple and efficient attention mechanism. Features $\bm{\mathcal{F}_1}$ and  $\bm{\mathcal{F}_2}$ are passed through several convolutional layers and a sigmoid activation function to generate attention maps $\mathbf{M}_1$ and $\mathbf{M}_2$. These maps are then multiplied by $\widehat{\mathbf{Y}}_1$ and $\widehat{\mathbf{Y}}_2$, respectively. Then, added them to the output generated by the Decoder to obtain the final fusion result $\mathbf{Y}_f$ for the Y channel.

	\subsection{Color Enhancement}
		In unsupervised methods for image processing, the color channels Cb and Cr are often processed using a standard Eq~\ref{equ:CbCr}, while the processing is limited to the Y channel. Consequently, the resulting images lack vibrancy and have lower saturation. To address this limitation, we propose a color enhancement skill (CE). 
		With the aim to enhance the color channels while preserving the overall image quality, we first convert the input images $\mathbf{I}_1$ and $\mathbf{I}_2$ into the YCbCr color space to obtain $Cb_1$, $Cr_1$, $Cb_2$, and $Cr_2$. Then, $Cb_f$ and $Cr_f$ are obtained using Eq~\ref{equ:CbCr}. Next, $Y_f$, $Cb_f$, and $Cr_f$ are converted into the RGB and HSL color spaces. 
		Subsequently, we obtain the enhancement factor $\alpha$ using the following mathematical formula:
		\begin{equation}
			\alpha=\left\{
			\begin{aligned}
				(1-S)/S, \quad\quad\quad     & S +\delta > 1 \\
				\delta/(1-\delta), \quad\quad\quad &  S + \delta \leq 1 
			\end{aligned} 
			\right. 
		\end{equation}
		where $S$ denotes $S$ channel of HSL color space, $\delta$ represents the color enhancement factor. Finally, we modify the RGB three channels using the obtained $\alpha$ as follows:
		\begin{align}
			\mathbf{Y}=\mathbf{Y}+(\mathbf{Y}-\mathbf{L}) \times \alpha 
		\end{align}
		where $\mathbf{Y} \in \{R,G,B\}$, $\mathbf{L}$ denotes the L channel of HSL color space. After the above processing, we will get the image with higher saturation and richer color information.

	\begin{table*}[t]
		\centering
		\small
		\caption{Quantitative comparison on the SICE \cite{cai2018learning} dataset and Benchmark \cite{zhang2021benchmarking} dataset in terms of various metrics. The top 2 results are highlighted in \textcolor{red}{\textbf{red}} and \textcolor{blue}{\textbf{blue}}, respectively.}
		\label{tab:main_metrics1}
		\resizebox*{0.9\textwidth}{!}{%
			\begin{tabular}{c|ccccc|ccccc}
				\hline
				Datasets    & \multicolumn{5}{c|}{SICE}  & \multicolumn{5}{c}{Benchmark}  \\ \hline
				Metrics     & PSNR $\uparrow$   & CS $\uparrow$ & CC $\uparrow$ 
				& NMI $\uparrow$ & \multicolumn{1}{c|}{$Q_{ncie}$ $\uparrow$}    & PSNR $\uparrow$    & CS $\uparrow$ & CC $\uparrow$ 
				& NMI $\uparrow$ & $Q_{ncie}$ $\uparrow$ \\ \hline
				
				DSIFT  & 58.1404  & 0.4593 & 0.2827 & 0.5580  	&\multicolumn{1}{c|}{0.8150} 
				& 56.6248  & 0.3384 & 0.4737 & 0.5994 & 0.8171 \\
				
				GFF   &58.0796 	 	&0.3898 	&0.2465 	&0.6683 	&\multicolumn{1}{c|}{0.8174} 
				&56.5521  	&0.3094 	&0.4132 	&0.6360 	&0.8179 \\
				
				SPDMEF  &\textcolor{blue}{\textbf{58.5985}}  	&0.4883 	&0.8092 	&0.5935 	&\multicolumn{1}{c|}{0.8131}
				&57.1662  	&0.3750 	&0.8317 	&0.6816 	&0.8178 \\
				FMMEF &58.4304 	&0.4758 	&0.5449 	&0.4718 		&\multicolumn{1}{c|}{0.8116} 
				&57.0949 	&0.3715 	&0.6524 	&0.5046 	&0.8135 \\
				
				MEFNet  &58.2663  	&0.4125 	&0.4660 	&0.6763	&\multicolumn{1}{c|}{\textcolor{red}{\textbf{0.8201}}}
				&56.6227  	&0.3471 	&0.5108 	&0.6624 	&0.8166 \\

				MEFGAN &58.4329  	&\textcolor{blue}{\textbf{0.7323}} 	&0.8634 	&0.5697 	&\multicolumn{1}{c|}{0.8118}
				&56.9687  	&\textcolor{red}{\textbf{0.5044}} 	&0.8513 	&0.5906 	&0.8141 \\
				
				IFCNN &58.3080 	&0.4002 	&0.7942 	&0.4071 	&\multicolumn{1}{c|}{0.8088}
				&56.9596 	&0.3042 	&0.8173 	&0.4292 	&0.8110 
				\\
				
				U2Fusion &58.3545 	&0.3142 	&\textcolor{blue}{\textbf{0.8977}} 	&\textcolor{blue}{\textbf{0.6784}} 	&\multicolumn{1}{c|}{0.8138} 
				&57.0603 	&0.2849 	&\textcolor{red}{\textbf{0.9070}} 	&\textcolor{blue}{\textbf{0.7520}}	& 0.8181
				\\
				
				AGAL &58.5971  	&0.5333 	&0.8920	&0.6604 	&\multicolumn{1}{c|}{0.8140} 
				&\textcolor{blue}{\textbf{57.1929}} 	&0.4136 	&\textcolor{blue}{\textbf{0.9062}} 	&0.7401	&\textcolor{blue}{\textbf{0.8183}} 
				
				\\
				
				DPEMEF &58.5404  	&0.4767 	&0.8734 	&0.5601 	&\multicolumn{1}{c|}{0.8116}
				&57.1796  	&0.3579 	&0.8875 	&0.6169 	&0.8149 
				\\
				\hline
				\textbf{BHFMEF} &\textcolor{red}{\textbf{58.7459}} 	 	 &\textcolor{red}{\textbf{0.7350}}	&\textcolor{red}{\textbf{0.9054}} 	&\textcolor{red}{\textbf{0.7210}} 	&\multicolumn{1}{c|}{\textcolor{blue}{\textbf{0.8182}}}
				&\textcolor{red}{\textbf{57.3213}} 		&\textcolor{blue}{\textbf{0.4707}} 	&0.9042	&\textcolor{red}{\textbf{0.7573}} 	&\textcolor{red}{\textbf{0.8185}} 
				\\
				\hline
			\end{tabular}%
		}
	\end{table*}
	
	\begin{table*}[htbp]
		\centering
		\small
		\caption{Quantitative comparison on the dataset~\cite{gallo2009artifact} and dataset~\cite{ma2015perceptual} in terms of various metrics. The top 2 results are highlighted in \textcolor{red}{\textbf{red}} and \textcolor{blue}{\textbf{blue}}, respectively.}
		\label{tab:main_metrics2}
		\resizebox*{0.9\textwidth}{!}{%
			\begin{tabular}{c|ccccc|ccccc}
				\hline
				Datasets    & \multicolumn{5}{c|}{dataset~\cite{gallo2009artifact}}                                                     & \multicolumn{5}{c}{dataset~\cite{ma2015perceptual}}                                                    \\ \hline
				Metrics     & PSNR $\uparrow$    & CS $\uparrow$ & CC $\uparrow$ 
				& NMI $\uparrow$ & \multicolumn{1}{c|}{$Q_{ncie}$ $\uparrow$}    & PSNR $\uparrow$  & CS $\uparrow$ & CC $\uparrow$ 
				& NMI $\uparrow$ & $Q_{ncie}$ $\uparrow$ \\ \hline
				
				DSIFT  & 55.3715  	 &0.3044 	 &0.5676  &	0.5923   	&\multicolumn{1}{c|}{0.8154 } 
				& 59.3119  &0.1665 &0.2207 &0.6012  &0.8177\\
				
				GFF   &55.2629 	 	&0.3076 &0.4611 &0.6328 
				&\multicolumn{1}{c|}0.8165 
				&59.3359  &0.1261 &0.0938 &0.6654 &0.8163 \\
				
				SPDMEF  &55.6466  &0.3440 &0.7988 &0.6316 
				&\multicolumn{1}{c|}{0.8156 }
				&59.3359  &0.1261 &0.0938 &0.6654 &0.8163 \\
				
				FMMEF &55.5231  &0.3493 &0.6628 &0.4916 
				&\multicolumn{1}{c|}{0.8120 } 
				&59.5876  &0.1774 &0.4719 &0.4807 &0.8131 \\
				
				MEFNet  &55.2349  &0.3031 &0.5494 
				&\textcolor{blue}{\textbf{0.7357}} 		&\multicolumn{1}{c|}{\textcolor{red}{\textbf{0.8222}}}
				&59.6060  &0.1711 &0.5349 &0.7604 &0.8142 
				\\
				MEFGAN &55.6308	&\textcolor{blue}{\textbf{0.4548}} 	&0.8405 	&0.5314	&\multicolumn{1}{c|}{0.8122}
				&59.8707  &\textcolor{blue}{\textbf{0.2710}} &0.8695 &0.5984 &0.8138  \\
				
				IFCNN &55.4494	&0.2972 &0.7525 &0.3888 &\multicolumn{1}{c|}{0.8097}
				&59.7784  &0.1500 &0.8257 &0.4255 &0.8102 \\
				
				U2Fusion &55.6616  &0.2931 &0.8791 
				&0.7015 
				&\multicolumn{1}{c|}{0.8162} 
				&\textcolor{blue}{\textbf{59.9853}}  &0.2275 &0.9104 &\textcolor{blue}{\textbf{0.8459}} &\textcolor{blue}{\textbf{0.8194}} \\
				
				AGAL &55.7888  &0.4180 
				&\textcolor{blue}{\textbf{0.8799}} 	&0.6827 	&\multicolumn{1}{c|}{0.8160} 
				&59.9314  &\textcolor{red}{\textbf{0.2943}} &\textcolor{red}{\textbf{0.9178}} &0.8013 &0.8184 \\
				
				DPEMEF&\textcolor{blue}{\textbf{55.8197}}  
				&0.3284  	&0.8579 	&0.6153 	&\multicolumn{1}{c|}{0.8144}
				&59.8780  &0.1520 &0.8950 &0.6974 &0.8162 \\
				\hline
				\textbf{BHFMEF} &\textcolor{red}{\textbf{57.3038}} 	 &\textcolor{red}{\textbf{0.4835}}	&\textcolor{red}{\textbf{0.9031}} 	&\textcolor{red}{\textbf{0.7606}} 	&\multicolumn{1}{c|}{\textcolor{blue}{\textbf{0.8188}}}
				&\textcolor{red}{\textbf{60.0354}}  &0.2534 &\textcolor{blue}{\textbf{0.9120}} &\textcolor{red}{\textbf{0.8661}} &\textcolor{red}{\textbf{0.8204}} \\
				\hline
			\end{tabular}%
		}
	\end{table*}

	\subsection{Loss Functions}
		The loss function contains two parts, i.e., the gamma correction loss $\mathcal{L}_{GC}$ and the exposure fusion loss $\mathcal{L}_{EF}$:
		\begin{equation}
			\begin{array}{l}
				\mathcal{L}_{Total} = \mathcal{L}_{GC} + \mathcal{L}_{EF}.
			\end{array}
		\end{equation}
		
		$\textbf{Gamma Correction Loss}$: To guide the training of GCM, we use the spatial consistency loss $\mathcal{L}_{spa}$, exposure control loss $\mathcal{L}_{exp}$ and illumination smoothness loss $\mathcal{L}_{is}$. The $\mathcal{L}_{spa}$ can be
		\begin{align}
			\mathcal{L}_{spa} = \frac{1}{K} \sum_{i=1}^K \sum_{j\in \Omega(i)} (|(Y_i - Y_j)|-|(I_i - I_j)|)^2,
		\end{align}
		where $K$ is the number of local region, and $\Omega(i)$ is the four neighboring regions (top, down, left, right) centered at region $i$. We denote $Y$ and $I$ as the average intensity value of local region in the enhanced version and input image, respectively. We empirically set the size of local region to $4\times4$.
		\begin{align}
			\mathcal{L}_{exp} = \frac{1}{M} \sum_{k=1}^M |(Y_k - \tau)|,
		\end{align}
		where $M$ represents the number of nonoverlapping local regions of size $16\times16$, $Y$ is the average intensity value of a local region in the enhanced image. $\tau$ is set as the median of dynamic range, e.g., 0.5. 
		The $\mathcal{L}_{GC}$ can be expressed as:
		\begin{align}
			\mathcal{L}_{GC} = \mathcal{L}_{spa} +  W_{exp}\mathcal{L}_{exp} + W_{is}\mathcal{L}_{is},
		\end{align}
		where $W_{exp}$ and $W_{is}$ are the weights of losses.   \begin{align}
			\mathcal{L}_{is} = \frac{1}{N} \sum_{n=1}^N (|\nabla_{x}A^n|+|\nabla_{y}A^n|)^2,
		\end{align}
		where $N$ is the number of iteration, $\nabla_{x}$ and $\nabla_{y}$ represent the horizontal and vertical gradient operations, respectively.
		
		$\textbf{The Exposure Fusion Loss}$: To produce desired results, we design intensity loss, gradient loss, SSIM loss to constrain the network. Their definitions are as follows:
		\begin{align}
			\label{equ:14}
			\mathcal{L}_{int} = \sum\limits_{i}\frac{1}{HW} \|I_f - I^i\|,\  I^i\in\{\widehat{\mathbf{Y}}_{1}, \mathbf{Y}_{1}, \mathbf{Y}_{2}, \widehat{\mathbf{Y}}_{2}\},
		\end{align}
		where $H$,$W$ denotes the height and width of image, respectively.
		\begin{align}
			\mathcal{L}_{grad} =  \frac{1}{HW} \| |\nabla I_f| - \max(|\nabla I^i|)  \|_1,  I^i\in \{\widehat{\mathbf{Y}}_{1}, \mathbf{Y}_{1}, \mathbf{Y}_{2}, \widehat{\mathbf{Y}}_{2}\},
		\end{align}
		where $\nabla$ indicates the Sobel gradient operator, which could measure texture information of an image. $|\cdot |$ stands for the absolute operation, $\|\cdot\|_1$ denotes the $\ell_1$-norm, and $\max(\cdot)$ refers to the element-wise maximum selection.
		\begin{equation}\label{equ:16}
			\begin{array}{l}
			\mathcal{L}_{ssim} = \sum\limits_{i}(1-ssim(I_f,I^{i})),\  I^i\in \{\widehat{\mathbf{Y}}_{1}, \mathbf{Y}_{1}, \mathbf{Y}_{2}, \widehat{\mathbf{Y}}_{2}\},
			\end{array}
		\end{equation}
		where $ssim(\cdot)$ represents the structural similarity operation. Then, $\mathcal{L}_{EF}$ can be expressed as:
		\begin{equation*}
			\begin{array}{l}
			\mathcal{L}_{EF} = W_{int}\mathcal{L}_{int} +  W_{grad}\mathcal{L}_{grad} + W_{ssim}\mathcal{L}_{ssim}, 
		\end{array}
	\end{equation*}
		where $W_{int}$, $W_{grad}$ and $W_{ssim}$ are the hyper-parameters that control the trade-off of each sub-loss term.

\section{EXPERIMENT}
	In this section, we begin by presenting the training and test datasets we utilized, along with the implementation details. We then conduct a qualitative and quantitative comparison of BHF-MEF with ten current state-of-the-art methods on four distinct datasets. Subsequently, we perform ablation experiments to validate the efficacy of each proposed module.
	\subsection{Implementation Details}
		We trained our network on the SICE~\cite{cai2018learning} dataset, which comprises 589 multi-exposure image sequences. We randomly selected 489 sequences for training, while the remaining 100 sequences were reserved for testing. To further evaluate the effectiveness of our proposed method, we conducted experiments on three additional datasets [9, 11, 12], each of which includes 100, 5, and 16 pairs of multi-exposure images, respectively. 
		Our network was trained on an NVIDIA GeForce RTX 3090 GPU with a batch size of 2 for 200 epochs. We utilized an ADAM optimizer and cosine annealing learning rate adjustment strategy, with a learning rate of 1e-6 and a weight decay of 1e-8. During the training phase, we employed a patch size of 256 $\times$ 256 and applied data augmentation techniques such as random flipping and rotation. The GCM's parameter \emph{$n$} was set to 8, and the weights \emph{$W_{exp}$}, \emph{$W_{is}$}, \emph{$W_{int}$}, \emph{$W_{gra}$}, and \emph{$W_{ssim}$} were set to 2, 200, 10, 3, and 2, respectively, to balance the scale of losses. $\delta$ was set to 0.2.
	
	\begin{figure*}[htbp]
		\centering
		\includegraphics[width=\linewidth]{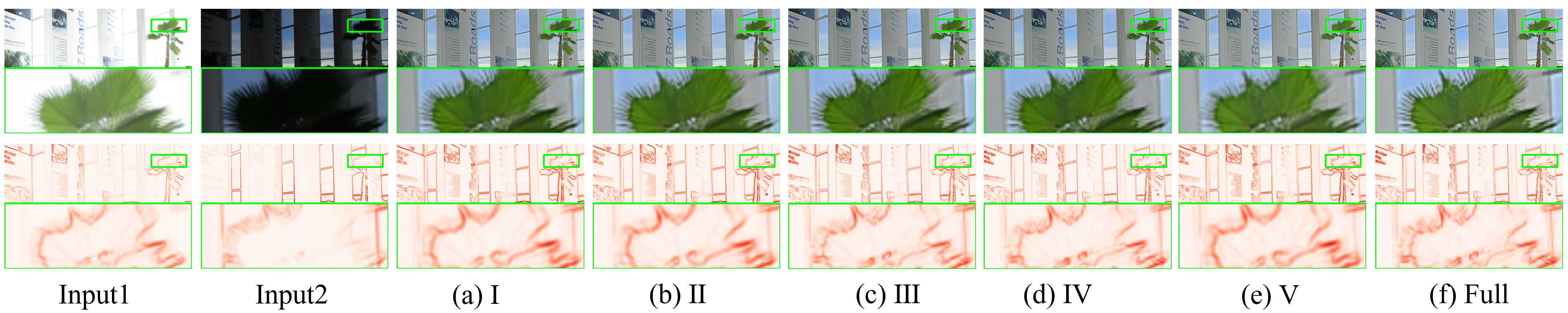}
		\caption{Visual comparison of ablation experiments on TEM. The second row is the corresponding gradient-based details of the first row. The number \uppercase\expandafter{\romannumeral1},\uppercase\expandafter{\romannumeral2},\uppercase\expandafter{\romannumeral3},\uppercase\expandafter{\romannumeral4},\uppercase\expandafter{\romannumeral5} here corresponds to \uppercase\expandafter{\romannumeral1},\uppercase\expandafter{\romannumeral2},\uppercase\expandafter{\romannumeral3},\uppercase\expandafter{\romannumeral4},\uppercase\expandafter{\romannumeral5} in Figure~\ref{fig:Ablation_TEM}.}
		\label{fig:ablation2}
	\end{figure*}

	\begin{table}[htbp]
		\centering
		\caption{Quantitative comparison of ablation on GCM, CE and Denoise. The top results are highlighted in \textcolor{black}{\textbf{black}}.}
		\label{tab:GCM}
			\begin{tabular}{c|ccccc}
				\hline
				Metrics     & PSNR $\uparrow$  & CS $\uparrow$ & CC $\uparrow$ 
				& NMI $\uparrow$ & $Q_{ncie}$ $\uparrow$  \\ \hline
				$n=0 \to n=8$ & 58.6955   &0.6471 &0.8726 &0.6166 &0.8128\\
				$n=1 \to n=8$ &58.6916 &0.6435 &0.8731 &0.6157 &0.8128  \\	
				w/o CE  &58.5707 &0.4658 &0.8606 &0.5666 &0.8116 \\		
				w/o Denoise &58.5522 &0.5269 &0.8540 &0.5337 &0.8110 \\
				Full &\textbf{58.7459} 	&\textbf{0.7350}	&\textbf{0.9054}	&\textbf{0.7210}	&\textbf{0.8182} \\
				\hline
			\end{tabular}
	\end{table}
	
	\begin{table}[htbp]
		\centering
		\caption{Quantitative comparison of ablation on TEM. The top results are highlighted in \textcolor{black}{\textbf{black}}.}
		\label{tab:TEM}
		\resizebox{\columnwidth}{!}{%
			\begin{tabular}{c|ccccc}
				\hline
				Metrics     & PSNR $\uparrow$  & CS $\uparrow$ & CC $\uparrow$ 
				& NMI $\uparrow$ & $Q_{ncie}$ $\uparrow$  \\ \hline
				
				\uppercase\expandafter{\romannumeral1}  &58.6456 &0.6023 &0.8614 &0.5583 &0.8115 \\
				\uppercase\expandafter{\romannumeral2}   &58.6496 &0.5964 &0.8621 &0.5622 &0.8116 \\
				\uppercase\expandafter{\romannumeral3}   &58.6483 &0.6019 &0.8618	&0.5588	&0.8115	\\
				\uppercase\expandafter{\romannumeral4}   &58.6551	&0.6062	&0.8639	&0.5646	&0.8116 \\
				\uppercase\expandafter{\romannumeral5}   &58.5581 	&0.5444 &0.8583 &0.5334 &0.8109 \\
				Full &\textbf{58.7459} 	&\textbf{0.7350} &\textbf{0.9054} &\textbf{0.7210}	&\textbf{0.8182} \\
				\hline
			\end{tabular}
		}
	\end{table}

	\begin{figure}[htbp]
		\centering
		\includegraphics[width=\linewidth]{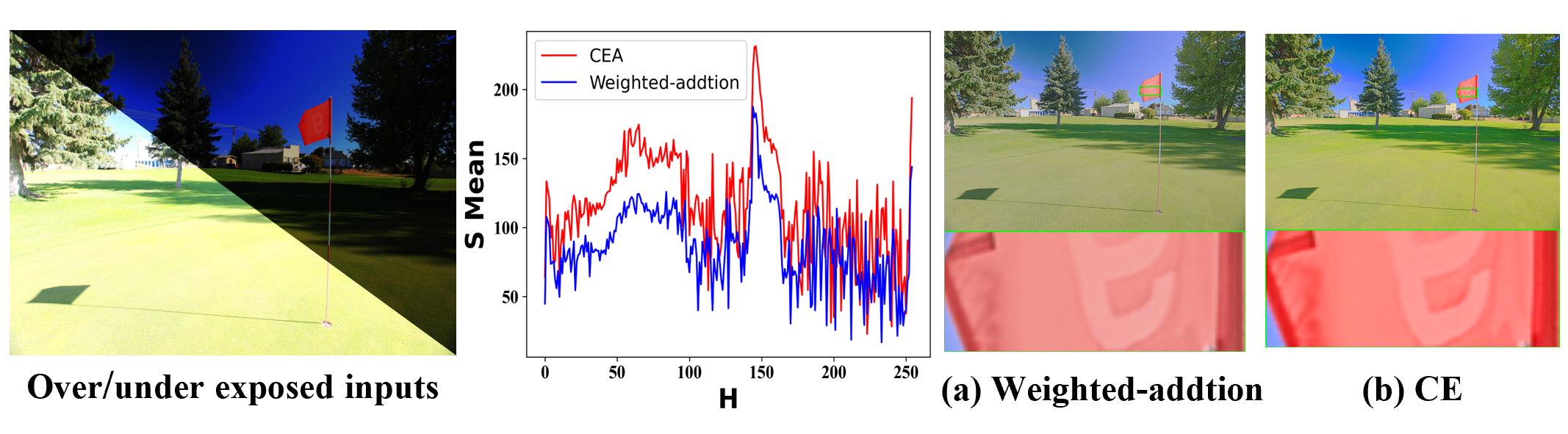}
		\caption{Comparison of ablation experiments on CE. The image (a) is colored by the traditional weighted-addition method in Eq~\ref{equ:CbCr}, and (b) is colored by our CE.}
		\label{fig:CE}
	\end{figure}

	\begin{figure}[htbp]
		\centering
		\includegraphics[width=\linewidth]{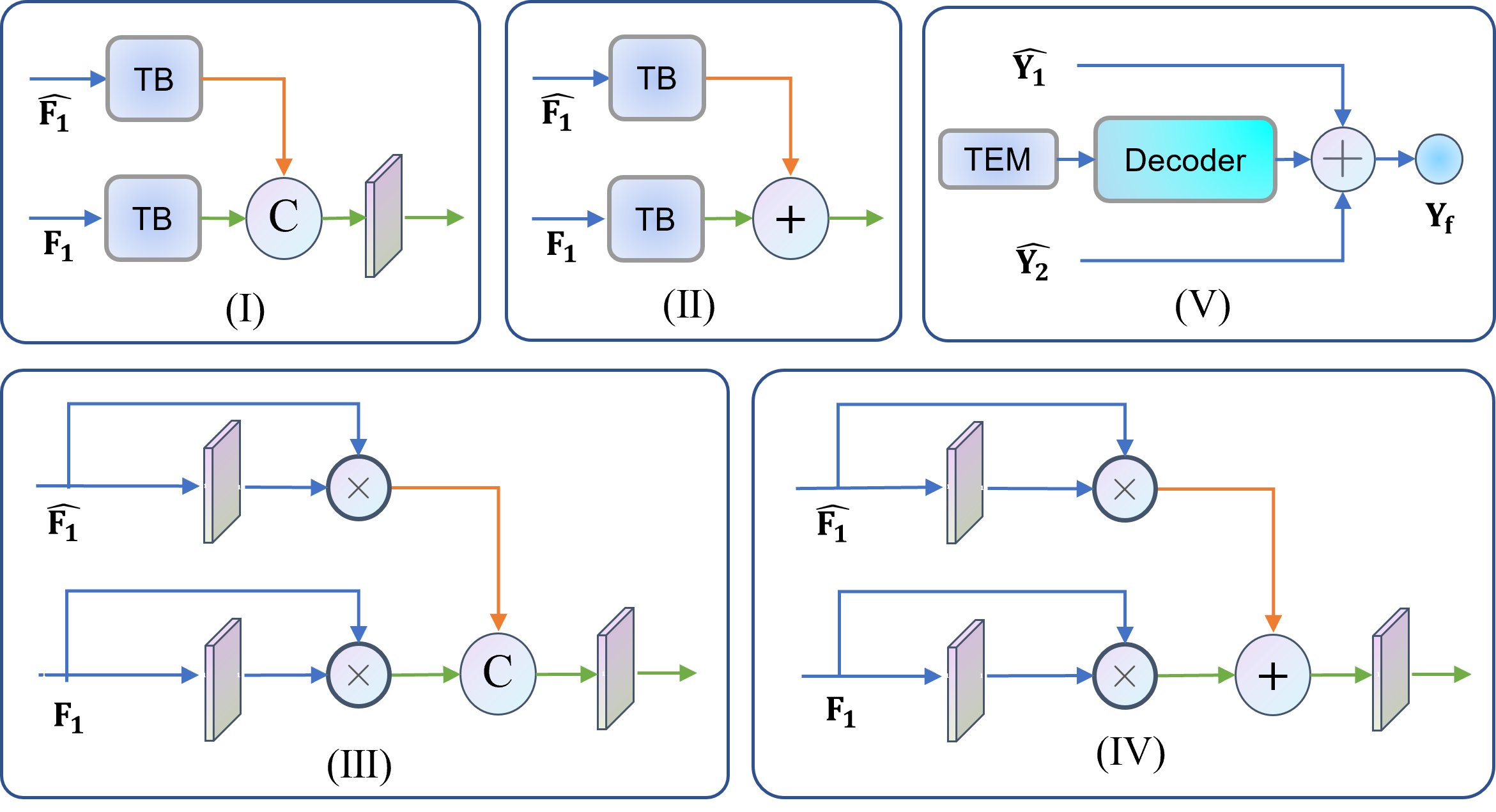}
		\caption{The specifics of the ablation experiments on TEM.}
		\label{fig:Ablation_TEM}
	\end{figure}

	\begin{figure}[htbp]
		\centering
		\begin{center}
			\begin{tabular}{c@{\extracolsep{0.1mm}}c@{\extracolsep{0.1mm}}c}
				
				\includegraphics[width=0.155\textwidth]{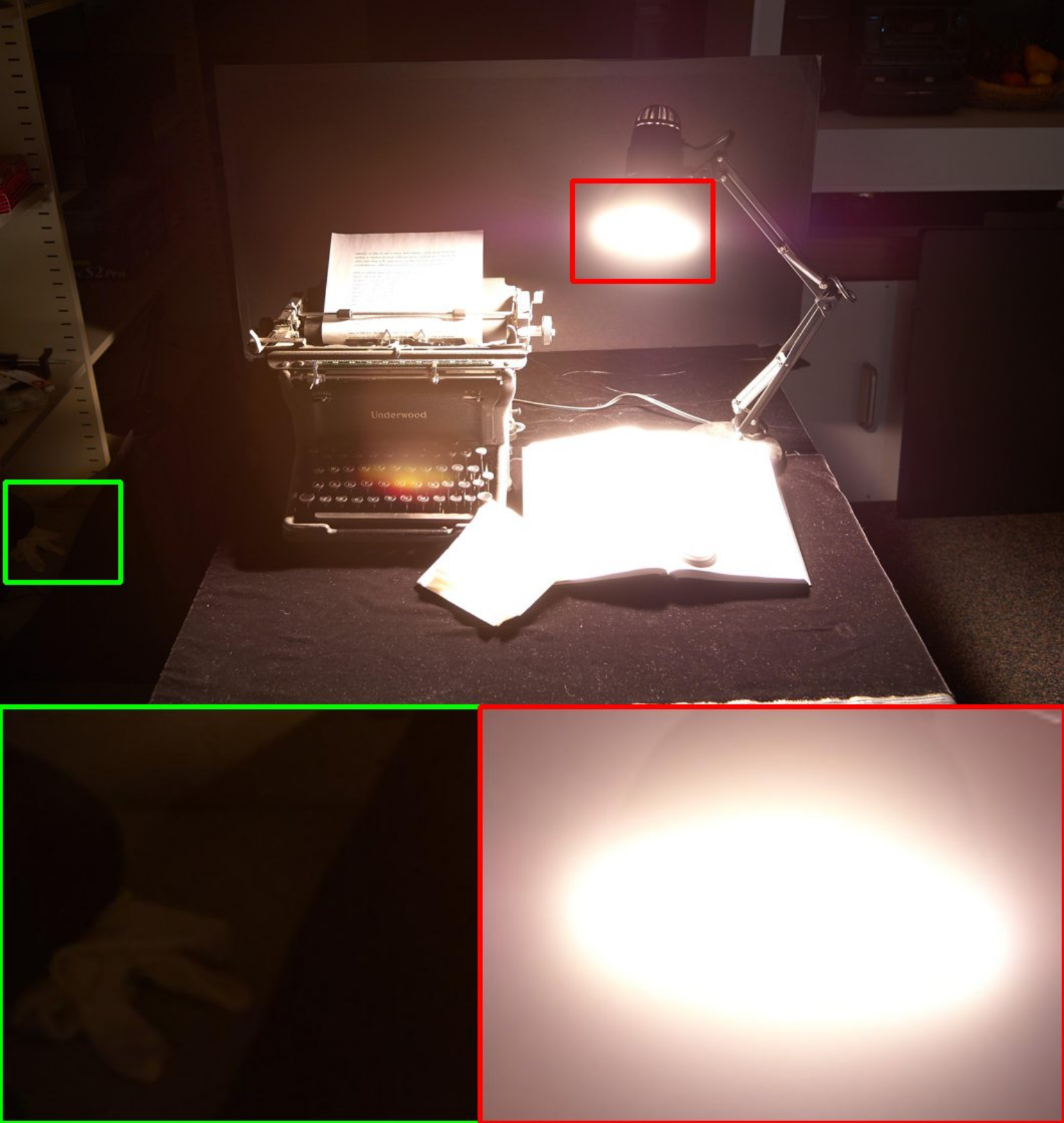}&
				\includegraphics[width=0.155\textwidth]{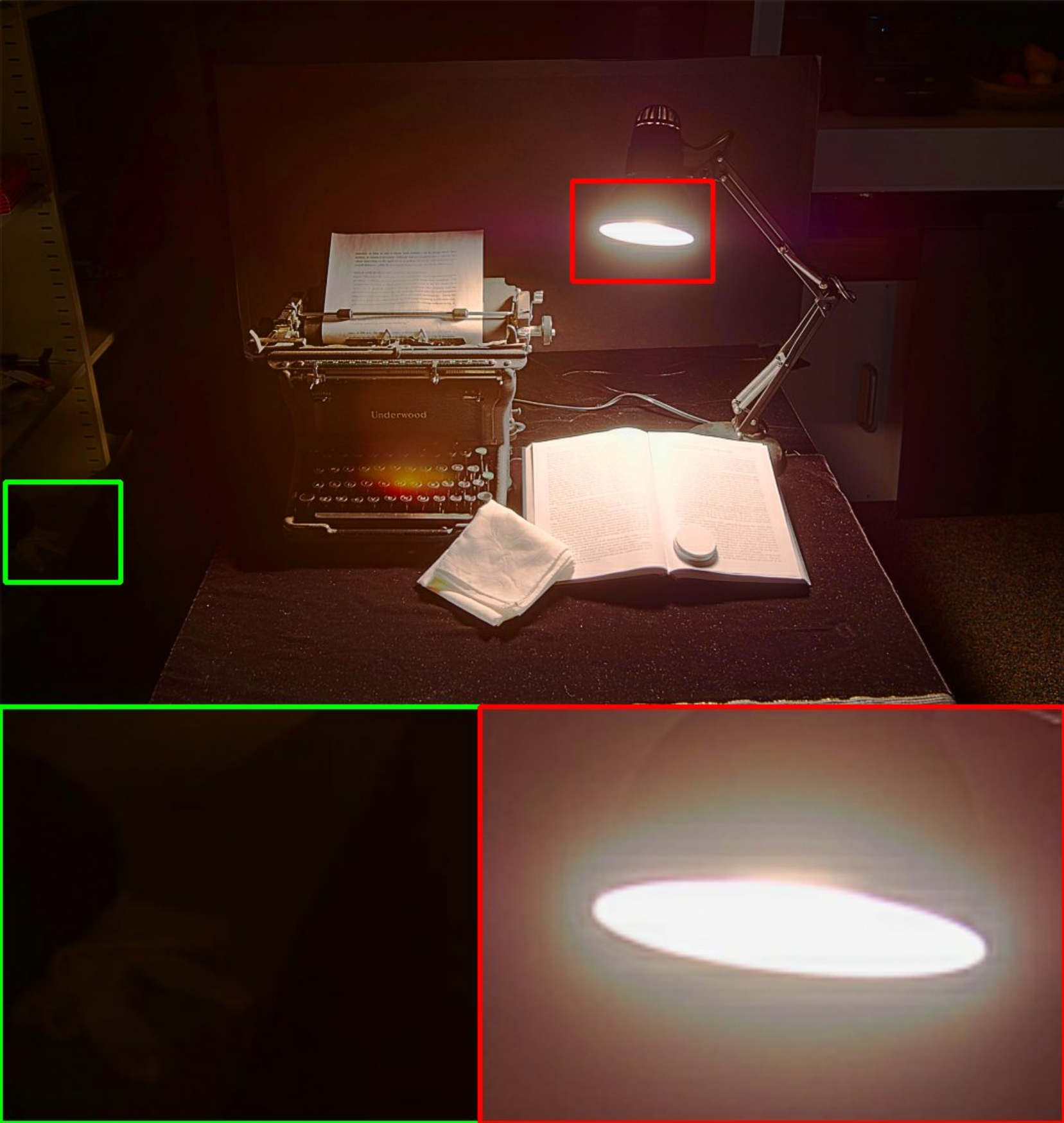}&
				\includegraphics[width=0.155\textwidth]{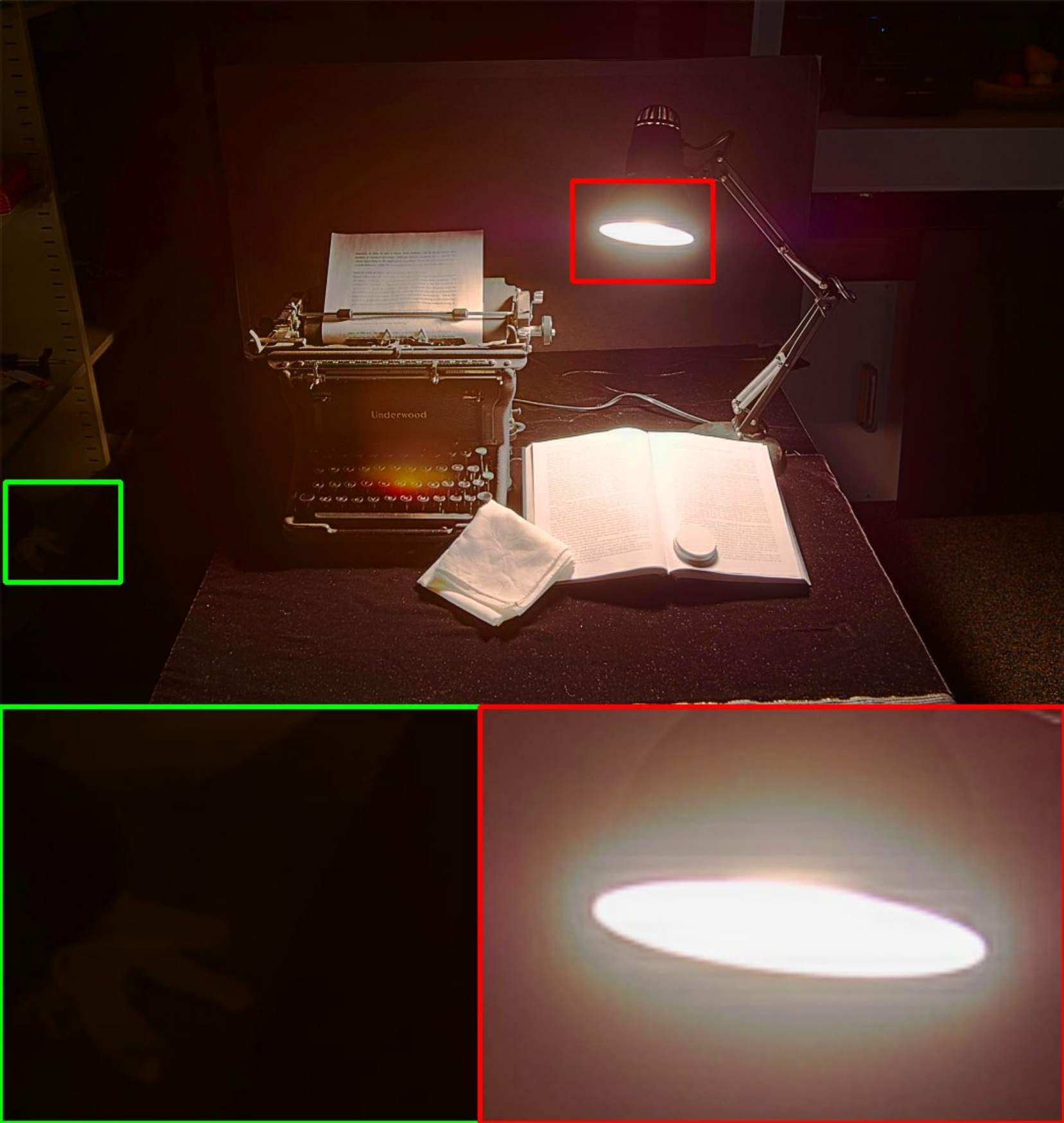}\\
				Input 1& (a) $n=0 \to n=8$& (b) $n=1 \to n=8$ \\
				\includegraphics[width=0.155\textwidth]{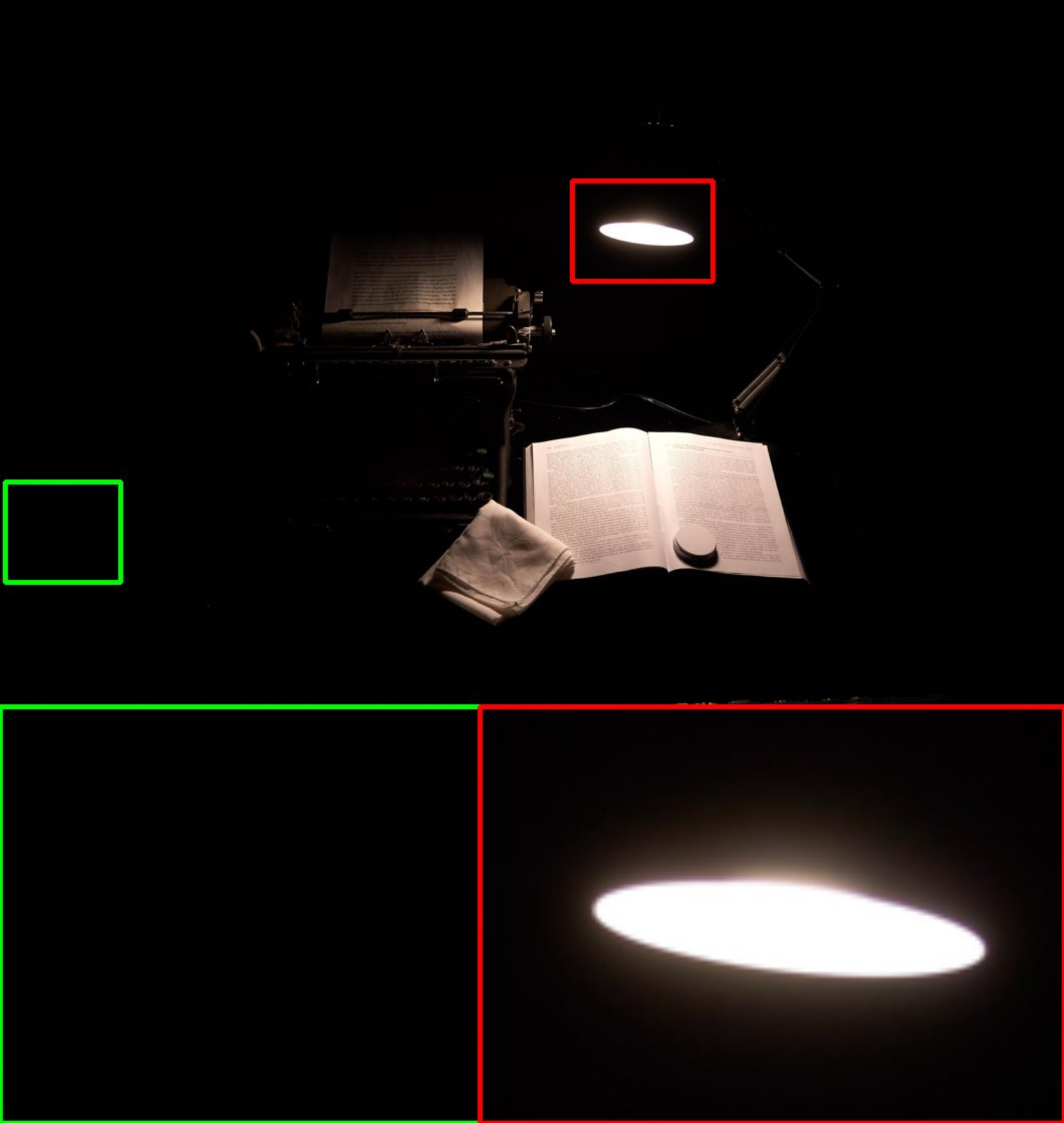}&
				\includegraphics[width=0.155\textwidth]{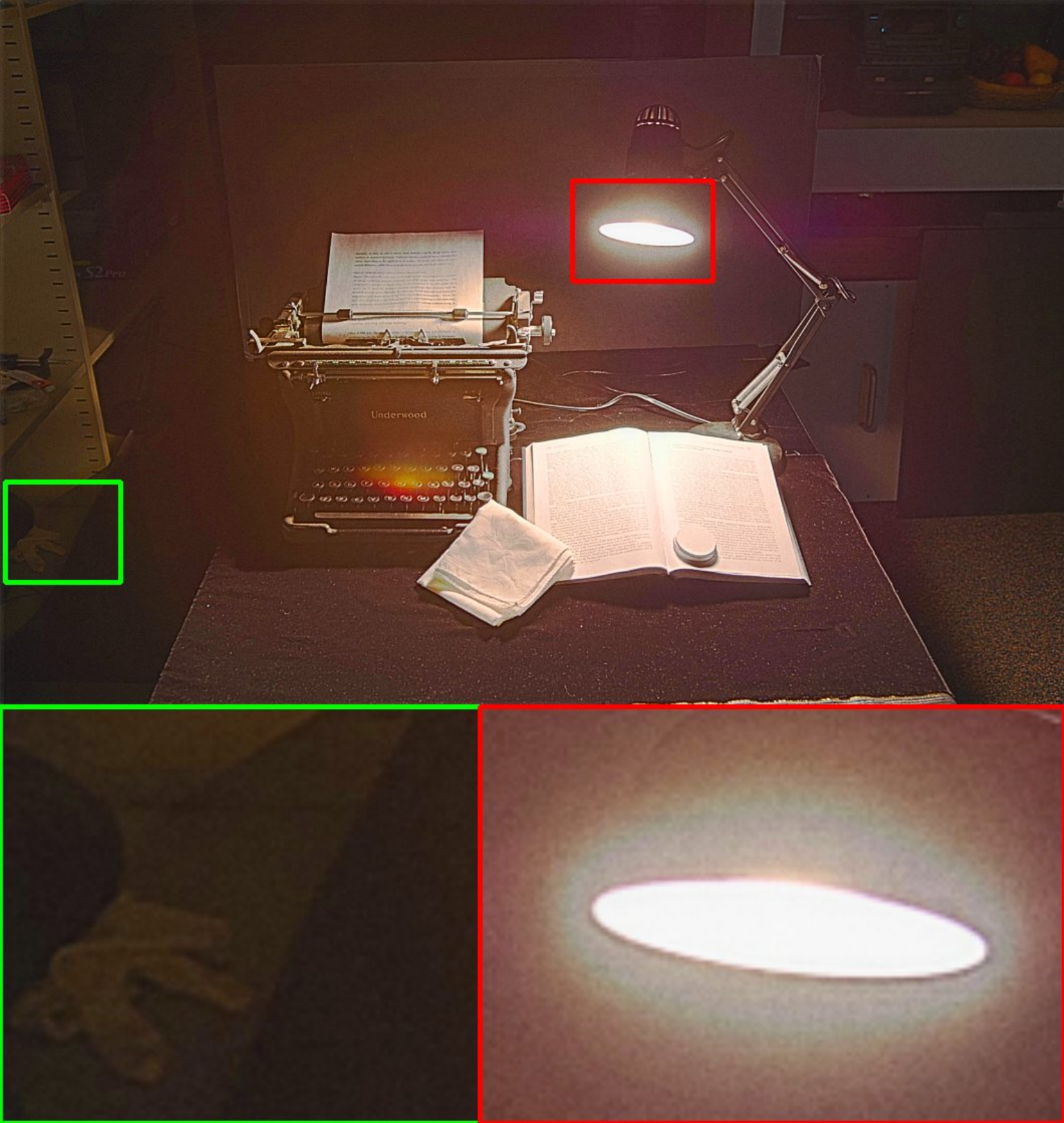}&
				\includegraphics[width=0.155\textwidth]{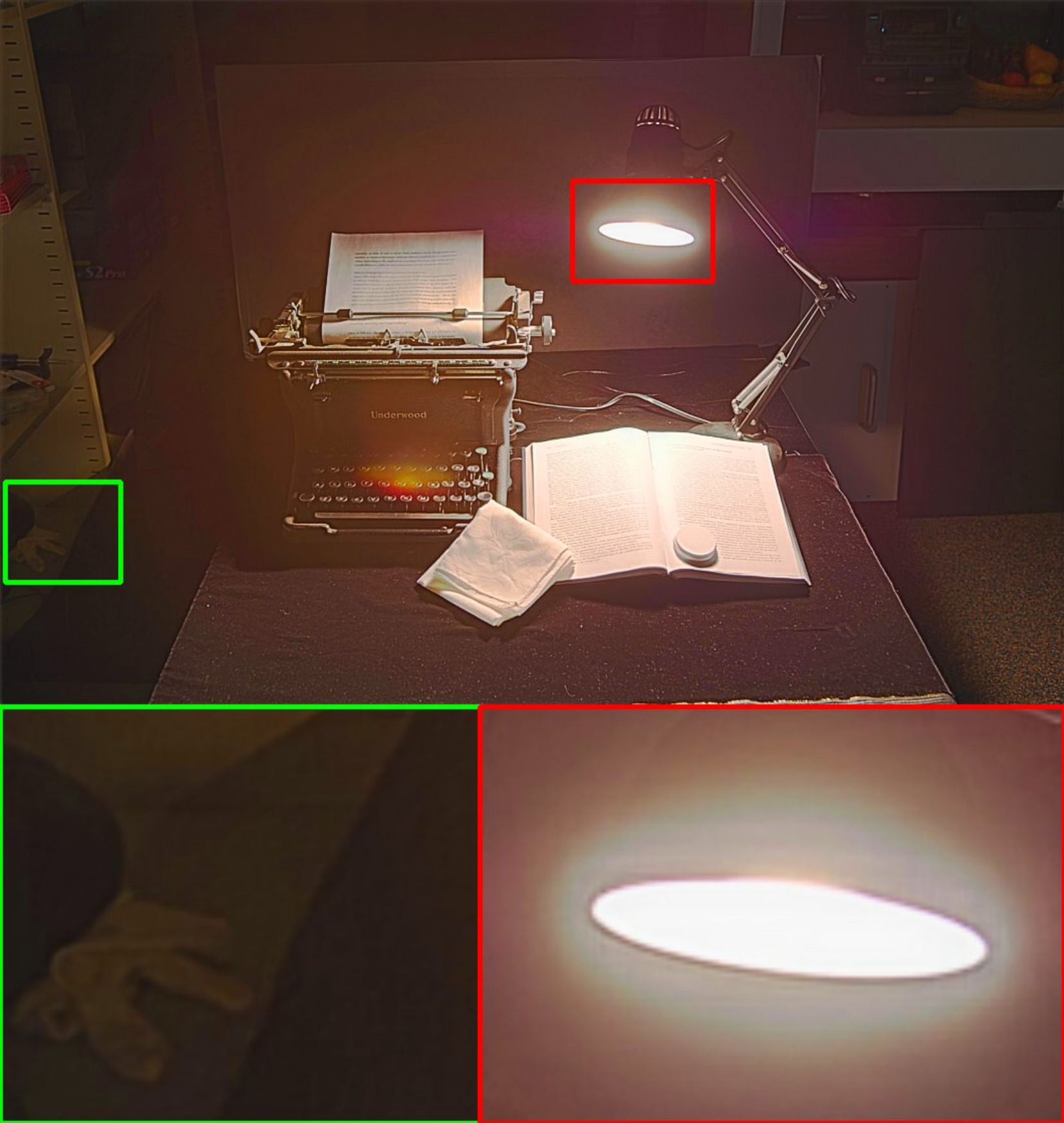}\\
				Input 2&  (c) w/o Denoise& (d) Full\\
			\end{tabular}
		\end{center}
		\caption{Visual comparison of ablation experiments on GCM and Denoise.}
		\label{fig:ablation1}
	\end{figure}

	\subsection{Experimental Results}
		We compare our method with ten state-of-the-art methods including DSIFT \cite{liu2015dense}, GFF \cite{li2013image}, SPEMEF \cite{ma2017robust}, FMMEF \cite{li2020fast}, MEFNet \cite{ma2019deep}, MEFGAN \cite{xu2020mef}, IFCNN \cite{zhang2020ifcnn}, U2Fusion \cite{xu2020u2fusion}, AGAL \cite{liu2022attention}, DPEMEF \cite{han2022multi}. To ensure a rigorous and unbiased comparison, we followed a standardized approach by implementing all competing methods using default parameters and publicly available pre-trained models. Furthermore, to fully demonstrate the superiority of our method, we conducted experiments on four distinct datasets \cite{cai2018learning,zhang2021benchmarking,gallo2009artifact,ma2015perceptual}. It is noteworthy that the prevalent quantitative evaluation metric for multi-exposure image fusion in computer vision involves comparing the results with under-exposed and over-exposed images. Consequently, to achieve optimal results, the weights of each image in R in Eq~\ref{equ:14} and Eq~\ref{equ:16} should be set to {0, 0.5, 0.5, 0}. Likewise, to attain the best visual effect, the corresponding weights should be configured to $\{0.25, 0.25, 0.25, 0.25\}$.

	\textbf{Qualitative Results.} 
		First, we conduct qualitative comparisons on four datasets. For comprehensive comparison, we analyze these results in Figure~\ref{fig:SICE} from two aspects, i.e., details and color. Observed that in these figures, our method obviously performs the best while other methods fail in details and color. DSIFT, GFF, FMMEF, MEFNet, U2Fusion usually introduce brightness inconsistency and halo artifacts. Although SPDMEF and MEFGAN avoid above problems, they suffer from color distortion. Apart from that, DPEMEF, AGAL fail to maintain the detail information exist well in source images since these algorithms adopt elementary fusion rules, which can't preserve details well during forward propagation. The IFCNN model exhibits strong performance in preserving intricate details, yet the resulting images suffer from issues related to over-sharpening and a tendency towards monochromatic and weakly saturated coloration. Given that our network incorporates GCM and TEM, we can effectively mine latent information from the source image using the former, while the latter can fully supplement this information into the generation process, resulting in images with rich and intricate details. Moreover, our proposed color-related algorithm, CE, ensures that the final generated image has vivid and diverse colors.

	\textbf{Quantitative Results.} 
		In addition to qualitative comparisons, we use PSNR~\cite{jagalingam2015review}, CS~\cite{yang2021reference}, CC~\cite{shah2013multifocus}, NMI~\cite{hossny2008comments}, \emph{$Q_{ncie}$}~\cite{WANG2005287} to evaluate our method. The Peak Signal-to-Noise Ratio (PSNR) represents the degree of distortion between the fused image and its original. It is calculated by dividing the peak signal level by the root-mean-square error between the fused and original images. Colorfulness (CS) measures the saturation and vividness of colors in an image. Correlation Coefficient (CC) measures the linear correlation between the pixel intensities of two images and quantifies their similarity. Normalized Mutual Information (NMI) measures the mutual information between two images, which describes how much information about one image is shared with the other. \emph{$Q_{ncie}$} is an image quality assessment method for multi-resolution analysis, mainly used to compare the degree of change between two images.
		Table~\ref{tab:main_metrics1} and Table~\ref{tab:main_metrics2} presents the objective evaluation for all comparison methods on four datasets. It can be seen that most of the five indicators have achieved the best results.

	\subsection{Ablation Study}
		To illustrate the effectiveness of each structure, we conducted ablation experiments on GCM, TEM, CE and the Denoise module. These experiments can be categorized as module ablation and structural ablation. Module ablation include GCM, CE and Denoise module ablation. Structural ablation is about TEM and the specific design is shown in Figure~\ref{fig:Ablation_TEM}.
	
		\textbf{Analyze the Effectiveness of GCM.} Our proposed GCM module is designed to extract and incorporate potential information from the source image, thereby enhancing the details of the entire fusion process. To evaluate the effectiveness of this module, we set two experiments. One is that we replaced the enhanced image generated by the GCM module with a copy of the original image, ensuring that the model parameters remain as consistent as possible. The other one is that we set the GCM's parameter n to 1. As illustrated in Figure~\ref{fig:ablation1}-(d), we can observe implicit details in the red area. However, in Figure~\ref{fig:ablation1}-(a)(b), the details are drowned in dark regions. It's obvious that the information in the original image is not fully utilized. The quantitative comparisons is shown in Table~\ref{tab:GCM}. 
		
		\textbf{Illustrating the Effectiveness of TEM.} The TEM mainly consists of attention-guided detail completion and a simple and efficient attention mechanism. The proposed attention-guided detail completion is refined from self-attention mechanism and it use the enhanced images for full detail completion of the fusion process. To evaluate the effectiveness of TEM, we designed five sets of comparative experiments. The specifics of the ablation experiments are illustrated in Figure~\ref{fig:Ablation_TEM}-(\uppercase\expandafter{\romannumeral1},\uppercase\expandafter{\romannumeral2},\uppercase\expandafter{\romannumeral3},\uppercase\expandafter{\romannumeral4},\uppercase\expandafter{\romannumeral5}). 
		To evaluate the effectiveness of attention map, we devised one set of comparative experiments. It involves directly incorporating the results of $f1'$, $f2'$, and the Decoder output without utilizing the attention map $M_1$ and $M_2$. The specific design of this ablation is alse illustrated in Figure~\ref{fig:Ablation_TEM}-(\uppercase\expandafter{\romannumeral5}). The quantitative and quantitative comparisons are shown in Figure~\ref{fig:ablation2} and Table~\ref{tab:TEM}, respectively.

		\textbf{Discuss the Structure of CE.} CE is an algorithm that modifies the RGB data using the S and L channels in the HSL color space, thereby increasing the saturation of the image. As can be observed in Figure~\ref{fig:CE}, the network with CE produces images with more vivid and diverse colors and more saturation on almost every color scale compared to the results generated by the network without CE. The quantitative result is shown in Table~\ref{tab:GCM}.

		\textbf{Analysize the Effectiveness of Denoise.} Low-light images typically contain more noise than regular images, and this noise is further amplified during the fusion process. Furthermore, our GCM module enhances image details, which may also amplify the noise present in the source image. Therefore, denoising is crucial for effective image enhancement. As demonstrated in Figure~\ref{fig:ablation1}, if the enhanced image is not denoised, the resulting fusion image will contain a significant amount of noise. 	
		Apart from that, the quantitative results of the ablation study are shown in Table~\ref{tab:GCM}. Our complete network achieves the highest performance across all metrics. 
	
\section{CONCLUSION}
	In this paper, we proposed BHF-MEF, a novel multi-exposure image fusion method via boosting the hierarchical features. The gamma correction module processed the over/under exposure images to explore the information latent in source images. Then the texture enhancement module fully supplement the details in the fusion process and the proposed color enhancement algorithm resulting in an image with enhanced color information. Experimental results illustrated that the developed method is not only numerically superior to existing methods, but also visually richer in detail and color.


\begin{acks}
	This work is supported by Natural Science Foundation of China (Grant No. 62202429, U20A20196) and Zhejiang Provincial Natural Science Foundation of China under Grant No. LY23F020024, LR21F020002.
\end{acks}


\bibliographystyle{acm}
\balance
\bibliography{sample-base}

\end{document}